\documentclass[preprint,authoryear,10pt,a4paper]{elsarticle}
\usepackage{lineno}
\usepackage{color}
\usepackage{geometry}
\usepackage[colorlinks,linkcolor=blue]{hyperref}
\usepackage{float}
\usepackage{graphicx}
\usepackage{subfigure}
\usepackage[utf8]{inputenc} 
\usepackage[T1]{fontenc}    
\usepackage{url}            
\usepackage{booktabs}       
\usepackage{amsfonts}       
\usepackage{nicefrac}       
\usepackage{microtype}      
\usepackage{lipsum}		
\usepackage{doi}
\usepackage{amsmath}
\usepackage{amssymb}
\usepackage{enumerate}
\usepackage[title]{appendix}
\usepackage[ruled,vlined]{algorithm2e}

\modulolinenumbers[1]

\journal{Transportation Research Part C: Emerging Technologies}

\biboptions{authoryear}

\begin{document}

\begin{frontmatter}

\title{Potential destination discovery for low predictability individuals based on knowledge graph}

\author[mymainaddress]{Guilong Li}
\ead{liglong3@mail2.sysu.edu.cn}

\author[mymainaddress]{Yixian Chen}
\ead{chenyx96@mail2.sysu.edu.cn}

\author[mymainaddress]{Qionghua Liao}
\ead{liaoqh3@mail2.sysu.edu.cn}

\author[mymainaddress,mymainaddress1]{Zhaocheng He\corref{mycorrespondingauthor}}
\cortext[mycorrespondingauthor]{Corresponding author}
\ead{hezhch@mail.sysu.edu.cn}

\address[mymainaddress]{Guangdong Provincial Key Laboratory of Intelligent Transportation Systems, School of Intelligent Systems Engineering,
Sun Yat-sen University, Guangzhou 510275, China}
\address[mymainaddress1]{The Pengcheng Laboratory, Shenzhen 518055, China}

\begin{abstract}
Travelers may travel to locations they have never visited, which we call potential destinations of them. Especially under a very limited observation, travelers tend to show random movement patterns and usually have a large number of potential destinations, which make them difficult to handle for mobility prediction (e.g., destination prediction). In this paper, we develop a new knowledge graph-based framework (PDPFKG) for potential destination discovery of low predictability travelers by considering trip association relationships between them. We first construct a trip knowledge graph (TKG) to model the trip scenario by entities (e.g., travelers, destinations and time information) and their relationships, in which we introduce the concept of private relationship for complexity reduction. Then a modified knowledge graph embedding algorithm is implemented to optimize the overall graph representation. Based on the trip knowledge graph embedding model (TKGEM), the possible ranking of individuals' unobserved destinations to be chosen in the future can be obtained by calculating triples' distance. Empirically. PDPFKG is tested using an anonymous vehicular dataset from 138 intersections equipped with video-based vehicle detection systems in Xuancheng city, China. The results show that (i) the proposed method significantly outperforms baseline methods, and (ii) the results show strong consistency with traveler behavior in choosing potential destinations. Finally, we provide a comprehensive discussion of the innovative points of the methodology.

\end{abstract}


\end{frontmatter}


\section{Introduction} \label{sec:introduction}

Destination prediction has always been the focus of research in the transportation field. As the increasingly rapid development of surveillance technologies in recent years, massive mobility data at individual level can be passively collected from both invasive and non-invasive detection systems. Such rich data after long-term observation enables traffic engineers to explore and propose many destination prediction approaches for different applications. However, things will be more difficult if only a short period data is available or at the early stage of observation. \cite{gonzalez2008understanding} found that in contrast with the random trajectories predicted by random walk and other models, human trajectories show a high degree of temporal and spatial regularity based on six-month observation. That is, the discovery of human mobility patterns or regularity requires sufficient observations, which also applies to individual trip activities. Under limited observation, the data records are not adequate to mine the traveler's mobility patterns, resulting in them appearing to make random trips and showing low predictability. On the other hand, insufficient observations mean a significant number of possible destinations of individuals are not exposed. We define these destinations as the potential destinations of individuals, i.e., locations that have not been visited by a traveler but will be visited in the future where short or long time horizon (e.g., in the next week or in the next year) is possible. Individuals' potential destinations should not be ignored to form a complete perception of their trip behavior. Especially with limited observations, under which travelers will frequently explore new destinations, some prediction methods will not be applied without first discovering the potential destinations of individuals. In other words, potential destination discovery is basic work supporting tasks like next destination prediction. For instance, for a traveler with only several observed trips (e.g., 3) and different destinations, we cannot reasonably determine the candidate set of predicted destinations without its potential destinations being discovered.

Potential destination discovery (or prediction) of low predictability travelers aims to discover the potential destinations of an individual based on limited historical records from the whole travelers with low predictability. Although many researchers have extensively studied in the field of destination prediction, few existing studies have considered travelers with low predictability, or travelers' potential destination under short-term observation period. Besides, most of previous research pay little attention to potential destinations (\cite{zhao2018individual},\cite{ebel2020destination},\cite{manasseh2013predicting}), or even ignores them totally when predicting. (\cite{gambs2012next},\cite{lu2013approaching}). In \cite{hariharan2004project}, trip destinations were just defined as the places travelers have visited. \cite{krumm2006predestination} is the first work that introduces a concept of new destination with a similar meaning of potential destination, in order to improve overall prediction performance. However, no validation is performed on them. Since then, many related works have taken the potential destination or other similar concepts into consideration, but they have not developed methods specialized for potential destinations, nor have they done specialized evaluations (\cite{zhao2021mdlf}). Besides, they all failed to directly make prediction for low predictable individuals. One way is that most previous studies focus on implementing prediction by mobility pattern extraction or regularity analysis based on highly predictable individuals, where low predictable individuals as a minority receive less attention (\cite{zhao2018individual,rossi2019modelling,manasseh2013predicting}).
Additionally, in some research, travelers with relatively few trips are treated as abnormal individuals to adapt the proposed prediction methods better. For instance, \cite{cheng2021probabilistic} filtered out individuals whose number of trips is less than 20. Under such settings, the majority of travelers to be predicted are highly predictable.

As our goal is to discover (or predict) potential destinations for individuals with low predictability, in this paper, we propose a generic framework that can be adapted to all kinds of mobility data under such conditions. To achieve this, the first problem facing us is how to enhance the predictability of a traveler when mobility pattern extraction is almost impossible for data sparsity. According to \cite{gonzalez2008understanding}, human travel patterns shared inherent similarity despite of the diversity among their travel history. This motivates us to leverage useful information from other travelers, instead of a single individual. In this way, an unpredictable individual will become a predictable one as part of the whole. Then we need to deal with other remaining 3 challenges. (1) How to organize all individuals' mobility data into one structure so that they can easily learn effective information from each other. (2) How to develop a proper optimization algorithm that can implement an overall optimization solution. (3) how to design meaningful metrics to validate the prediction results. To deal with these problems, we propose a Potential Destination Prediction Framework based on Knowledge Graph (PDPFKG). Firstly, we present a trip knowledge graph (TKG) schema to organize the individual-level mobility data into a well-defined structure, in which the fundamental objects (e.g., travelers, destinations and time information) in the transportation field are modeled as entities with relevant relationships among them. Then, we adopt TransH, a popular knowledge graph embedding model proposed by \cite{wang2014knowledge}, on TKG with modifying the training strategy and objective function. Afterward, the entities and relationships in the TKG are all projected into a vector space and their positions are optimized based on the associations among them. In this way, the possibility ranking of each traveler's unobserved destination can be estimated by their core triples' distances.
Finally, we propose two result refinement ways to support the aggregated and individual level evaluations. On this basis, three metrics are adopted for aggregated-level evaluations from different perspectives. In addition, average and deviation are chosen to explore the overall performance at the individual level and the inter-individual variation.

To validate PDPFKG, we apply it to an anonymous vehicular dataset from 138 intersections in Xuancheng city, collected from a video-based detection system based on automatic vehicle identification (AVI) technology provided by the Chinese local transport bureau. Experiment results demonstrate that the results given by PDPFKG present a strong correlation with the potential destination choice behaviour of individuals on aggregated-level. In addition, individual performance shows a single-peaked distribution, and the introduction of new valid information can improve individual performance overall, indicating that PDPFKG is scientific and has great potential.

In summary, this paper mainly makes the following contributions.

\begin{itemize}
    \item We develop a generic framework PDPFKG based on knowledge graph for potential destination discovery of individuals with low predictability.

    \item We customize PDPFKG to tackle a set of challenges, including a trip knowledge graph schema for mobility objects' association organization, an introduction of private relationships for complexity reduction, an adapted embedding model for learning relationships information.

    \item We demonstrate the validity of PDPFKG through extensive experiments and discussions based on a city-scale vehicular dataset, and we reveal that the association information of travelers is worth mining. 
\end{itemize}

\section{Literature review} \label{sec:LR}

During the last two decades, a great deal of work has been devoted to trip destination and human mobility prediction. This section aims at reviewing recent studies exploring destination prediction problem from two aspects: data condition and method. In Table~\ref{tab:LR}, the summary of reviewed studies is presented, where the middle three columns (scenario, data type, and data size) show the data condition of studies and the last column illustrates the used method. Specifically, the scenario column indicates the specific scene of the study; the data type column shows whether the study used OD-only data (e.g., smart card data) or trajectory records (e.g., location-based GPS data and bluetooth data); and the data size column shows time span or average number of records for each individual.

\begin{table}[ht]
\centering
\caption{Representative studies of human mobility and destination prediction.}
\label{tab:LR}
\begin{tabular}{ccccc}
\hline
Studies & Scenario & Data type & Data size & Method  \\ \hline

\cite{ashbrook2002learning} &  Human mobility  & Trajectory  & 4 months & Markov model  \\ 
\cite{krumm2006predestination} & Vehicle & Trajectory & 43 records/id & Bayesian inference\\
\cite{burbey2008predicting} & Human mobility & Trajectory & 11 weeks & Partial-match\\
\cite{nadembega2012destination} & Human mobility & Trajectory & 9 months & Cluster-based\\
\cite{noulas2012mining} & Human mobility & Trajectory & 5 months & M5 model tree\\
\cite{xue2013destination} & Taxi & Trajectory & 3 months & Bayesian Inference\\
\cite{chen2019mpe} & Taxi & Trajectory & 12 months & Deep learning\\
\cite{wang2020attention} & Sharing bike & OD-only & 40 records/id & Deep learning\\
\cite{besse2017destination} & Taxi & Trajectory & 12 months & Distribution-based\\
\cite{imai2018early} & Human mobility & Trajectory & 2 months & Probabilistic model\\
\cite{neto2018combining} & Vehicle & Trajectory & 3 months & Markov model\\
\cite{dai2018cluster} & Sharing bike & Trajectory & 12 months & Cluster-based\\
\cite{zhao2018individual} & Public transit & OD-only & 24 Months & Markov model\\
\cite{rossi2019modelling} & Taxi & Trajectory & 12 months & RNN\\
\cite{zong2019trip} & Vehicle & Trajectory & 2 months & Hidden markov model\\
\cite{rathore2019scalable} & Taxi & Trajectory & 869 records/id & Markov chain\\
\cite{ebel2020destination} & Taxi & Trajectory & 1731 records/id & LSTM\\
\cite{mo2021individual} & Subway & OD-only & 30 months & Hidden markov model\\
\cite{liang2021vehicle} & Vehicle & Trajectory & 528 records/id & Machine Learning\\
\cite{jiang2021dp} & Vehicle & OD-only &  7 months & Bayesian-based\\
\cite{sun2021joint} & Vehicle & Trajectory &  11 months & LSTM\\
\cite{zhao2021mdlf} & Subway & OD-only &  over 1 month  & Deep learning\\
\cite{cheng2021probabilistic} & Subway &  OD-only &  3 month   &  Probabilistic model\\
\hline
\end{tabular}
\end{table}


Most of previous studies mainly focus on prediction of highly predictable travelers. In general, a longer observation period means higher predictability of the individual. As shown in Table~\ref{tab:LR}, the majority of existing studies were based on long-term observation, which results in adequate records and guarantees the majority of individuals have high predictability. Besides, some studies carried out data processing to make the target individuals sufficiently predictable.For example, \cite{manasseh2013predicting,neto2018combining} limited targets to specific highly regular individuals. \cite{zhao2018individual} selected users with at least 60 active days of transit usage, which excluded occasional users and short-term visitors. \cite{cheng2021probabilistic} only focused on passengers with a minimum of 20 observations in 3 months. \cite{imai2018early} filtered out individuals whose trip number was less than $5$. \cite{zhao2021mdlf} excluded individuals whose active days were less than $2$. Besides, \cite{wang2020attention} excluded trip records with relatively shorter travel time. \cite{liang2021vehicle} excluded trips of individuals whose routing patterns are different from target individuals. \cite{alvarez2010trip} and \cite{chen2019mpe} implied the trip regularity of individuals existed under their observation period.

Recently, some studies tried to make predictions for individuals with sparse data. For instance, \cite{wang2017moving} predicted the moving destinations using sparse dataset. \cite{xue2013destination} forecast destinations by solving data sparsity problem. \cite{xue2015solving} handled the problem of data sparsity for destination prediction by sub-trajectory synthesis. \cite{imai2018early} predicted destinations at an early stage when the destinations had not been fully observed. \cite{zhao2021mdlf} challenged destination prediction for occasional trips of individuals, including ones with few trips.
Individuals with sparse data are considered more challenging to predict because of lower predictability, and lower prediction accuracy can be tolerated. Existing studies based on sparse data mainly used trajectory data (\cite{wang2017moving,xue2013destination,xue2015solving,imai2018early}), whose information is more abundant than OD-only data for the process of mobility is recorded. Thanks to this, the above researches deal with sparse data through trajectory synthesis and characteristics extraction.
\cite{zhao2021mdlf} used OD-only data, but there are a large number of regular travelers in its dataset, except for partial inactive individuals. Besides, it excluded individuals with few activity days, whose predictability is very low. 

According to the above analysis, existing studies tend to make predictions for highly predictable individuals whose possible destinations are almost totally observed, while low predictability individuals have not been well considered yet (especially for OD-only data type).

Previous studies paid little attention to potential destinations of individual's trips. This is partly because all possible destinations of individuals are almost entirely observed under long-term observation conditions, for which potential destinations prediction hardly affect the overall results. For instance, \cite{lu2013approaching} achieved high prediction accuracy ignoring potential destinations. On the other hand, it is due to the limitation of existing prediction methods. 

As shown in Table~\ref{tab:LR}, the methods for destination prediction have been dominated by data-driven models (e.g., neural network-based), which became popular with a great quantity of data becoming available. Most of these models adopt supervised learning, and more training data is better, for which sufficient data is required. Considering the principle of these models, they learn the pattern of data and tend to reproduce data that have been trained. These make them poor predictors for sparse data and data without patterns, while they are unable to or hardly predict untrained data. So these models are neither good at dealing with individuals with low predictability (data not regular and sparse), nor good at predicting potential destination (unobserved). For the above reasons, some studies ignored potential destinations when predicting. For example, \cite{lu2013approaching,gambs2012next},  which based on Markov Chain model, can not predict locations that users have never visited before. 

There were some studies that considered potential destinations, but their major attention was still put on observed ones, and no mechanism was specially developed for potential destinations. For example, \cite{jiang2019destination,asahara2011pedestrian} gave the probability of unobserved destinations by statistics of the groups. \cite{neto2018combining} can predict places never visited by the user by combining Markov Model and Partial Matching. \cite{zhao2018individual} shared the spatial choice set of all users, making new destinations for the individual may be predicted. \cite{zhao2021mdlf} utilized crowd feature to handle individual data missing (e.g., individuals have not traveled at the given origin and time), by which destinations an individual never appeared to might be predicted. The ways these studies predict potential destinations are mainly statistical-based (\cite{asahara2011pedestrian,zhao2018individual,neto2018combining,jiang2019destination,zhao2021mdlf}), without establishing and modeling relationships among individuals.  Besides, none of these studies specifically validated the predicted result on unobserved destinations, leading to an unknown performance on potential destination predictions. In summary, these methods are not focused on prediction or discovery of the individual's potential destinations, although they might give results from unobserved destinations.

Potential destination discovery or prediction is somewhat analogous to the recommendation system problem (\cite{wang2013customized,hernandez2020fog}), but we believe it is more challenging. First, urban traffic zones containing multiple types of POIs are hard to classify and label compared to a product, video, or specific POI, which has explicit purpose or attributes. On the other hand, the information about individual search, browsing, scoring, etc., is critical for recommendation systems, while we are not sure exactly where the individual is going and what their purpose is when observing an individual's destination.
Both of these make it difficult to portray or categorize individuals by their observed destinations. Further, potential destination prediction has less common sense or principles to rely on. Most methods of recommendation systems, especially for combating data sparsity, were developed heavily based on prior knowledge or common sense. For example, video recommendations (\cite{davidson2010youtube,lee2017large}) believe that people prefer to choose things that are similar or related to what they have chosen. Besides, prior knowledge, such as people tend to explore the POIs near the ones they favour, is the basis for model construction of POI recommendations (\cite{yin2017spatial,qian2019spatiotemporal}). However, there is little prior knowledge related to potential destination prediction, and these mentioned are not applicable to this task. For example, it is unreasonable to believe an individual who frequents a traffic zone where a hospital is located will visit the traffic zones where other hospitals are located. These considerations indicate significant challenges for potential destination prediction, and they even make it ambiguous whether an individual's potential destination can be predicted or discovered.

\section{Preliminaries}\label{sec:pre}

\subsection{Problem description} \label{sec:pre1}

\begin{figure}[H]
\setlength{\abovecaptionskip}{0.cm}
\centering
\includegraphics[width=0.98\textwidth]{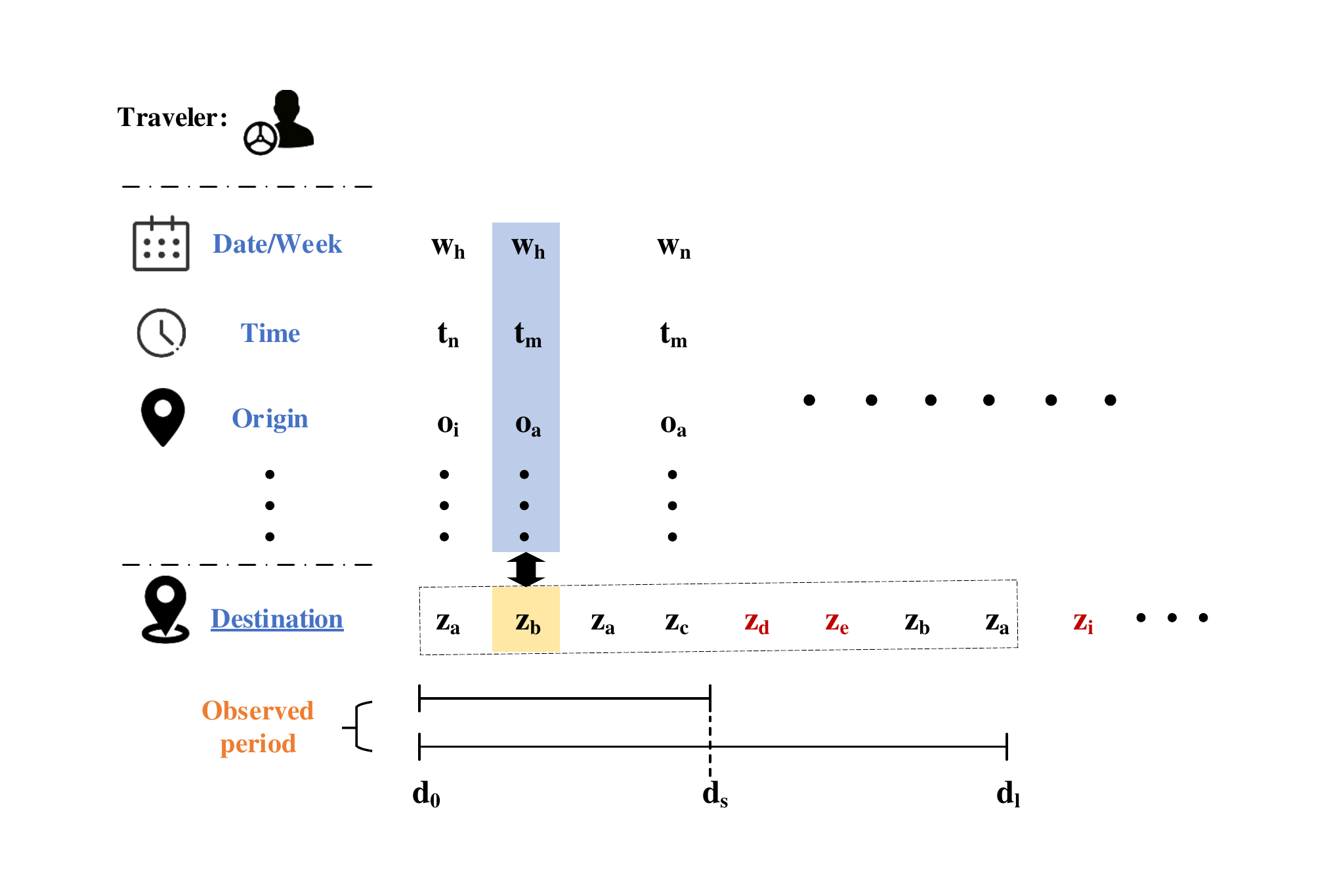}
\caption{Schematic illustration of individual's destination prediction.}
\label{fig:difi_des}
\end{figure}

Fig.~\ref{fig:difi_des} shows the basic elements of individual destination prediction problem. The popular ideas of methods shown in Table.1 can be summarized into two categories. The first one is modeling the destinations (or trajectory) as a sequence like $l = (z_a,z_b,z_a,z_c,z_d,z_e,z_b,z_a)$ (dotted box in Fig.~\ref{fig:difi_des}). On this basis, it learns the transfer probability between elements to execute prediction. For example, $z_b$ is more likely to transfer to $z_a$ will be learned by analysis of $l$ based on this idea.  
The other one aims to construct a mapping from relevant information such as trip time, trip origin (blue box in Fig.~\ref{fig:difi_des}) to the trip destination (yellow box in Fig.~\ref{fig:difi_des}). This idea generally adopts supervised learning for features (e.g. ($w_h,t_m,o_a,\dots$)) to label (e.g. $z_b$) fitting, and achieves prediction when the features are given. Nevertheless, for potential destinations of a traveler like $z_i$ under observed period $T^o=(d_0,d_l)$ shown in Fig.~\ref{fig:difi_des}, these ideas become inapplicable since they are neither trained as the label for the individual nor present in its destination sequence. Potential destinations will be gradually observed with continuous observation. Conversely, a shorter observation period leads to more potential destinations. For example, $z_d$ and $z_e$ would become potential destinations for the traveler in Fig.~\ref{fig:difi_des} if the observed period shortened to $(d_0,d_s)$. In any case, the potential destinations cannot be ignored unless all possible destinations can be observed.

This paper focuses on discovering or predicting the potential destinations of travelers, and the problem can be defined as follows with the notations defined in Table.~\ref{tab: summary_notations}. Given a dataset recording travelers' trip activities during a short observation period $T^o$, and a specific traveler $v_n$, we aim to give the possibility ranking of each unobserved traffic zone ($z_j \in Z-Z^o_n$) being visited by $v_n$ in the  future.

\subsection{Concepts and notations}\label{sec:pre2}

\begin{figure}[h]
\setlength{\abovecaptionskip}{0.cm}
\centering
\includegraphics[width=0.72\textwidth]{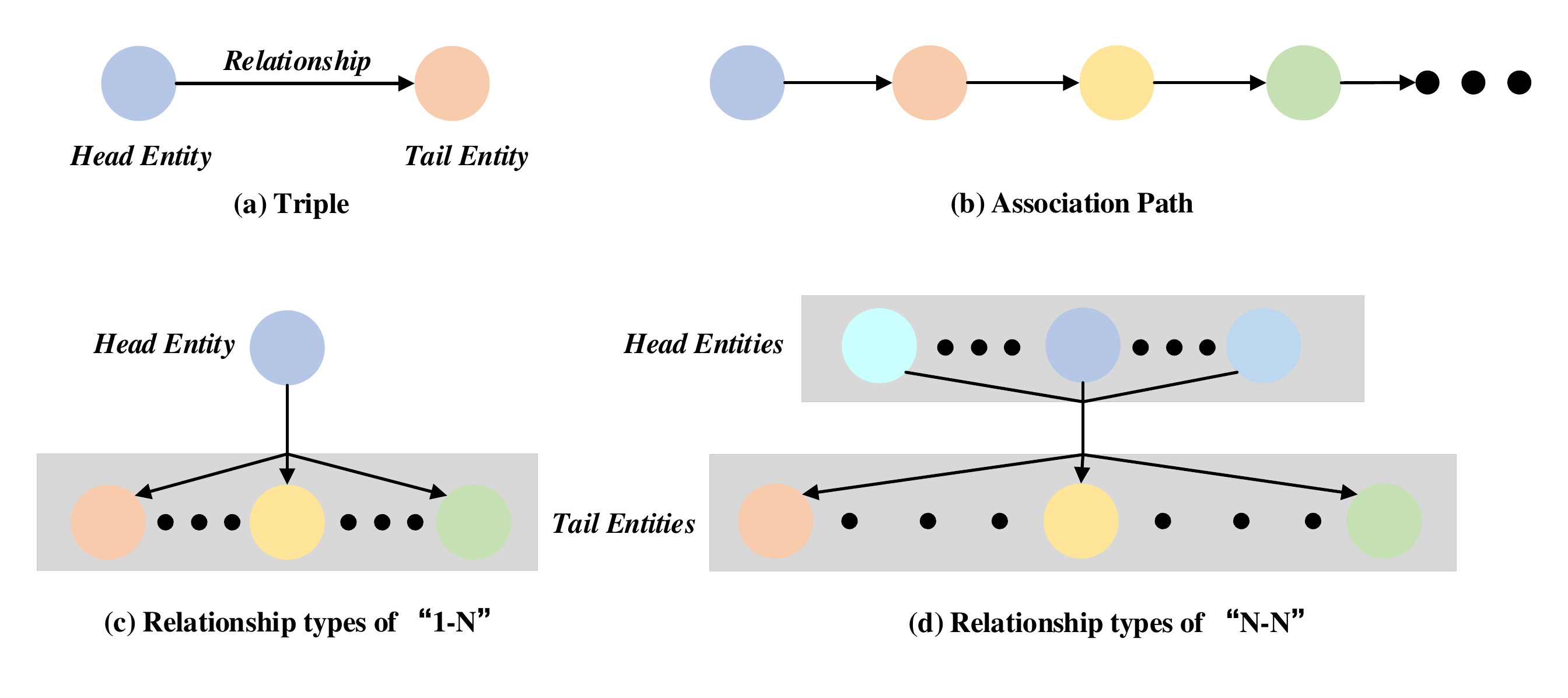}
\caption{Notations of knowledge graph.}
\label{fig:introduction of KG}
\end{figure}

First, we introduce some basic concepts used in this paper about the knowledge graph. Following the previous study(\cite{ji2021survey}), we use $G$ to represent a knowledge graph, and it can be expressed as $G=\{E,R,F\}$ where $E$, $R$ and $F$ are sets of entities, relationships and facts. A fact is denoted as a triple $(h,r,t)\in F$, where $h$ and $t$ are elements of the entity set $E$ and $r$ is the element of $R$. For the triple $(h,r,t)$, $r$ generally has a direction from $h$ to $t$. The triple with a directional relationship shown in Fig.~\ref{fig:introduction of KG}(a) can be represented as $(h:Entity)-[r:Relationship]\rightarrow(t:Entity)$, in which entities represented by $h$ and $t$ are called head entity and tail entity, respectively. The triple is the unit structure in knowledge graph and the information expressed by it is called fact. A specific triple expresses a specific fact. 

The direct association of entities can be expressed by triples, and the indirect association information of them can be expressed by association path (or meta-path by \cite{sun2012mining}) in knowledge graph. Entities that are not directly related but can be associated with one or more other entities are considered to be indirectly associated. The path that makes these entities associated is called the association path. It can be considered as a chain of multiple connected triples, or a sequence of alternating entities and relationships as shown in Fig.~\ref{fig:introduction of KG}(b). The formation of an association path depends on different triples containing the same entity, and they should be the head entity and tail entity, respectively.

The relationships of a knowledge graph can be divided into four types of complexity: $1-1$, $1-N$, $N-1$, and $N-N$ ($N>1$). The former represents the number of head entities connected to the relationship, and the latter represents the number of its tail entities. For example, the relationship of $1-N$ type means that there is only one head entity and its tail entity is multiple, as shown in Fig.~\ref{fig:introduction of KG}(c). The relationship of $1-1$ type that has only one head and one tail entity is called simple relationship. Relatively, $1-N$, $N-1$ and $N-N$ are called complex relationship, of which $N-N$ is the most complex type, as shown in Fig.~\ref{fig:introduction of KG}(d). For a complex relationship, its complexity depends on the quantity of its head and tail entities, i.e., the value of $N$. The more the quantity, the more complex it is.

Other crucial notations in this paper are summarized in Table.~\ref{tab: summary_notations}.

\begin{table}[ht]
\centering
\caption{Summary of notations.}
\label{tab: summary_notations}
\begin{tabular}{l|l}
\hline
       Notations               & Description  \\
                      \hline
$V$  & Set of all low predictability vehicle individuals.   \\
$Z$  & Set of all traffic zones.   \\
$v_n$  & Vehicle individual $n$ in $V$.    \\
$z_j$         & Traffic zone $j$ in $Z$.        \\
$T^o$        & Observation period.     \\
$T^f$          & The period considered in the future (after $T^o$).     \\
$\textbf{\textit{M}}$         & The trip knowledge graph embedding model (TKGEM).        \\
$a^e_{n,j}$         & Predicted ranking of $z_j$ for $v_n$ given by $\textbf{\textit{M}}$.\\
$a^h_{j}$         & Hotness ranking of $z_j$.\\
$a^c_{n,j}$         & Combined ranking of $z_j$ for $v_n$.\\
$Z_n^o$        & Set of traffic zones traveled as the destination by $v_n$ within $T^o$. \\
$Z_n^f$        & Set of traffic zones traveled as the destination by $v_n$ within $T^f$. \\

$U$        & Distribution of aggregated potential destination's predicted rankings. \\
$H$        & Distribution of the average of individuals' predicted results. \\

\hline
\end{tabular}
\end{table}

\section{Methodology}\label{sec:Method}

\subsection{Trip knowledge graph construction}\label{sec:Structure}

Individual-level trip data is always displayed in tabular form, where each record describes the information (e.g., vehicles, origins, departure time and destinations) of one trip made by a user. It benefits trip information retrieval and storage, but the associations between individuals are separated, which is valuable for destination prediction, especially when the individual data is sparse.
To efficiently represent the association information among individuals, we propose to apply knowledge graph to organize all the data into one structure where the mobility-related objects are properly connected by designing a trip knowledge graph schema. Specifically, it includes two steps: entity extracting and relationship building.

\subsubsection{Entity extracting}

At this step, we have to decide which types of entities should be included in trip knowledge graph. As our goal is to infer potential destinations on a domain-specific knowledge graph, the extracted entities should serve the interests of the prediction task. Further, to make our graph model more general with all kinds of mobility data, we only consider the most common and available elements.

First, a type of entity representing the traveler's (vehicles) identity is required for prediction at the individual level, and we denoted it as $Veh\_id$. Then each entity of $Veh\_id$ corresponds to a specific individual. Similarly, the element of spatial geography is also needed to represent the destination and origin of trips. Next, we considered the information relevant to the choice of trip destinations. 
\cite{yuan2013time} pointed out the significance of the time factor in the points of interest (POI) recommendation task. In addition, \cite{zong2019trip} boosted the effectiveness of its model on the next location prediction task by adding weekday versus holiday information. Therefore, both the factors of POI and trip time are extracted as entities.  

In summary, we determine the entity types whose meanings are shown in Table~\ref{tab:KG_entity}, where the entity of $Zone$ is a spatial concept used to represent the destination and origin of the trip (see Section~\ref{sec:data describe} for detail). When extracting entities, we need to ensure that they are consistent with the real world. In other words, different entities must uniquely represent one actual object of the real world. For example, a zone only responds to one entity of $Zone$, although it was visited by different individuals and appeared in multiple records. 

\begin{table}[ht]
\centering
\caption{Entity types of the trip knowledge graph.}
\label{tab:KG_entity}
\begin{tabular}{l|l}
\hline
\multicolumn{1}{l}{\textbf{Entity Type}}      & \multicolumn{1}{c}{\textbf{Meaning}}      \\ 
\hline

Veh\_id & Unique identification of the individual \\
Day\_nat    &Nature of the day, including working day and holiday   \\
Time\_span & Time span of the day, e.g. morning peak  \\
Zone    & Traffic zone           \\
POI    & Point of interest  \\ \hline

\end{tabular}
\end{table}

\subsubsection{Relationship building}

Relationship building is to describe the relationship of different types of entities. According to Section \ref{sec:pre}, entities and the relationship between them form a triple, which describes a fact. Hence, relationship building is essentially constructing various triples for describing facts. Therefore, what facts should be expressed is the issue in this step. According to entities extracted in Table~\ref{tab:KG_entity}, the facts that need to be described in trip knowledge graph can be divided into two categories: 1) Individual historical trips, including historical trip destinations, etc.; 2) Traffic zone contains POI. 
These require building relationships between $Veh\_id$ and $Time\_span$, $Day\_nat$, and $Zone$ respectively, as well as $Zone$ and $POI$.

In Section \ref{sec:pre}, we have introduced that the relationships have different degrees of complexity. It has no impact on humans' understanding but for knowledge graph embedding. 
To handle complex relationships, embedding models will also be more complex. In addition, the more complex a relationship is, the higher the dimension required to describe it. On the other hand, the embedding dimension of relationships with different types is usually the same. Thus, the principles of relationship building are as follows. 1) Balance the complexity of different type relationships; 2) Avoid excessive complexity of relationships; 3) Make the complexity of relationships independent of the data scale. These three principles will ensure: 1) The optimal dimension of each type of relationship is consistent; 2) The relationship can be expressed in finite dimensions; 3) The model is migratory across datasets.

Based on these principles, building the relationship $Has\_POI$ between $Zone$ and $POI$ is a feasible way to describe the fact that the traffic zone contains POI. In this way, for example, the fact that there are schools located in traffic zone $z_j$ can be represented by $(Zone:z_j)-[Has\_POI] \rightarrow (POI:school)$. Then $Has\_POI$ is a complex relationship of type $N-N$ since a traffic zone may contain multiple POIs, and one kind of POI may be distributed in different traffic zones. However, the complexity of $Has\_POI$ is measurable and not excessive because of the small size of its head and tail entities. Besides, The trip-independent fact description makes it unaffected by the trip data scale.

For building relationships between $Veh\_id$ and the other three types of entity, we take $Veh\_id$ and $Zone$ as an example. Following the way of defining $Has\_POI$, the relationship $Choose\_D$ will be built between $Veh\_id$ and $Zone$. Human beings can interpret this, but it violates all of the principles of relationship building.
First, although both of $Has\_POI$ and $Choose\_D$ are of type $N-N$, $Choose\_D$ is much more complex than $Has\_POI$ for its number of head and tail entities is not at the same level with $Has\_POI$. Second, its complexity varies with the scale of the trip data. For example, as more individuals are considered, the number of its head entities increases accordingly, leading to increased complexity. Lastly, when the number of individuals considered is huge, the dimension required to express it will become unacceptable.
To address the issue, we propose the concept of private relationship. We define the $Choose\_D\_id$ as a group of relationships. Each relationship corresponds to a specific individual, i.e., the head entity of each relationship is only one entity of $Veh\_id$. In other words, each individual has a private relationship of $Choose\_D\_id$. The complexity type of $Choose\_D\_id$ is reduced to $1-N$ versus $Choose\_D$, and it's not excessive in complexity for the number of tail entities ($Zone$) is small. More importantly, it is independent of data scales. For instance, when the quantities of individuals change, the number of relationships of $Choose\_D\_id$ changes accordingly, while the complexity is hardly affected. In addition, $Choose\_D\_id$ is closer to $Has\_POI$ in complexity compared to $Choose\_D$. The same problem exists in building relationships between $Veh\_id$ and other types of entities like $Time\_span$, and we also adopt private relationships to handle them. 

\begin{figure}[h]
    \centering
    \includegraphics[width=0.65\textwidth]{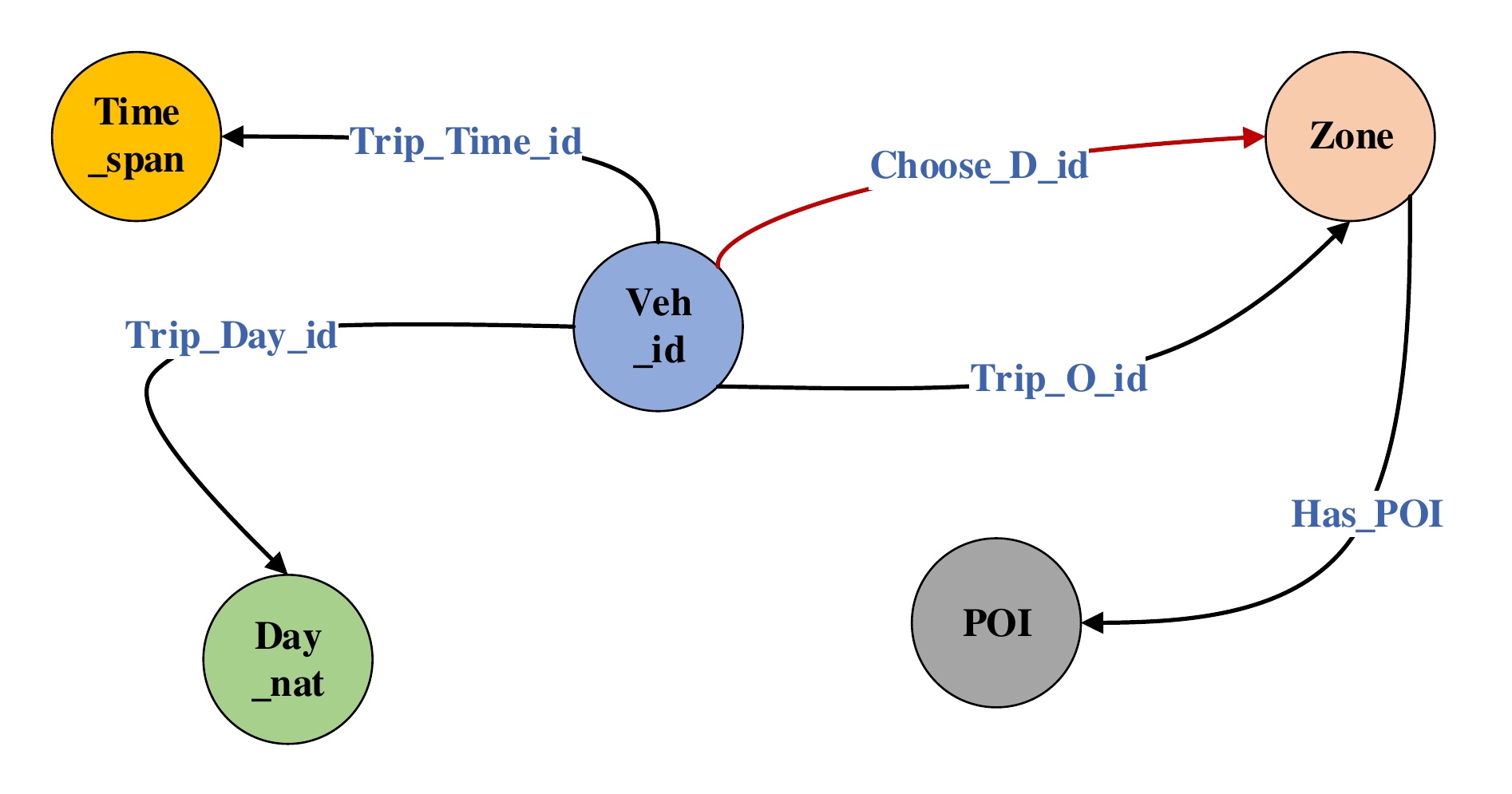}
    \caption{The schema of the trip knowledge graph.}
    \label{fig:KG_structure}
\end{figure}

\begin{figure}[H]
\setlength{\abovecaptionskip}{0.cm}
\centering
\includegraphics[width=13cm]{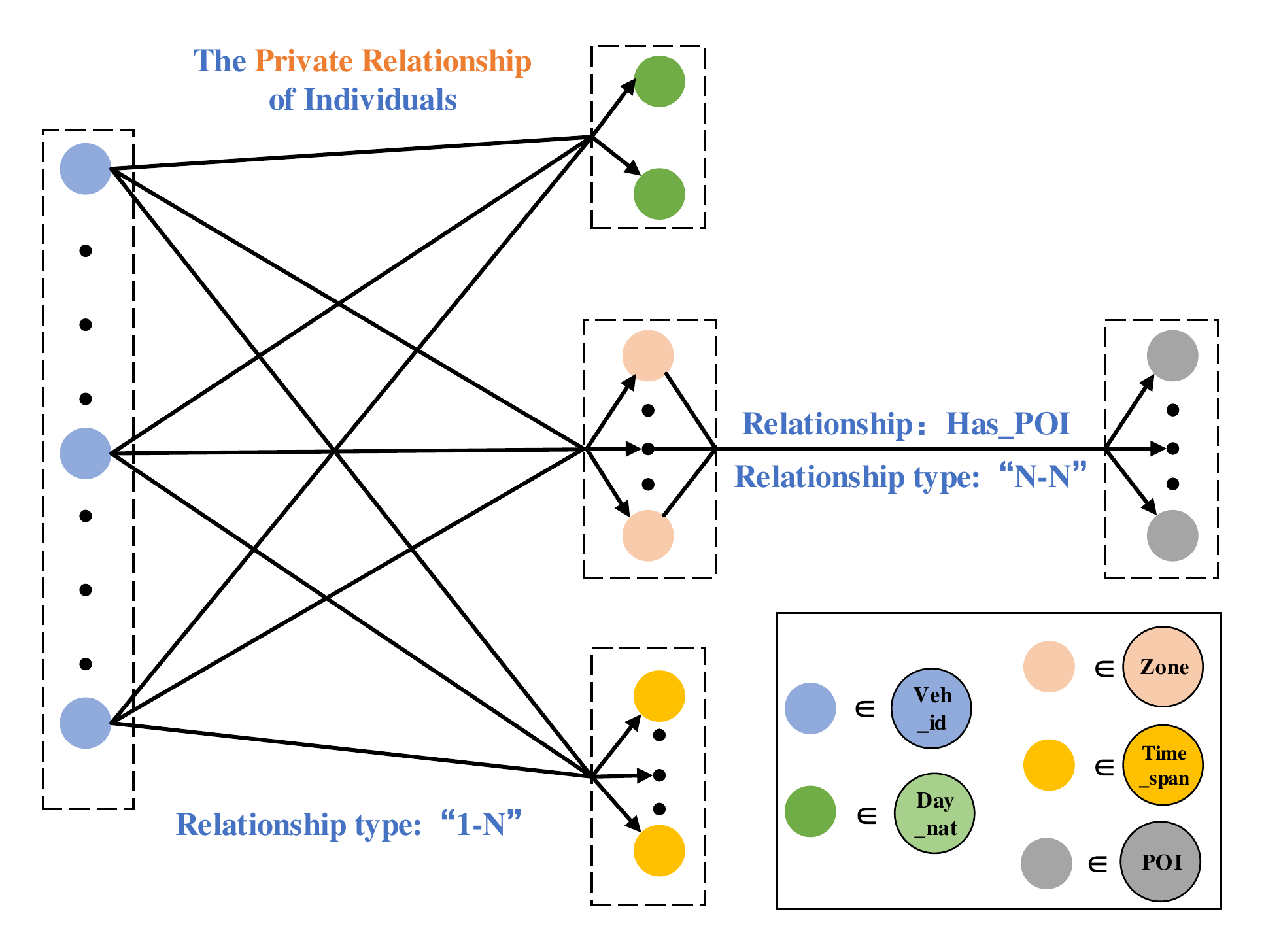}
\caption{Micro entities association of the trip knowledge graph.}
\label{fig:Joint_KG}
\end{figure}

So far, we have completed the construction of the trip knowledge graph (TKG). The schema of TKG is shown in Fig.~\ref{fig:KG_structure} and all types of triple and facts described by them see Table \ref{tab:KG_triple}. The triple marked with $*$ is called core triple, which is the triple that performs the prediction. In TKG, all types of entities have an association path, and the microscopic association between different types of entities is shown in Fig.~\ref{fig:Joint_KG}.

\begin{table}[ht]
\centering 
\caption{Triples of the trip knowledge graph.}
\label{tab:KG_triple}
\begin{tabular}{l|l}
\hline
\multicolumn{1}{c}{\textbf{Triple} } & \multicolumn{1}{c}{\textbf{Fact}}      \\ 
\hline

(Veh\_id)-[Choose\_D\_id]$\rightarrow$(Zone)* & The vehicle chooses the zone as destination to trip \\
(Veh\_id)-[Trip\_O\_id]$\rightarrow$(Zone) & The vehicle trips with the zone as origin \\
(Veh\_id)-[Trip\_Time\_id]$\rightarrow$(Time\_span) & The vehicle trips during the time span (e.g., morning peak)  \\
(Veh\_id)-[Trip\_Day]$\rightarrow$(Day\_nat) & The vehicle trips on the day with the day nature (e.g., workday) \\
(Zone)-[Has\_POI]$\rightarrow$(POI) & The zone contains the point of interest\\ \hline

\end{tabular}
\end{table}

\subsection{Trip knowledge graph embedding}\label{sec:embedding}

In TKG, information about all individuals' trips has been associated. But it is described by natural language that it cannot be computed and does not have the ability to predict. This section will use a modified knowledge graph embedding algorithm to map TKG to a continuous space and obtain parametric expressions of its entities and relationships. 

\subsubsection{Graph embedding models}\label{sec:embedding1}

The purpose of knowledge graph embedding is to map the entities and relationships in a continuous space. After embedding, entities and relationships of knowledge graph will have a parametric representation, and then the knowledge graph is calculable. The translation model is a classical category of models for implementing knowledge graph embedding, including many specific models. All these models consider the tail entity of the triple as a translation of the head entity through relationships. Their differences are mainly in the complexity of the models, and it is reflected in the different number of parameters, which affects the ability of the models to handle complex relationships. In general, the more complex the model, the better it is able to handle complex relationships. Next, we will introduce the generic models of TransE and TransH. The former can most directly represent the principle of the translation model. And the latter deals with complex relationships in a relatively easy way and matches the schema of TKG well. For a more comprehensive understanding of graph embedding algorithms and translation models, the \cite{wang2017knowledge} can be consulted.

\begin{figure}[ht]
    \centering
    \includegraphics[scale=0.58]{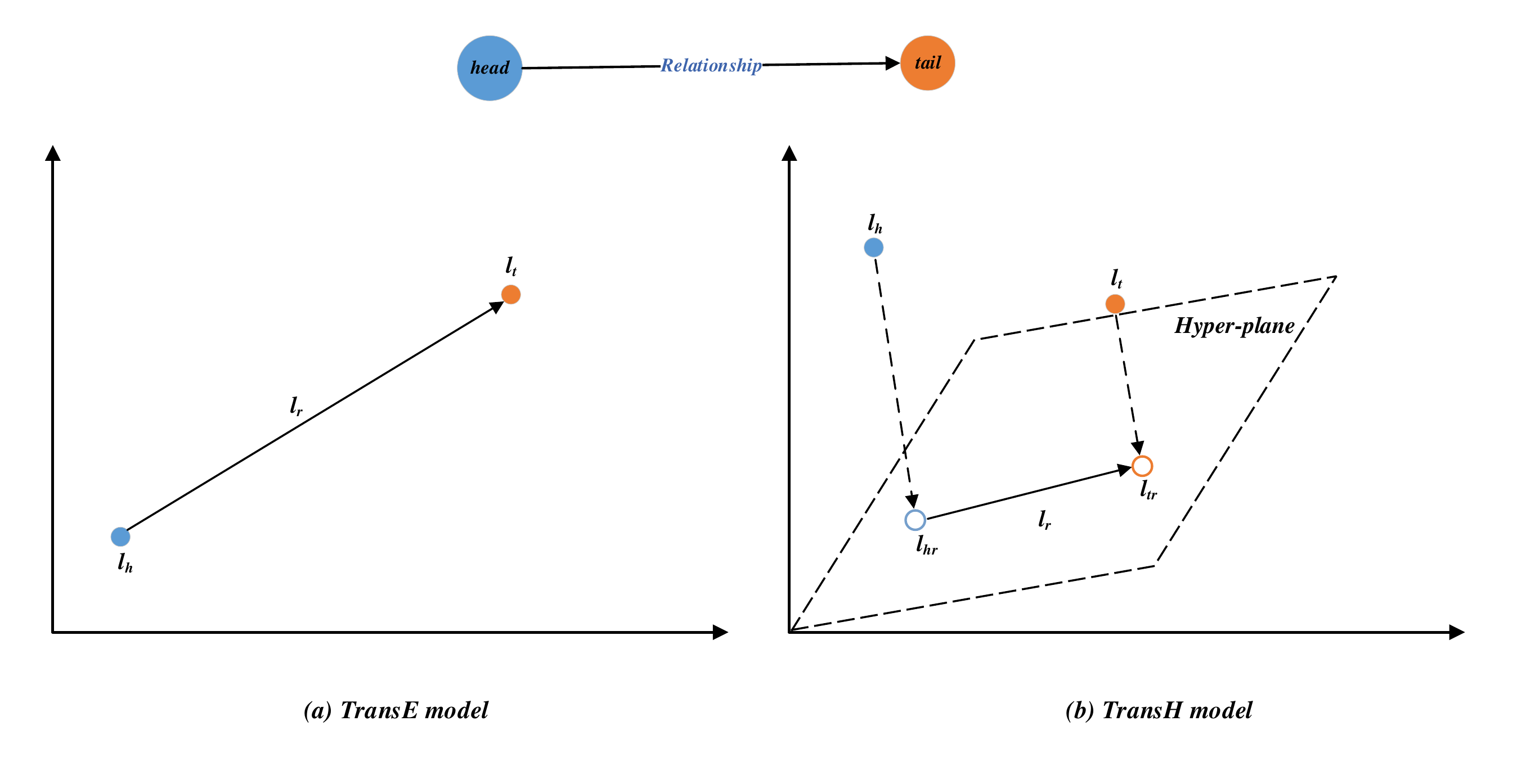}
    \caption{Translation embedding models.}
    \label{fig:Trans model}
\end{figure}

TransE model is the first and the classic algorithm of translation models for knowledge graph embedding. It regards the relationship in the knowledge graph as a translation vector between entities. For each triple like $(h,r,t)$, TransE regards $\boldsymbol{l}_r$, the vector representation of relationship $r$, as the translation between the head entities' vector $\boldsymbol{l}_h$ and tail entities' vector $\boldsymbol{l}_t$. Based on this idea, we can also regard $\boldsymbol{l}_t$ as the translation of $\boldsymbol{l}_h$ through relationship $\boldsymbol{l}_r$. As shown in Fig.~\ref{fig:Trans model}, for a triple $(h,r,t)$ which is short for $(head)-[relationship]\rightarrow(tail)$, the goal of TransE is to iteratively update the parameters of vector $\boldsymbol{l}_h,\boldsymbol{l}_r,\boldsymbol{l}_t$ as much as possible so that the formula $\boldsymbol{l}_h+\boldsymbol{l}_r \approx \boldsymbol{l}_t$ holds. The loss function of the TransE model is defined in Eq.~\eqref{eq:loss function}. In the geometric sense, $|\boldsymbol{l}_h+\boldsymbol{l}_r-\boldsymbol{l}_t|_{L_1/L_2}$ is the distance from the head entity of the triple $(h,r,t)$ to the tail entity through the translation of the relationship $r$. So, the result of Eq.~\eqref{eq:loss function} is also considered as the distance of the triple.

\begin{equation}
    f_r(h,t)=|\boldsymbol{l}_h+\boldsymbol{l}_r-\boldsymbol{l}_t|_{L_1/L_2}\label{eq:loss function}
\end{equation}

TransE model is the most concise form of the translation model, which has only one continuous space. In this space, entities and relationships have unique representations. This gains a small calculated cost, but its ability of semantic representation is also limited. 
Using the trip scenario as an example. The meanings of traffic zones vary among individuals such as a hospital located traffic zone may be the workplace for a doctor but a healing place for a patient. Hence, the model should allow traffic zone entities to have different representations when considering different individuals. However, this is not achievable for TransE model, making it hard for traffic zone entities to converge. 
To address the problem of complex semantic representation, some enhanced models were proposed and TransH model (\cite{wang2014knowledge}) is one of them. 
By introducing the hyper-plane affiliated to the relationship, TransH model enables the same entity to have different representations in triples composed of different relationships. As shown in Fig.~\ref{fig:Trans model}(b), for the relationship $r$, TransH model uses the translation vector $\boldsymbol{l}_r$ and the normal vector $\boldsymbol{w}_r$ of the hyper-plane to express it at the same time. As for a triple $(h,r,t)$, TransH model projects the head entity vector $\boldsymbol{l}_h$ and the tail entity vector $\boldsymbol{l}_t$ along the normal to the hyper-plane corresponding to the relationship $r$, by which $\boldsymbol{l}_{hr}$ and $\boldsymbol{l}_{ht}$ will be obtained and their calculation equation is shown in Eq.~(\ref{eq:transh_head},~\ref{eq:transh_tail}). In this way, the model is capable of handling complex semantics. For instance, under the schema of TKG, the same traffic zone entity could have different representations after mapping by the hyper-planes of the individual's private relationships. In other words, the representation of traffic zone entities changes when forming triples with different individuals.
In this case, the loss function(also the calculation of the triple distance) of TransH model is changed to Eq.~\eqref{eq:transH loss function}. 

\begin{equation}
    \boldsymbol{l}_{hr}=\boldsymbol{l}_h-\boldsymbol{w}_r^T\boldsymbol{l}_h\boldsymbol{w}_r \label{eq:transh_head}
\end{equation}

\begin{equation}
    \boldsymbol{l}_{tr}=\boldsymbol{l}_t-\boldsymbol{w}_r^T\boldsymbol{l}_t\boldsymbol{w}_r \label{eq:transh_tail}
\end{equation}

\begin{equation}
     f_r(h,t)=\left|\boldsymbol{l}_{hr}+\boldsymbol{l}_r-\boldsymbol{l}_{tr}\right|_{L_1/L_2}\label{eq:transH loss function}
\end{equation}

The negative sampling strategy is commonly adopted when training models to improve the efficiency of training and enhance distinguishing ability, such as \cite{wang2020attention}, especially for translation models. The triples constructed in the knowledge graph based on observed data are considered the correct triples or positive samples. Other triples are generally called false triples or negative samples. Unlike positive samples derived from historical data, negative samples need to be constructed artificially. The general method of generating the set of negative samples is to randomly replace one of the head entities (\cite{pan2008one}), relationship and tail entity of positive samples with other entities or relationships. Denote the negative samples generated by this method as $S^-$. Then it can be described by Eq.~\eqref{eq:neg_sample}.

\begin{equation}
    S^- = \left\{(h^{\prime},r,t)\right\}\cup \left\{(h,r^{\prime},t)\right\} \cup \left\{(h,r,t^{\prime})\right\} \label{eq:neg_sample}
\end{equation}

Most translation-based models typically adopt the negative sample strategy with their optimization objective function shown in Eq.~\eqref{eq:objective function} where $S$ is the set of positive samples, $S^-$ is the set of negative samples and $\gamma$ is the acceptable distance between the positive and negative samples. 

\begin{equation}
    \sum_{(h,r,t)\in S} \sum_{(h^{\prime},r^{\prime},t^{\prime})\in S^-} \text{max}\left(0,f_r(h,t)+\gamma-f_{r^{\prime}}\left(h^{\prime},t^{\prime}\right)\right)\label{eq:objective function}
\end{equation}

\subsubsection{Analysis and optimization of knowledge graph embedding for trip knowledge graph} \label{sec:ana_opt}

According to Eq.~\eqref{eq:transH loss function},~\eqref{eq:objective function}, the optimization objective of translation-based embedding model is for a single triple or a pair of triples (a positive sample and a negative sample). Nevertheless, overall optimization can be achieved when implement it on TKG. Overall optimization means that the optimization process proceeds in the direction of considering the decreasing distances of all triples, and eventually converges to the overall optimum, not a single triple. The following will explain the reason. 

Through Section~\ref{sec:data describe} we know that the association path is formed by multiple triples. Then we consider the following association path:

\begin{equation*}
    (h_m:Entity)-[r_m:relationship] \rightarrow (e_c:Entity)-[r_n:Relationship] \rightarrow (t_n:Entity)
\end{equation*}

which is formed by triple $f_m:(h_m,r_m,e_c)$ and $f_n:(e_c,r_n,t_n)$. If the vector representation of entities and relationship of $f_n$ have been adjusted during training, triple $f_m$ will be affected simultaneously because the tail entity $e_c$ of $f_m$ serves as the head entity of $f_n$. Likewise, the adjustment of the $f_m$ triple affects the triples that form associated path with it. That is, The training of a triple will affect all triples of the associated paths it forms. Through the introduction of Section \ref{sec:Structure} and Fig.~\ref{fig:Joint_KG}, almost all entities in TKG have association paths among them. Thus, the update of one triple will affect the parameterized representation of the others. So, although the optimization objective is for a single triple, benefiting from the schema of TKG, the training will converge towards the overall optimal. The interplay of expressions of different entities and relationships realizes information transfer, enabling training to learn associative information.

Overall optimization can be achieved no matter which translation-based model is chosen. In this paper, we adopt TransH since it can handle the complex relationships in TKG, and it is concise with fewer parameters. In addition, TransH adopts the semantics of trips well.
For example, a traffic zone generally contains multiple POIs, making individuals use them in different ways. It implies the meaning of traffic zones to individuals varies. So traffic zones should be represented differently in triples formed by different individuals, which can be handled by hyperplane of the relationship. 

General embedding models were proposed for generic knowledge graph, and we find it doesn't match well with TKG (a domain-specific knowledge graph) and our task.
Denote the triple of $(Zone)-[Has\_POI] \rightarrow (POI)$  type as $F^{poi}$ and others in Table.~\ref{tab:KG_triple} as $F^{trip}$. $F^{trip}$ is constructed based on trip data, which is usually of a larger quantity. Whereas only relatively small number of triples contained by $F^{poi}$. So the scale of $F^{poi}$ and $F^{trip}$ is imbalanced. The triple that $(Veh\_id)-[Choose\_D]\rightarrow(Zone)$ of $F^{trip}$ and $F^{poi}$ have a public type of entity $Zone$. If no adjustment is made to the training strategy, the representation of $Zone$ entities will be mainly dominated by the $(Veh\_id)-[Choose\_D]\rightarrow(Zone)$, leading to the related information between traffic zones and POI being harder to learn. This will reduce training efficiency significantly. To solve this problem, we can do a pre-train for $F^{poi}$, which can make a better initialization of $Zone$. On the other hand, it is also feasible to augment $F^{poi}$ so that it has a scale comparable to other type triples before training.

Second, we find negative sampling strategy does not apply to TKG for the potential destination prediction task. According to Eq.~\eqref{eq:objective function}, the distance between negative samples is enlarged while it is reduced for positive samples during training. It means the negative sample is considered the opposite of the positive sample, or it would be misleading for training. So the fact described by negative samples should be truly false. 
However, in TKG, triples describing individual historical trips are constructed by trip data collected under short-term observation.
It means there are a large number of triples absent from TKG as they have not been observed, not because they are false. 
Potential destination prediction is essentially predicting unobserved facts. If the true triples describing the unobserved fact are regarded as negative samples during training, it will greatly affect the model's performance.
For example, $(Veh\_{id}:v_i)-[Choose\_D\_i] \rightarrow (Zone:z_b)$ may be considered and generated as a negative sample if $v_i$ has not chosen $z_b$ as the destination in history. However, $z_b$ may be a potential destination of $v_i$, so it would be a disaster if the triple was trained as a negative sample.
Therefore, to adapt to our data and task, we eliminate the negative sampling strategy and modify the optimization objective to Eq.~\eqref{eq:objective function2}. The idea of it is describing only the observed facts and considering that when the distance of a positive sample is less than $\gamma$ that setting for preventing over-fitting, then the fact is considered to be well expressed by the model without adjustment. 

\begin{equation}
    \sum_{(h,r,t)\in S} \text{max}\left(0,f_r(h,t)-\gamma\right)\label{eq:objective function2}
\end{equation}

\subsection{Potential destination discovery}\label{sec:po_pre}

Trip knowledge graph embedding model (TKGEM) will be obtained by implementing the modified knowledge graph embedding algorithm (see Section~\ref{sec:ana_opt}) on TKG. This section will introduce the potential destination discovery (or prediction) based on TKGEM. 

\begin{figure}[ht]
    \centering
    \setlength{\abovecaptionskip}{0.cm}
    \includegraphics[scale=0.58]{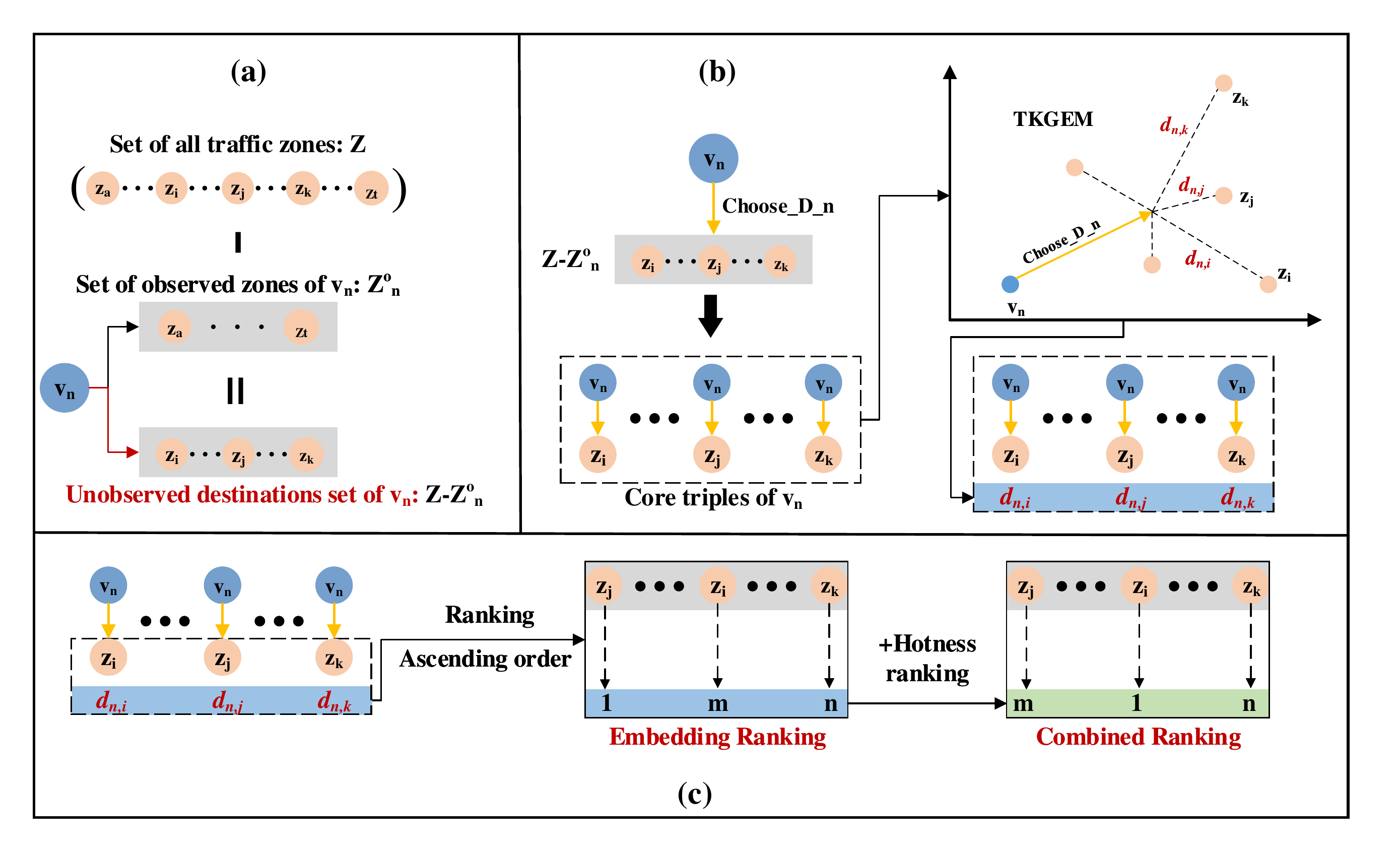}
    \caption{Framework of potential destination discovery of the individual.}
    \label{fig:Infer_framework}
\end{figure}

According to Section \ref{sec:embedding1}, the training is actually a process of decreasing the distance of positive samples. That is, TKGEM portrays the possibility that the fact is established by the distance of the triple that describes it. When the model converges, drawing on the knowledge graph completion task, we argue that the possibility of a fact is negatively correlated with its triple distance. On this basis, TKGEM makes individual-level discovery or prediction, and Fig.~\ref{fig:Infer_framework} shows the flow for the individual.

Following Fig.~\ref{fig:Infer_framework}, we introduce the process of the individual potential destination discovery by taking $v_n$ as an example. First, we identify the unobserved destination set $Z-Z^o_n$ of $v_n$ as its potential destination candidate set, each of which could be its potential destination. Then for each $z_j \in Z-Z^o_n$, a core triple of $v_n$ can be constructed, i.e., $(Veh\_id:v_n)-[Choose\_D\_n] \rightarrow (Zone:z_j)$, and its distance can be calculated by TKGEM, denoted as $d_{n,j}$. According to the basic idea of TKGEM, $d_{n,j}$ can be considered as portraying the possibility that $v_n$ will choose $z_j$ as the destination in the future. 
Although the distances of core triples cannot be mapped to quantitative possibilities, the ranking of the possibilities can be derived from the relative scale of the distances. For example, it can be concluded that $z_j$ has a higher probability of being a potential destination if $d_{n,i}>d_{n,j}$, even though we do not know exactly how large a possibility $d_{n,i}$ or $d_{n,j}$ corresponds to. On this basis, we can get the possibility ranking of each $z_j \in Z-Z^o_n$ being a potential destination for $v_n$. The above process is summarized in Algorithm \ref{alg:algorithm0}.

\begin{algorithm}[H]
 \caption{Algorithm of potential destination discovery for the individual.}
 \label{alg:algorithm0}
 \KwIn{Vehicle individual: $v_n$; TKGEM: $\textbf{\textit{M}}$; Zone to be ranked: $z_j \in Z-Z_n^o$}
 \KwOut{The possible ranking of $z_j$ for $v_n$ given by $\textbf{\textit{M}}$: $a^e_{n,j}$}
 $Z^u_n\leftarrow Z-Z_n^o$\;
 $h_n\leftarrow Entity \ (Veh\_id:v_n)$\;
 $r_n \leftarrow Relationship \ [Choose\_D\_n]$\;
 $S_n \leftarrow \varnothing$\;
 \For {$z_m$ in $Z^u_n$}{
 $t_m \leftarrow Entity \ (Zone:z_m)$\;
 $c_{n,m} \leftarrow Triple \ (h_n,r_n,t_m)$\;
 Calculate the distance of $c_{n,m}$ by $\textbf{\textit{M}}$, denoted as $d_{n,m}$\;
 $S_n \leftarrow d_{n,m}$\;
 }
 $a^e_{n,j} \leftarrow 1$\;
 \For{$d_{n,l}$ in $S_n$}{
 \If{$d_{n,l} < d_{n,j}$}{
 $a^e_{n,j}\leftarrow a^e_{n,j}+1$\;}
 }
 \textbf{Return} $a^e_{n,j}$
\end{algorithm}

Denote the traffic zone's ranking obtained by Algorithm \ref{alg:algorithm0} as its embedding ranking (PDPFKG-ER). According to the data modeling and training algorithm of TKG and TKGEM of PDPFKG, the embedding ranking is supposed to be given based on the association information. To confirm it empirically, as well as reflect the differences between PDPFKG and statistical models (e.g., deep learning models), we introduce a statistics-based ranking: traffic zone's hotness ranking (HR), which is ranked by the frequency of visits by all travelers, i.e., the hotness ranking of the most visited traffic zone is $1$. Further, the combined ranking (PDPFKG-CR) of embedding ranking and hotness ranking can be calculated by Algorithm \ref{alg:algorithm1}. All of these rankings will be demonstrated, analyzed, and discussed in the experimental and discussion section.

\begin{algorithm}[H]
 \caption{Algorithm of calculating traffic zone's combined ranking.}
 \label{alg:algorithm1}
 \KwIn{Vehicle individual: $v_n$; Zone to be ranked: $z_j \in Z-Z_n^o$}
 \KwOut{The combined ranking of $z_j$ for $v_n$: $a^c_{n,j}$}
 $Z^u_n\leftarrow Z-Z_n^o$\;
 $L_n \leftarrow \varnothing$\;
 \For{$z_m$ in $Z_n^u$}{
 $a^e_{n,m}$ $\leftarrow$ Algorithm \ref{alg:algorithm0}\;
 Get the hotness ranking of $z_m$, denoted as $a^h_{m}$\;
 $a^l_{n,j}$ $\leftarrow$ $a^e_{n,m} + a^h_{m}$\;
 \If{$a^l_{n,j}$ not in $L_n$}{
 $L_n \leftarrow a^l_{n,j}$\;
 \Else{$L_n \leftarrow a^l_{n,j}+1$\;}
 }
 }
 $a^c_{n,j} \leftarrow 1$
 \For{$a^l_{n,m}$ in $L_n$}{
 \If{$a^l_{n,m} < a^l_{n,j}$}{
 $a^c_{n,j}\leftarrow a^c_{n,j}+1$\;}
 }
\textbf{Return} $a^c_{n,j}$
\end{algorithm}

\section{Experiments} \label{sec:exper}

\subsection{Dataset description}\label{sec:data describe}

In this section, we performed the proposed method with a real-world vehicular dataset (without public vehicles) from Xuancheng, China.

\subsubsection{Data preparation}\label{sec:data scene}

\begin{figure}[ht]
    \centering
    \includegraphics[scale=0.70]{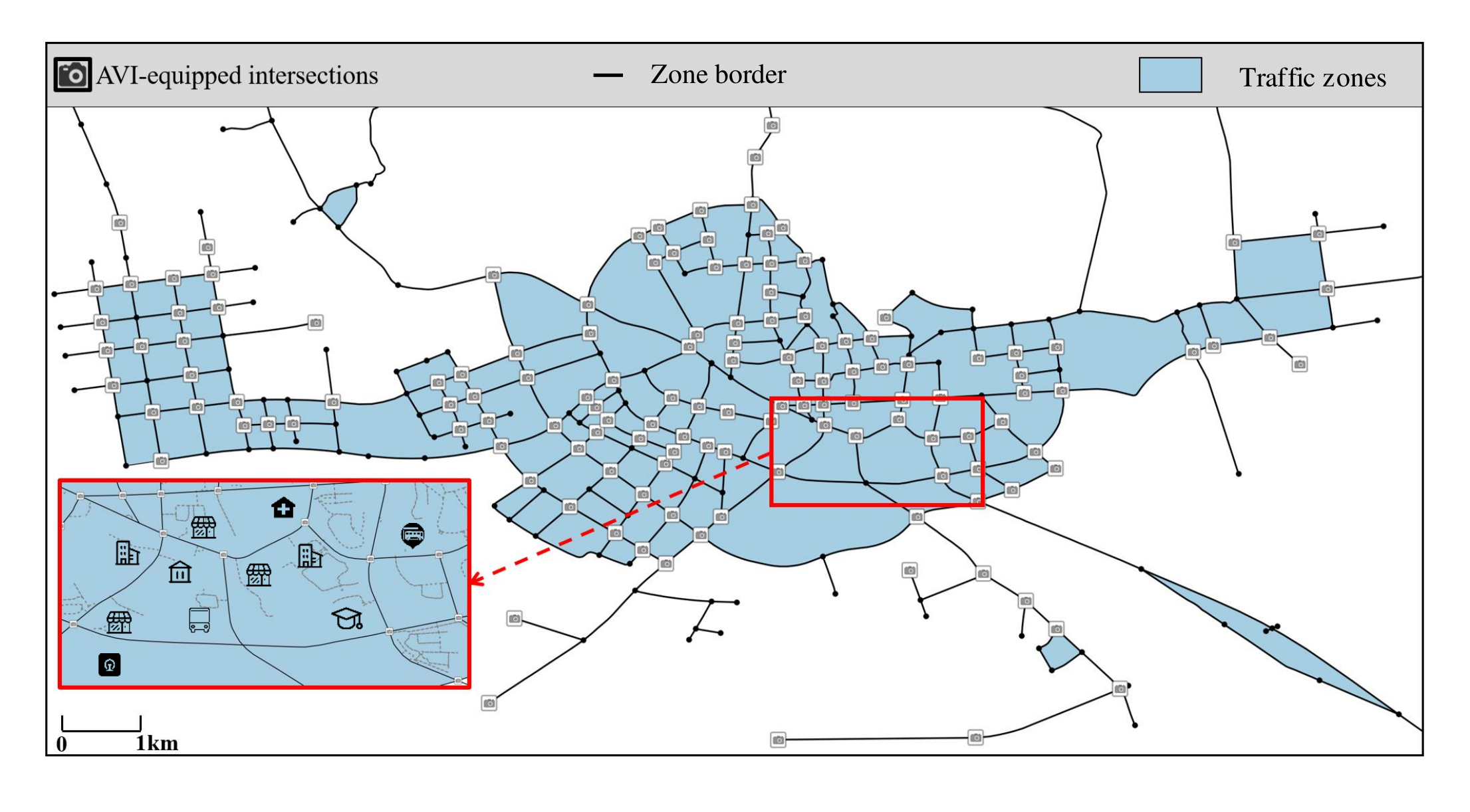}
    \caption{Road network and traffic zones of Xuancheng city.}
    \label{fig:data_des_zone}
\end{figure}

The original trip data was collated by the automatic vehicle identification systems deployed in the city road network. It records vehicles' activity passively on road network with their identity. 
The road network and AVI systems distribution in Xuancheng are shown in Fig.~\ref{fig:data_des_zone}, and the fields of the original data are shown in Table~\ref{tab: original_data_fields}. The traffic zone proposed in \cite{wang2021city} is used to describe the origin and destination of trips with the total number of $191$. There are multiple points of interest (POIs) inside of the traffic zones, and this information is publicly available on the Internet. 

\begin{table}[ht]
\centering
\caption{The fields of original trip data.}
\label{tab: original_data_fields}
\begin{tabular}{l|l}
\hline
       Field               & Description  \\
                      \hline
Vehicle\_id  & The id of vehicle, which is the identification of the vehicle.    \\
Date         & The date of the trip.        \\
Ftime        & The departure time of the trip.       \\
Fzone        & The origin of the trip. \\
Tzone        & The destination of the trip.     \\
\hline
\end{tabular}
\end{table}

The scale of original data covers five weeks from August 5th, 2019 to September 8th, 2019. Only the first-week data was extracted and assumed available under our experimental context to simulate a short-term observation. The other four weeks' data are used for validation and discussion.

There are some individuals who have strong regularities even under the observation of one week. To eliminate their interference with model evaluation, we filtered out them to ensure our scenario consisted of low predictability individuals totally. The filtering principles take into account trip frequency, destination distribution, as well as spatial and temporal correlation of trips (\cite{li2022urban}). For instance, individuals who have few recurrent destinations, indicating a poor spatial correlation, are more likely to be identified as low predictability individuals. As a result, $95,509$ individuals are selected and formed the target group for research, which accounts for $81.56\%$ of the total.

\subsubsection{Data analysis}

\begin{figure}[ht]
    \centering
    \includegraphics[width=0.92\textwidth]{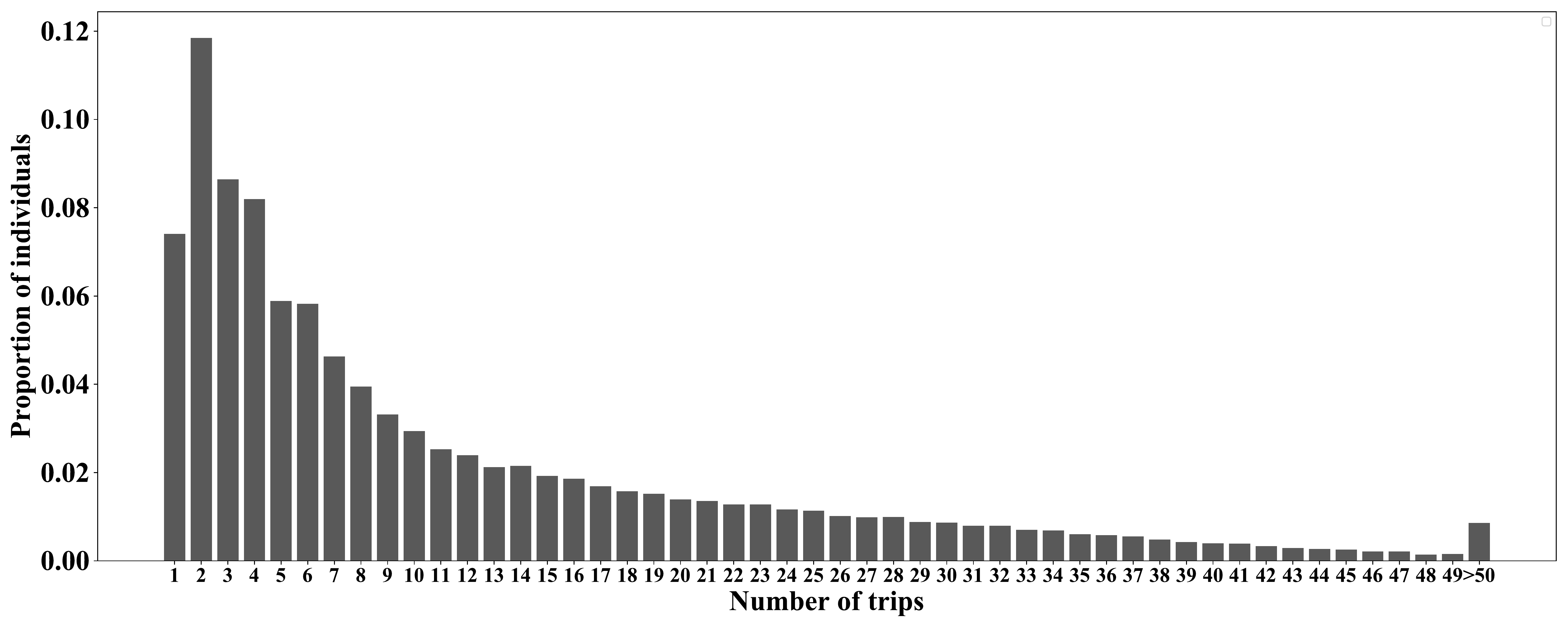}
    \caption{Trip frequency distribution of target individuals in one week.}
    \label{fig:trip_frc}
\end{figure}

To demonstrate the low predictability of our target individuals in detail, we analyzed their trip data from three perspectives. First, the trip frequency distribution of these individuals is shown in Fig.~\ref{fig:trip_frc}. It shows that the majority of individuals had less than $10$ trips observed. For more than $20\%$ of individuals, no more than two trips were observed, indicating the amount of individual data is extremely limited.

\begin{figure}[h]
\setlength{\abovecaptionskip}{0.cm}
\centering
\includegraphics[width=0.57\textwidth]{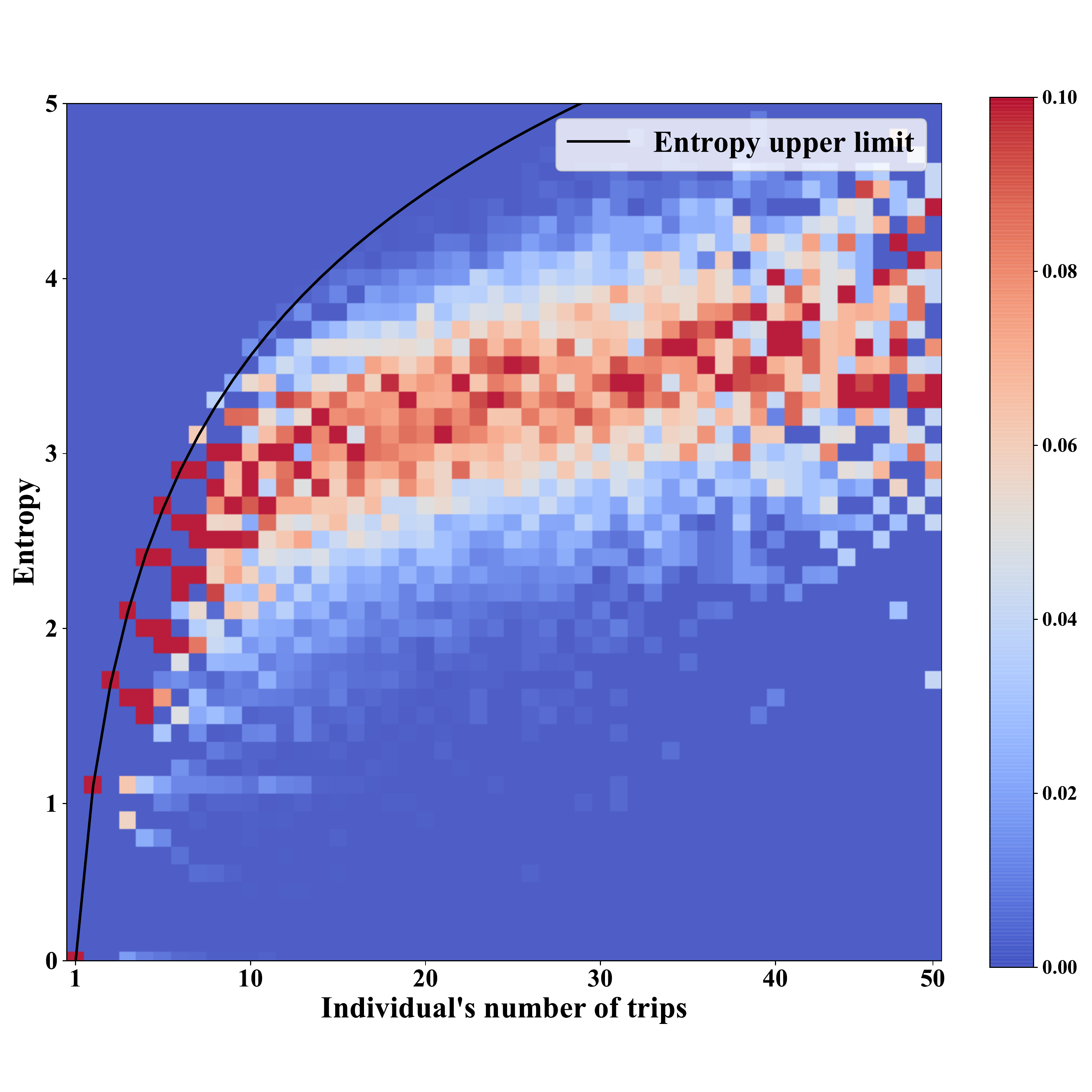}
\caption{Entropy distribution of target individuals.}
\label{fig:shang_sj}
\end{figure}

Entropy is a widely accepted metric to measure individual's trip regularity, such as entropy (\cite{scheiner2014gendered}), actual entropy (\cite{song2010limits}), and entropy rate (\cite{goulet2017measuring}). In this paper, the entropy of individual $v_n$ is defined as Equation.~\ref{eq:entropy}, where $p_n(z_j)$ is the historical probability that $z_j$ was chosen as the destination by $v_n$. The entropy distribution of target individuals is shown in Fig.~\ref{fig:shang_sj}, and the color blocks represent the proportion of individuals among all individuals with the same number of trips. (the entropy distribution of individuals we excluded is shown in Appendix.~\ref{ap:app}). It should be noted that the dynamic range of colors is set to $0-0.1$, and the scale higher than $0.1$ is also marked with the same color as $0.1$ (red), for better display of details. The black curve indicates the upper limit of the entropy under the corresponding number of trips. The high entropy of the target individuals shown in Fig.~\ref{fig:shang_sj} demonstrates these individuals are low predictability.  

\begin{equation}
    \textit{E}_n = \sum_{z_j \in Z^o_n}p_n(z_j)log_2p_n(z_j) \label{eq:entropy}
\end{equation}

Limited observations and randomness of trips lead to a large number of potential destinations for the target individuals. In addition to the potential destination, we define the accidental destination of $v_n$ as locations that were chosen as the destination during $T^o$ but not visited in $T^f$. Taking the case shown in Fig.~\ref{fig:difi_des} as an example, $z_c$ is an accidental destination of the traveler under the setting that $T^o=(d_0,d_s)$ and $T^f=(d_s,d_l)$. By denoting the percentage of accidental destinations and potential destinations of $v_n$ as $q^a_n$ and $q^p_n$ respectively, then they can be calculated by Eq.~(\ref{(1)},~\ref{(2)}). On this basis, we calculate the target individual's average of $q^a_n$ and $q^p_n$ with various lengths of $T^f$ and a fixed $T^o$ (7 days, i.e., the first week). The results are shown in Table~\ref{tab:percentage of accidental destination and potential destination}.
Besides, we demonstrate the distribution of the $q^a_n$ and $q^p_n$ of target individuals with $T^f = 14~\text{days}$ in detail by Fig.~\ref{fig:acc_po} (Appendix.~\ref{ap:app}).

\begin{equation}
q^a_n=\frac{|Z^o_n-Z^f_n|}{|Z^o_n|}*100\% \label{(1)}
\end{equation}
\begin{equation}
q^p_n=\frac{|Z^f_n-Z^o_n|}{|Z^f_n|}*100\% \label{(2)}  
\end{equation}

\begin{table}[ht]
\centering
\caption{The percentage of accidental destinations and potential destinations of target individuals.}
\label{tab:percentage of accidental destination and potential destination}
\begin{tabular}{lcc}
\hline
                      & Percentage of  accidental destinations & Percentage of potential destinations \\
                      \hline
$T^o$=7 days \& $T^f$=7 days         & 62.63\%              & 58.18\%             \\
$T^o$=7 days \& $T^f$=14 days        & 52.00\%              & 64.96\%             \\
$T^o$=7 days \& $T^f$=21 days        & 45.67\%              & 68.69\%             \\
$T^o$=7 days \& $T^f$=28 days        & 39.63\%              & 72.40\%             \\
\hline
\end{tabular}
\end{table}

According to Table~\ref{tab:percentage of accidental destination and potential destination}, the target individuals have a large proportion of both accidental destinations and potential destinations, indicating they still show strong randomness in destination choice after our observation period (i.e., the first week). In addition, the high percentage of potential destinations confirms individuals' observation is insufficient in our scenario. Further, it shows that potential destination prediction urgently needs research under limited observation.

\subsection{Experimental setting}
\subsubsection{Training and validation data splitting}

The low predictability individuals we focused on are those target individuals screened in Section \ref{sec:data scene}. The first-week original data (August 5th, 2019 to August 12th, 2019) is used for constructing the trip knowledge graph and training, containing $1,047,061$ trip records, and the remaining is for validation. In detail, the following two weeks (August 13th, 2019 to August 27th, 2019) are used for evaluation in Section \ref{sec:ex_eva}, and the last two weeks are applied in Section \ref{sec:diff_period} where we discuss the performance of PDPFKG over different future time horizons.

The constructed trip knowledge graph has $95,728$ unique entities and $381,513$ relationships. The types of entities are shown in Table~\ref{tab:KG_entity}. When extracting trip time entities, \textit{Ftime} in Table~\ref{tab: original_data_fields} were mapped as time spans, with a total of seven such as the morning peak. The schema of TKG is shown in Fig.~\ref{fig:KG_structure}, and its triples are shown in Table~\ref{tab:KG_triple}.

\subsubsection{Experiment setup} \label{sec:para}

We implemented the modified TransH model to TKG based on Pytorch framework.
The crucial parameters are as follows.

\begin{table}[ht]
\centering
\caption{Experimental parameters for TransH model.}
\label{tab:training setting}
\begin{tabular}{l|c}
\hline
\textbf{Parameter}      & \multicolumn{1}{c}{\textbf{Value}}      \\ 
\hline

Entity Dimension & 148 \\
Relationship Dimension & 148   \\
$\gamma$ & 1.0  \\
Learning Rate  & 0.003   \\
Training Batch Size & 1024  \\
Optimizer   & Adam (\cite{kingma2014adam})  \\ \hline

\end{tabular}
\end{table}

\subsubsection{Result refinements}

\begin{figure}[H]
\setlength{\abovecaptionskip}{0.cm}
\centering
\includegraphics[width=0.98\textwidth]{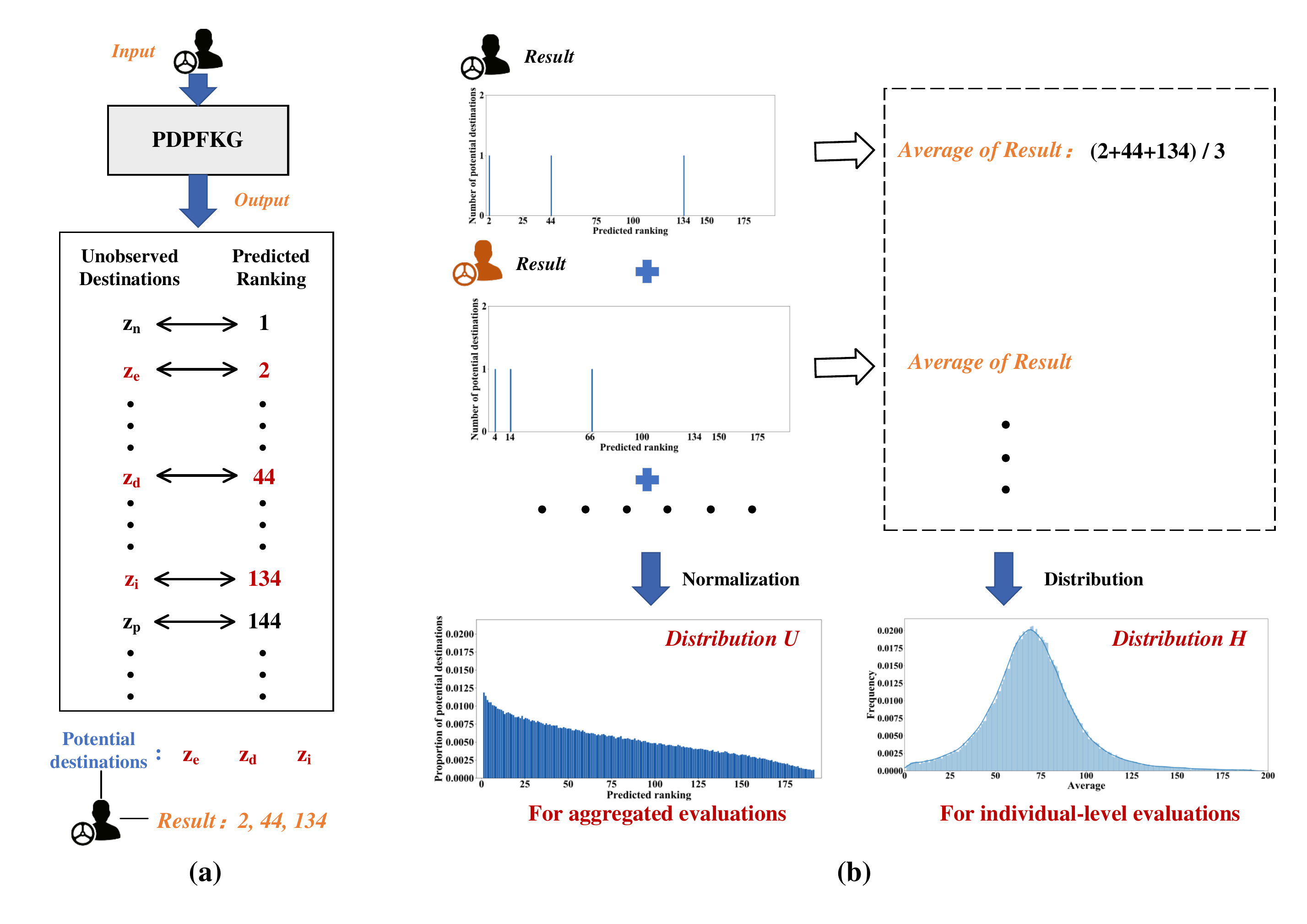}
\caption{Schematic illustration of result refinements.}
\label{fig:refinement}
\end{figure}

For $v_n$, its unobserved and potential destination sets can be presented as $Z^u_n=Z-Z^o_n$ and $Z^p_n=Z^f_n-Z^o_n$, respectively. As shown in Fig.~\ref{fig:refinement}(a), for each $z_j \in Z^u_n$, PDPFKG will give a predicted ranking of it, denote as $a^e_{n,j}$ (see Algorithm  \ref{alg:algorithm0}). In this case, $Z^p_n$ would be mapped as a set of ranking (e.g., $\{z_c,z_d,z_i\}\rightarrow \{2,44,134\}$ in Fig.~\ref{fig:refinement}(a)), denote as $R_n$. $R_n$ is considered the discovery or prediction result of $v_n$, and each element represents the predicted ranking of a specific potential destination of $v_n$.

$R_n$ varies among individuals (e.g., the black individual ($v_n$) and the orange one in Fig.~\ref{fig:refinement}(b)), while we can not judge which one should prevail. Thus, result refinements are needed to evaluate experimental results. As shown in Fig.~\ref{fig:refinement}(b), on the one hand, we aggregate the results of all individuals and normalize it as a discrete distribution, denoted as $U$. $p^U(i)$ indicates the proportion of potential destinations predicted by ranking $i$. On the other hand, the average of each individual's result is calculated for individual-level evaluation. Its distribution, denoted as $H$, reveals the performance varies among individuals.

\subsection{Evaluation metrics} \label{sec:metrics}

The distribution $U$ and $H$ are two objects we focus on, of which $U$ is the key for performance evaluation, and the differences in individual predictions can be gotten from $H$. Predicted ranking $i$ should have a negative correlation with $p^U(i)$ for an available $U$. Hence, we adopt spearman's rank correlation coefficient (spearman's $\rho$) to evaluate the overall relevance of $p^U(i)$ with $i$ of $U$. For a more refined and comprehensive evaluation of $U$, we further introduce the following metrics. 

\textbf{Confusion degree.} Denotes the value ranking of $p^U(i)$ in descending order among $U$ as $i'$, i.e., $\ p^U(i)<p^U(j) \Rightarrow i'>j'$. Then the degree of ranking confusion of $U$ can be calculated by Eq.~\eqref{eq:confusion degree1}. 

\begin{equation}
    D_{f} =\sum_{i \in U}\left|i' - i\right|  \label{eq:confusion degree1}
\end{equation}

\textbf{Concentration degree (Recall).} The concentration degree of the Top-$k$ rankings can be calculated by Eq.~\eqref{eq:concentrated}, which can also be interpreted as recall$@$k.

\begin{equation}
    D_c(k) = \frac{\sum_{i=1}^{k}p^U(i)}{\sum_{i \in U}p^U(i)} \label{eq:concentrated}
\end{equation}

$D_f$ is introduced for evaluating the correctness of prediction. $i=i', \ \forall i\in U$ is considered perfectly correct since it guarantees that $z_i$ is more likely be chosen than $z_j$ by $v_n$ statistically when the predicted ranking of $z_i$ is prioritized over $z_j$ ($a_{n,i}<a_{n,j}$). $D_c(k)$ measures the method's capability. It only focuses on the concentration of the distribution in the head while ignoring the shape and variation. $D_c(k)$ can be translated to recall$@$k by taking all individuals as a whole (not the average of each individual's recall). For example, $D_c(1)$ is equal to the recall of Top-1 (Recall$@$1).

For distribution $H$, we mainly focus on its shape, mean, and deviation to explore the performance at the individual level and the inter-individual variation.

\subsection{Baselines}

The following three categories of methods are chosen for comparison. They have the same input ($v_n$) and prediction logic as our method and can achieve potential destination prediction personality.

\begin{itemize}
    \item \textbf{Random choice (RC):} This method simulates predicting in the absence of context information. The method randomly gives the ranking.
    \item \textbf{Matrix decomposition (MD):} A typical class of methods for data imputation. individuals and traffic zones are use as the two dimensions of the matrix. The matrix is initialized by filling the number of trips of the individual visiting the corresponding traffic zone. For each individual's vector, the traffic zone corresponding to $0$ value constitutes its set of unobserved destinations, and the $0$ values will be filled by performing matrix decomposition. On this basis, the rankings of unobserved destinations can be given by comparing the values after imputation. We have implemented three common matrix decomposition methods, which are UV decomposition (MD-UV), QR decomposition (MD-QR), and SVD decomposition (MD-SVD).
    \item \textbf{Collaborative filtering (CF):} A classic method for recommendation system. In our context, individual and traffic zone as considered as user and item respectively. The frequency of individuals choosing a traffic zone as the destination is regarded as the user's score for the item. On this basis, we implemented two methods, collaborative filtering user-based (CF-U) and collaborative filtering item-based (CF-I). 
\end{itemize}

\subsection{Experimental results} \label{sec:perfor}

\subsubsection{Overall aggregated performance visualization} \label{sec:over_per_vis}

\begin{figure}[H]
\centering
\subfigure[Hotness ranking(HR)]{\includegraphics[width=0.45\textwidth]{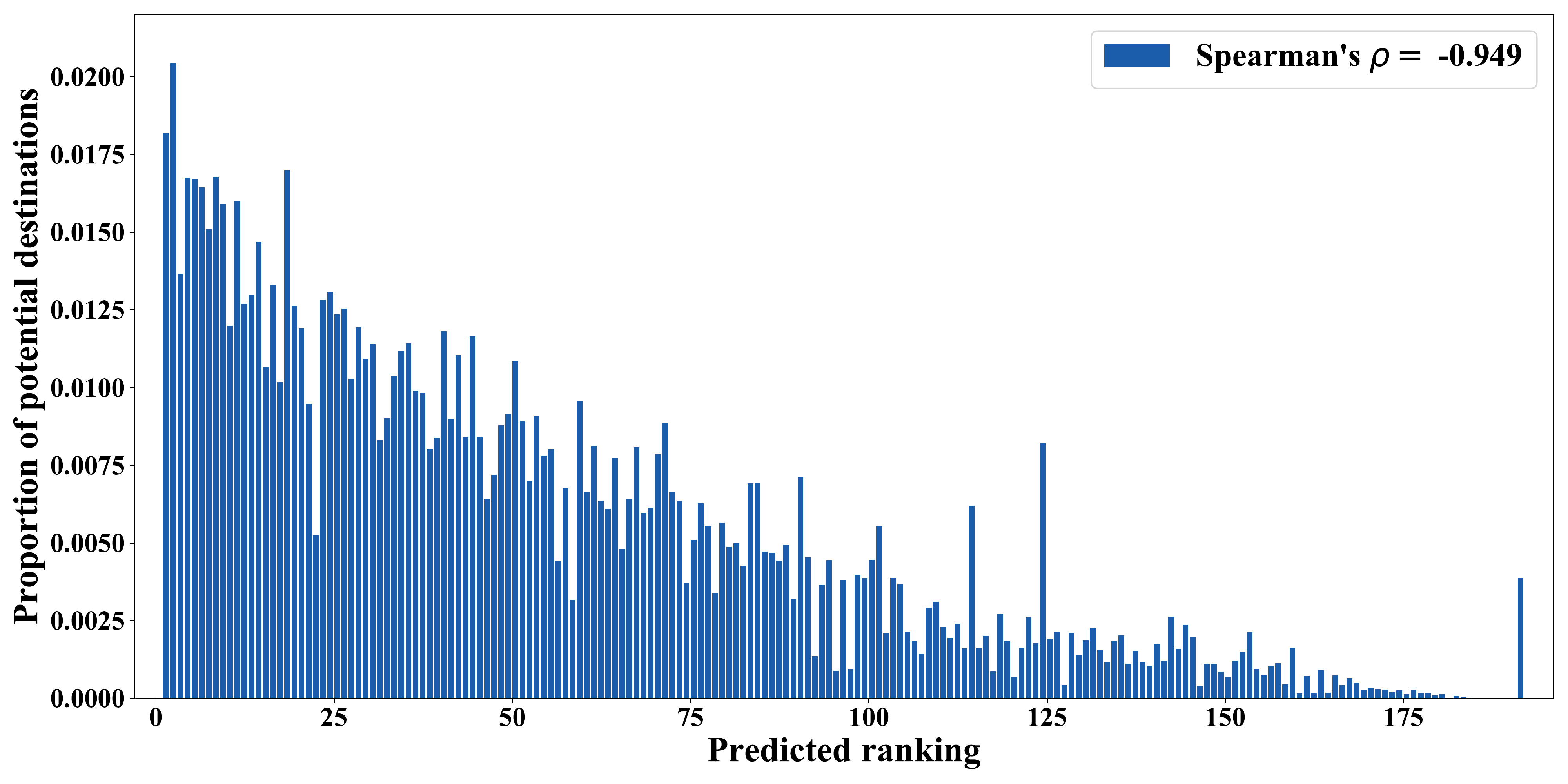}}
\subfigure[Embedding ranking(PDPFKG-ER)]{\includegraphics[width=0.45\textwidth]{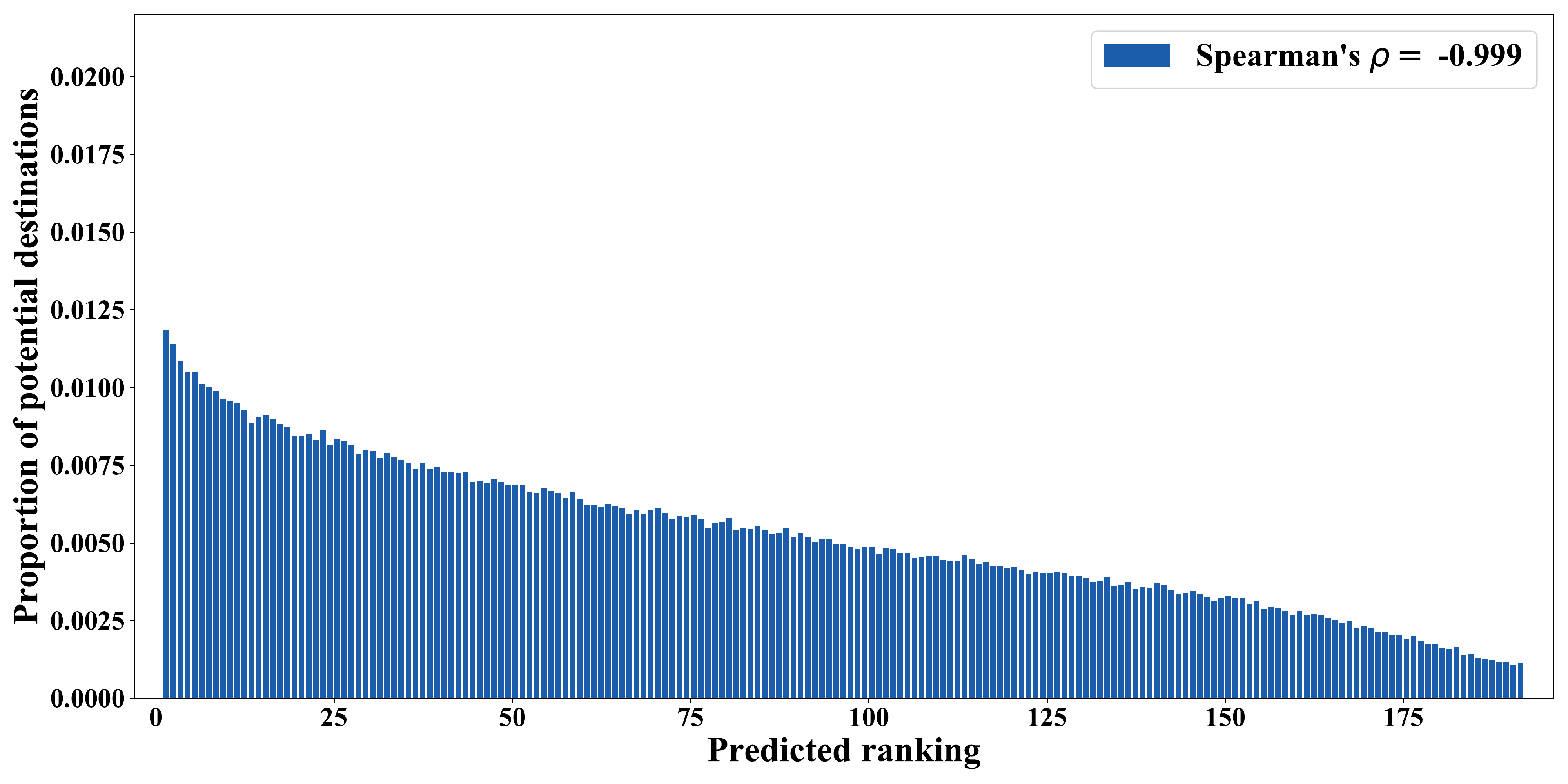}}
\subfigure[Combined ranking(PDPFKG-CR)]{\includegraphics[width=0.50\textwidth]{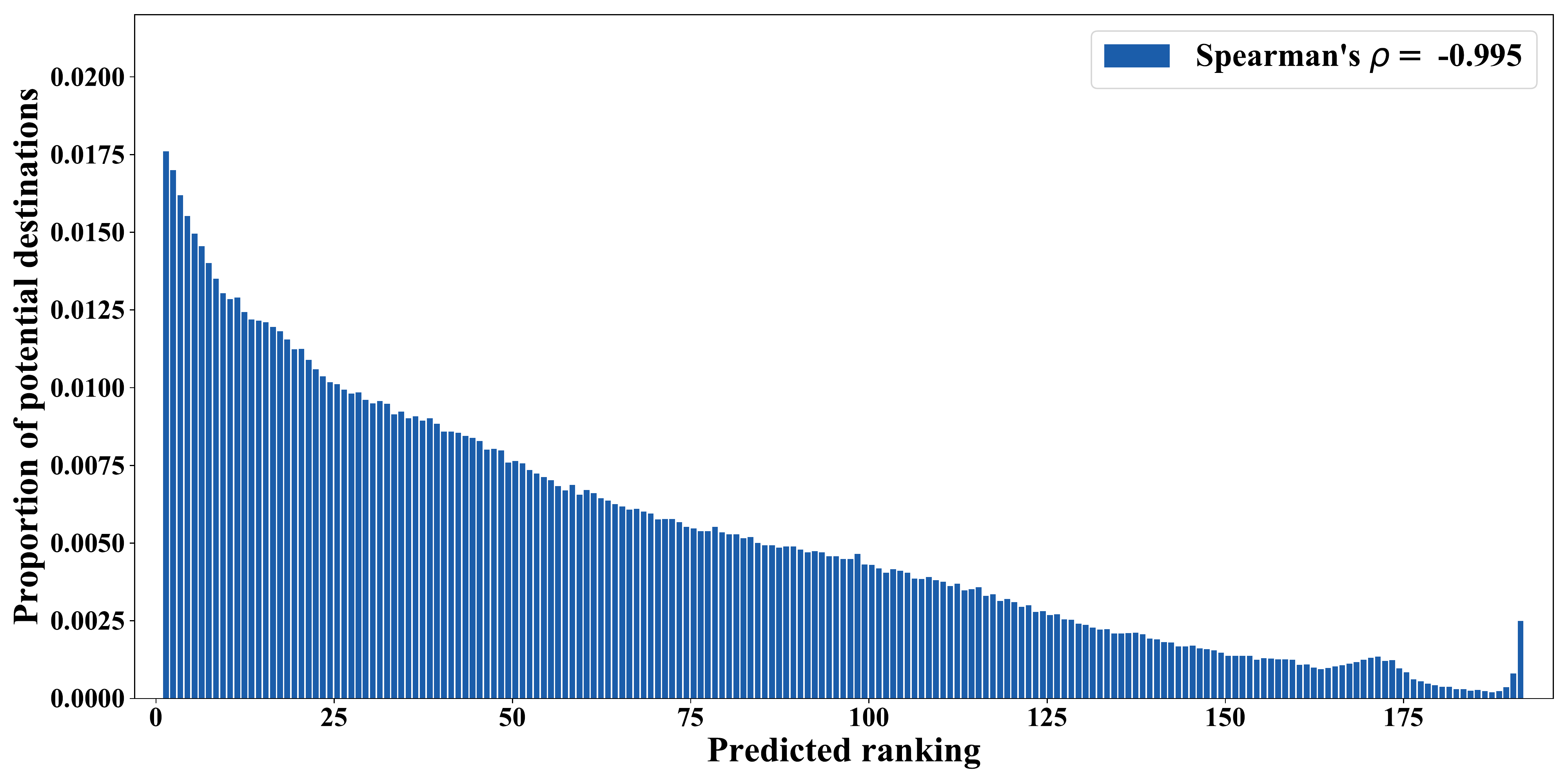}}
\caption{Visualization of $U$ (Performance) of different ranking ways.}
\label{fig:our method}
\end{figure}

Before quantitative evaluation, visualization of $U$ can intuitively show the overall performance. Fig.~\ref{fig:our method} shows $U$ refined by hotness ranking ($a^h$), embedding ranking ($a^e$) and combined ranking ($a^c$) mentioned in Section \ref{sec:po_pre}. The spearman's $\rho$ of them is shown as label.

The distribution shapes and the spearman's $\rho$ indicate that the predicted ranking strongly correlates with the quantity or proportion of potential destinations caught by it, especially for PDPFKG (ER and CR). It means their discovery or prediction results are valid overall. It should be emphasized that the performance of HR depends entirely on the context of the city and cannot achieve a personalized prediction. In other words, HR's performance in XuanCheng benefits from the significant difference and relative stability of the visiting hotness among different traffic zones in this city.
  
The distribution $U$ of PDPFKG-CR, which is refined by combining hotness ranking and embedding ranking (see Algorithm \ref{alg:algorithm1}), differs from that of HR and PDPFKG-ER. This variation demonstrates that the rankings given by hotness ranking and embedding ranking have a large difference. Further, it proves that the information learned by PDPFKG is different from the traffic zone's hotness information that can be obtained by simple statistics. On the other hand, combining ranking seems to be a feasible way for PDPFKG to integrate the statistical information according to the performance of PDPFKG-CR, which takes advantage of HR and PDPFKG-ER.

\begin{figure}[ht]
\centering

\subfigure[RC]{\includegraphics[width=0.45\textwidth]{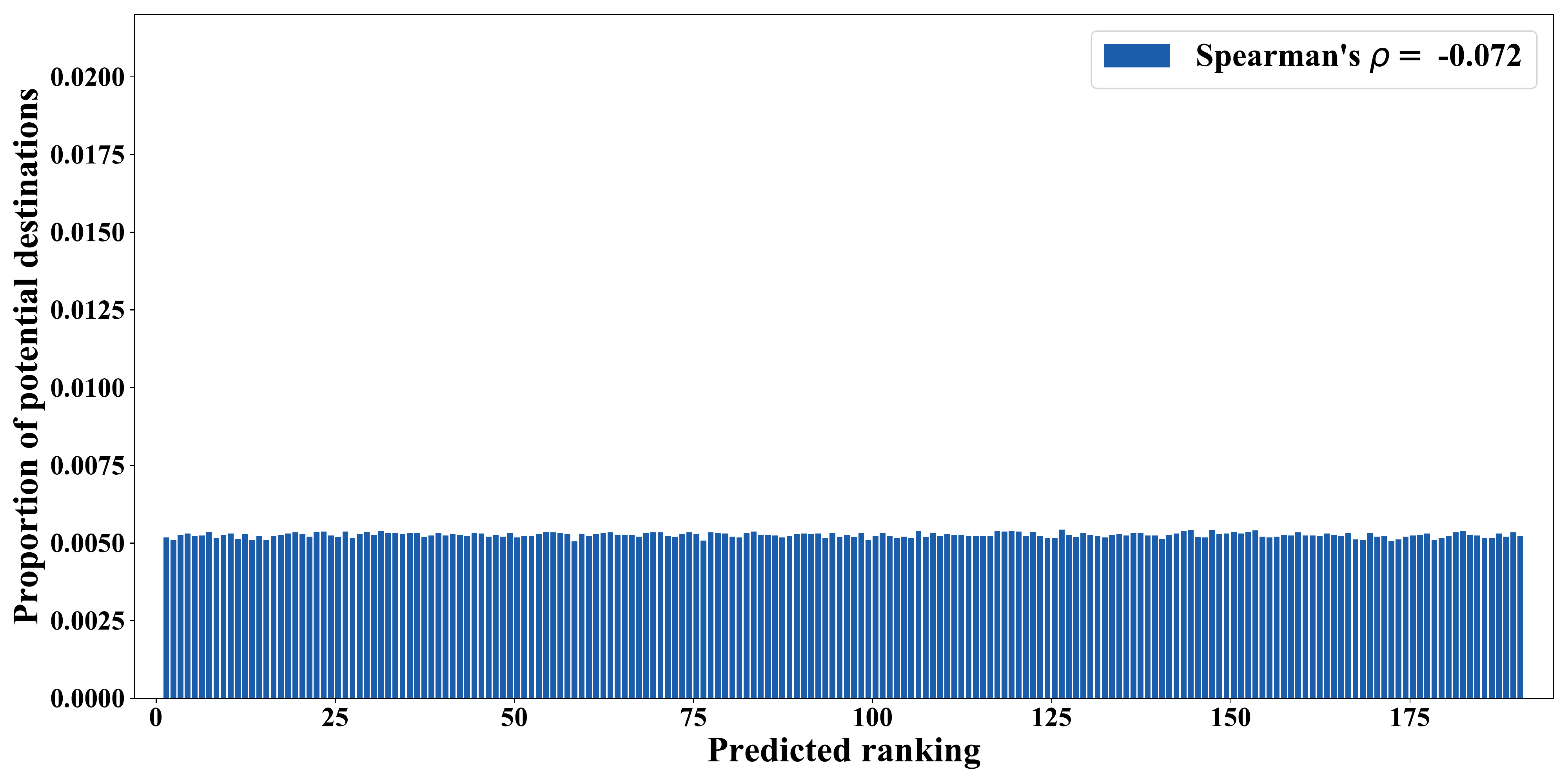}}
\subfigure[MD-UV]{\includegraphics[width=0.45\textwidth]{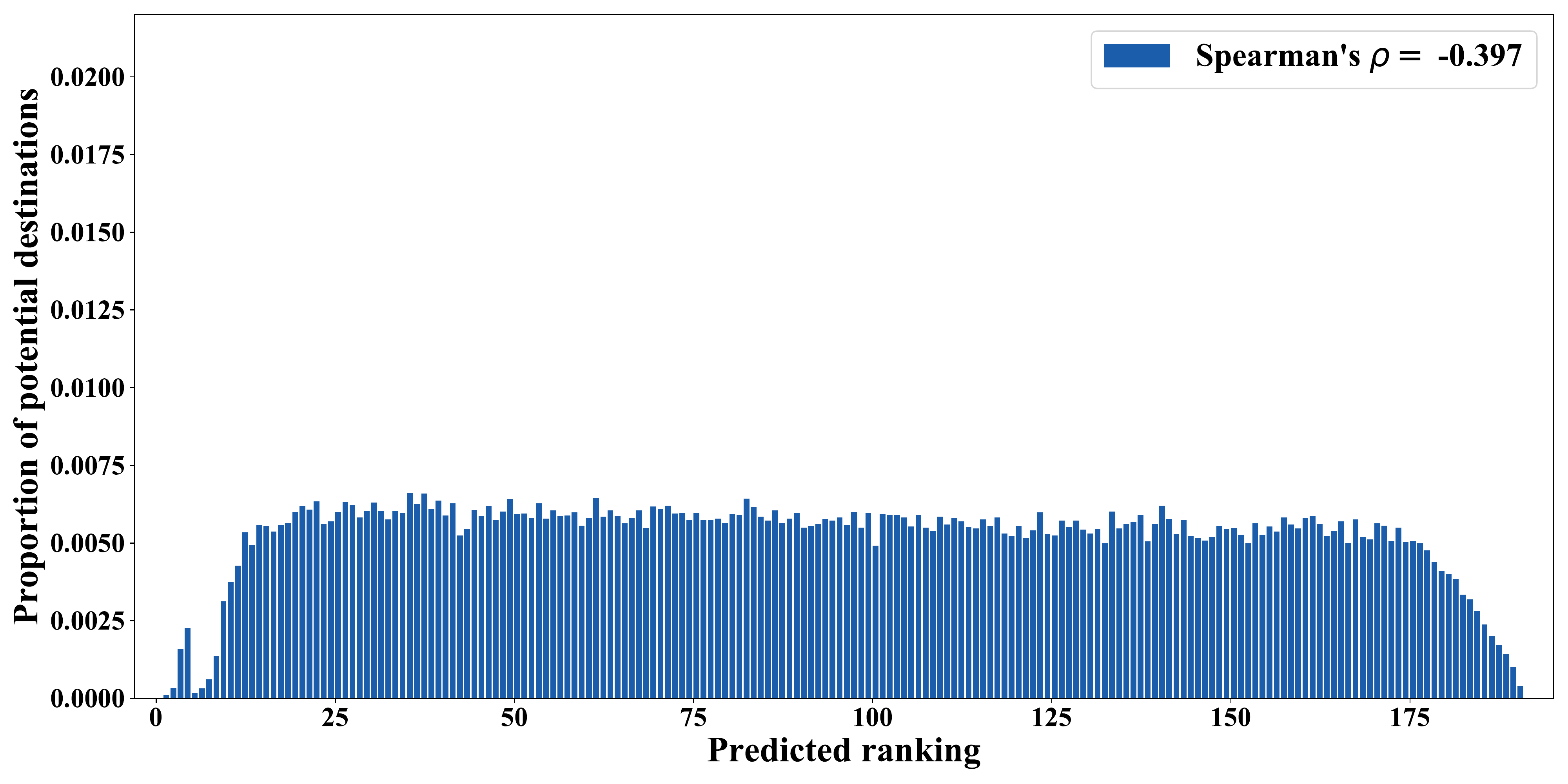}}
\subfigure[MD-QR]{\includegraphics[width=0.45\textwidth]{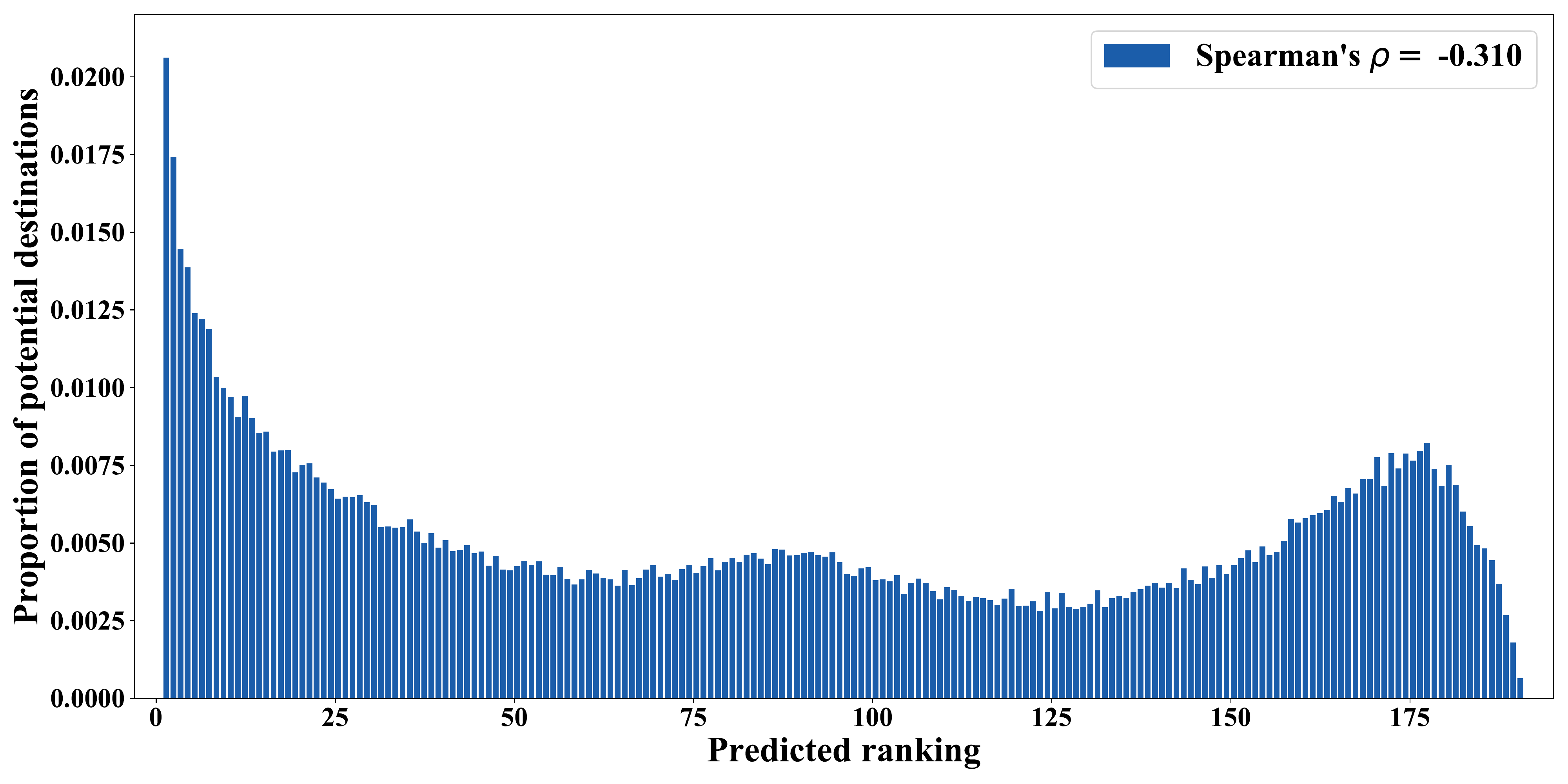}}
\subfigure[MD-SVD]{\includegraphics[width=0.45\textwidth]{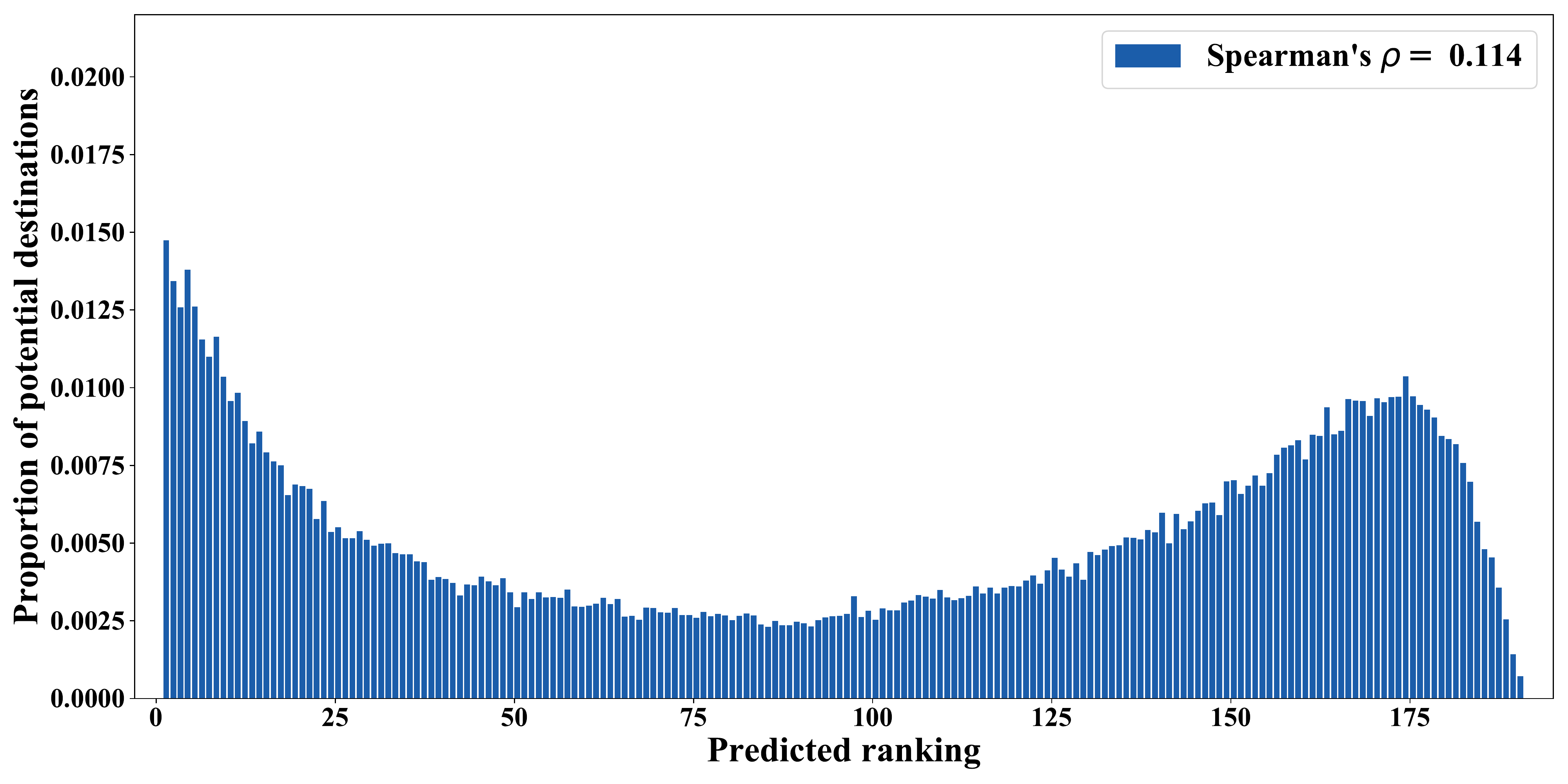}}
\subfigure[CF-U]{\includegraphics[width=0.45\textwidth]{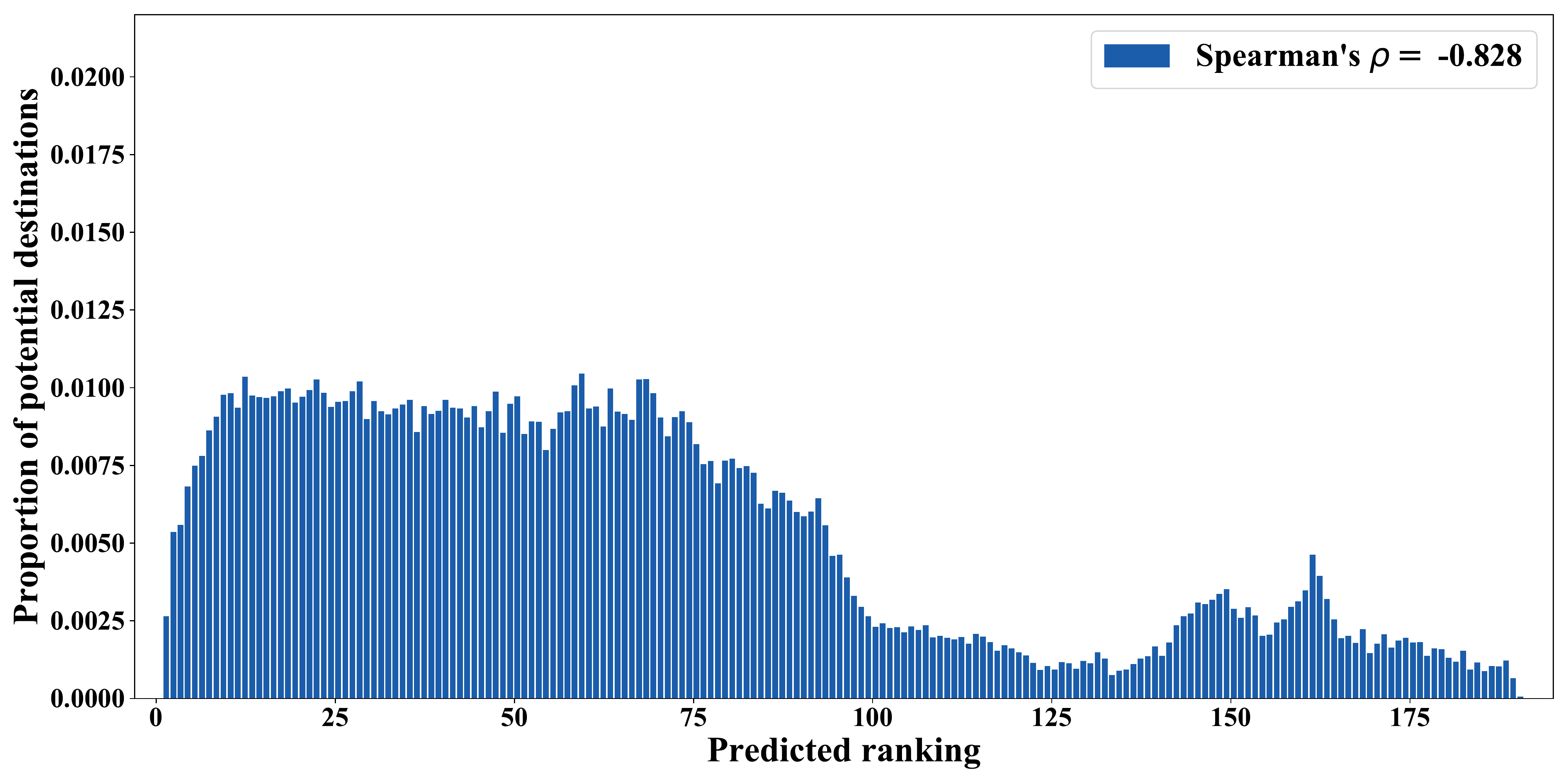}}
\subfigure[CF-I]{\includegraphics[width=0.45\textwidth]{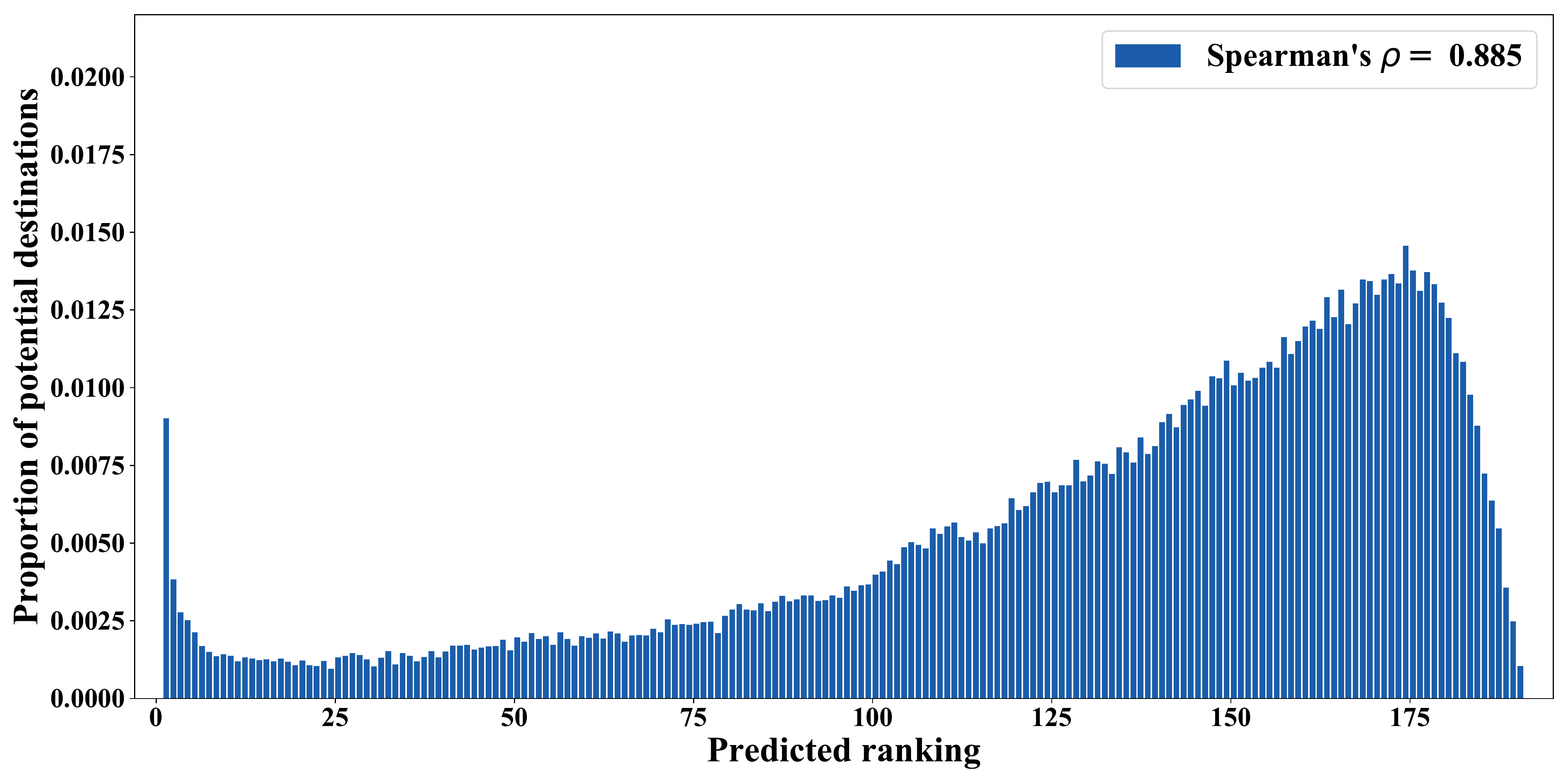}}
\caption{Visualization of $U$ (Performance) of baseline methods.}
\label{fig:comparison}
\end{figure}

The visualizations of the baseline methods' $U$ are shown in Fig.~\ref{fig:comparison}. 
In terms of spearman's $\rho$, all of them do not perform well, and are even worse than HR, which is only based on simple statistics. 
Specifically, RC almost follows a uniform distribution with spearman's $\rho$ close to $0$, meaning it is invalid. The performance of MD-UV is close to RC, except for a significant decline in the tail. Both MD-QR and MD-SVD perform well in the head. However, they all suffer from predicting a large number of potential destinations with very low rankings. This is most notable for CF-I, whose spearman's $\rho$ approaches $1$ instead of $-1$. 
CF-U' distribution $U$ drops rapidly in the middle, indicating it can roughly distinguish the possibility of potential destinations being chosen, i.e., it is reasonable to consider that destinations with very low ranks have a very low probability of being chosen. However, it can't provide valid information when the ranking is in a specific interval like $1 - 75$. 

\subsubsection{Experimental evaluations-aggregated level} \label{sec:ex_eva}

This section will show the evaluation of aggregated performances of different methods using the metrics introduced in Section.~\ref{sec:metrics}. 

\begin{figure}[ht]
\centering
\includegraphics[width=0.86\textwidth]{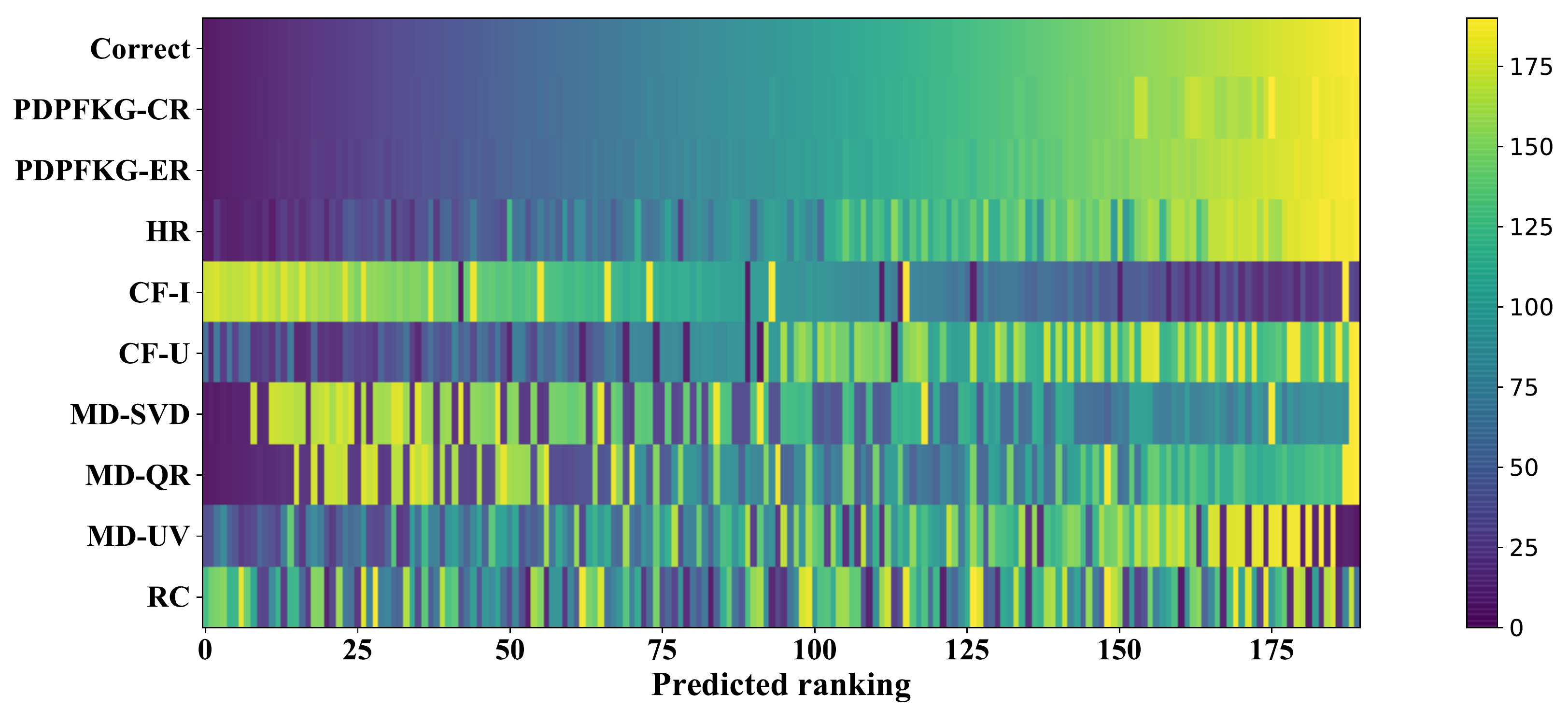}
\caption{Visualization of confusion degree performances of different methods.}
\label{fig:eva_rank}
\end{figure}

First, we use Fig.~\ref{fig:eva_rank} to intuitively demonstrate the ranking confusion degree ($D_f$) of the different methods. The horizontal axis indicates the predicted ranking ($i$), and the value ranking of $p^U(i)$ ($i'$) of different methods' $U$ is noted with gradient color. The completely correct result that satisfies $i=i', \ \forall i\in U$ is shown as a benchmark, noted as ``Correct". The performances of methods on $D_f$ can be judged visually by comparing their distributions of color blocks with the ``Correct". The values of $D_f$ of different methods are calculated and shown in Table~\ref{tab:metric_fs}.

\begin{table}[ht]
\centering
\caption{Performances of confusion degree of different methods.}
\label{tab:metric_fs}
\begin{tabular}{c|c|ccc|cc|c|cc}
\hline
       &        & \multicolumn{3}{c|}{MD}                                         & \multicolumn{2}{c|}{CF}            &      & \multicolumn{2}{c}{\textbf{PDPFKG}}            \\ \cline{3-7} \cline{9-10} 
Method & RC & \multicolumn{1}{c|}{MD-UV}    & \multicolumn{1}{c|}{MD-QR}    & MD-SVD   & \multicolumn{1}{c|}{CF-U}  & CF-I  & HR   & \multicolumn{1}{c|}{PDPFKG-ER}    & PDPFKG-CR   \\ \hline
$D_f$     & 7899   & \multicolumn{1}{c|}{5415}  & \multicolumn{1}{c|}{7627}  & 11245 & \multicolumn{1}{c|}{3828}  & 14030 & 2120 & \multicolumn{1}{c|}{\textbf{295}}   & \textbf{359}   \\ \hline

\end{tabular}
\end{table}

\begin{figure}[H]
    \centering
    \includegraphics[width=0.80\textwidth]{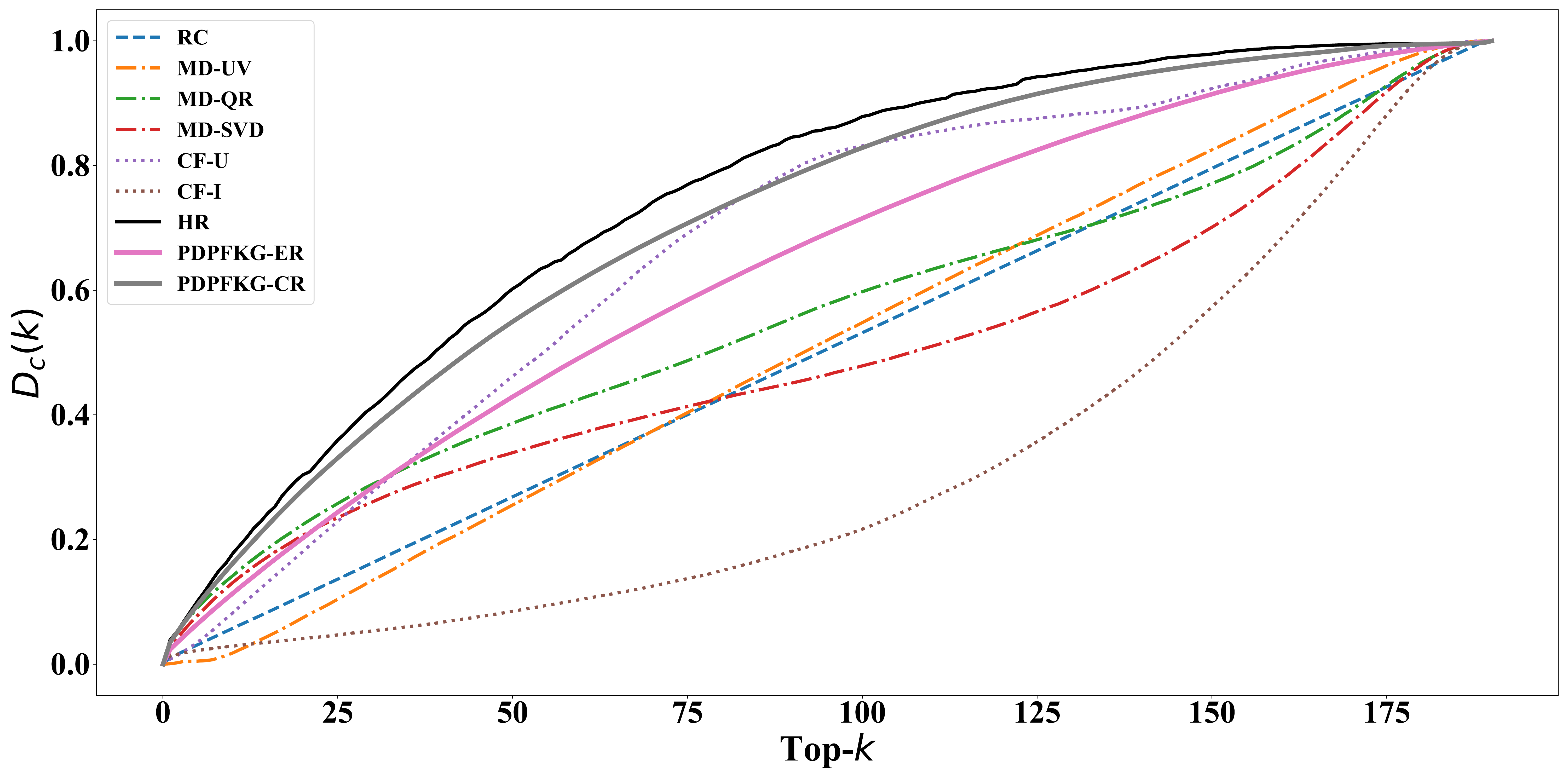}
    \caption{Performances of concentration degree of different methods.}
    \label{fig:model_acc}
\end{figure}

To present a comprehensive view of the concentration degree $D_c(k)$, we calculate $D_c(k)$ with various $k$ ($k=(1,2,3\dots,190)$) of different methods and make Fig.~\ref{fig:model_acc} by plotting $(k,D_c(k))$. As we have mentioned, it can be presented in another form about recall, and we provide it in Appendix.~\ref{ap:app} Fig.~\ref{fig:accurecy}.

According to Fig.~\ref{fig:eva_rank} and Table~\ref{tab:metric_fs}, PDPFKG's performance on confusion degree ($D_f$) is significant better than any other methods, and reach a level that close to the "Correct". It means the discovery or prediction results of PDPFKG has statistical reliability. Besides, the smoothness of PDPFKG-CR's distribution $U$ is notable since it suggests the probability of predicted rankings may be portrayed as concise functional forms. If PDPFKG has similar performances (i.e., distribution $U$ with low confusion degree and smoothness) in other cities, and there exists a unified functional, it will be important for exploring urban trip patterns of travelers.

Fig.~\ref{fig:model_acc} shows HR has the best performance on concentration degree ($D_c(k)$). However, we have mentioned HR's performance depends on the city's context and cannot be improved. The improvement on $D_c(k)$ of PDPFKG-CR over PDPFKG-ER comes from the integration of statistical information by combining hotness ranking. As the extremely low confusion degree is maintained, PDPFKG-CR performs superior to any other methods overall. The performance of PDPFKG-ER on $D_c(k)$ is not yet very ideal, but we will show PDPFKG has great potential for improvement in the following sections.

\subsubsection{Experimental evaluations-individual level} \label{sec:ex_eva_ini}

\begin{figure}[h]
\setlength{\abovecaptionskip}{0.cm}
\centering
\includegraphics[width=0.95\textwidth]{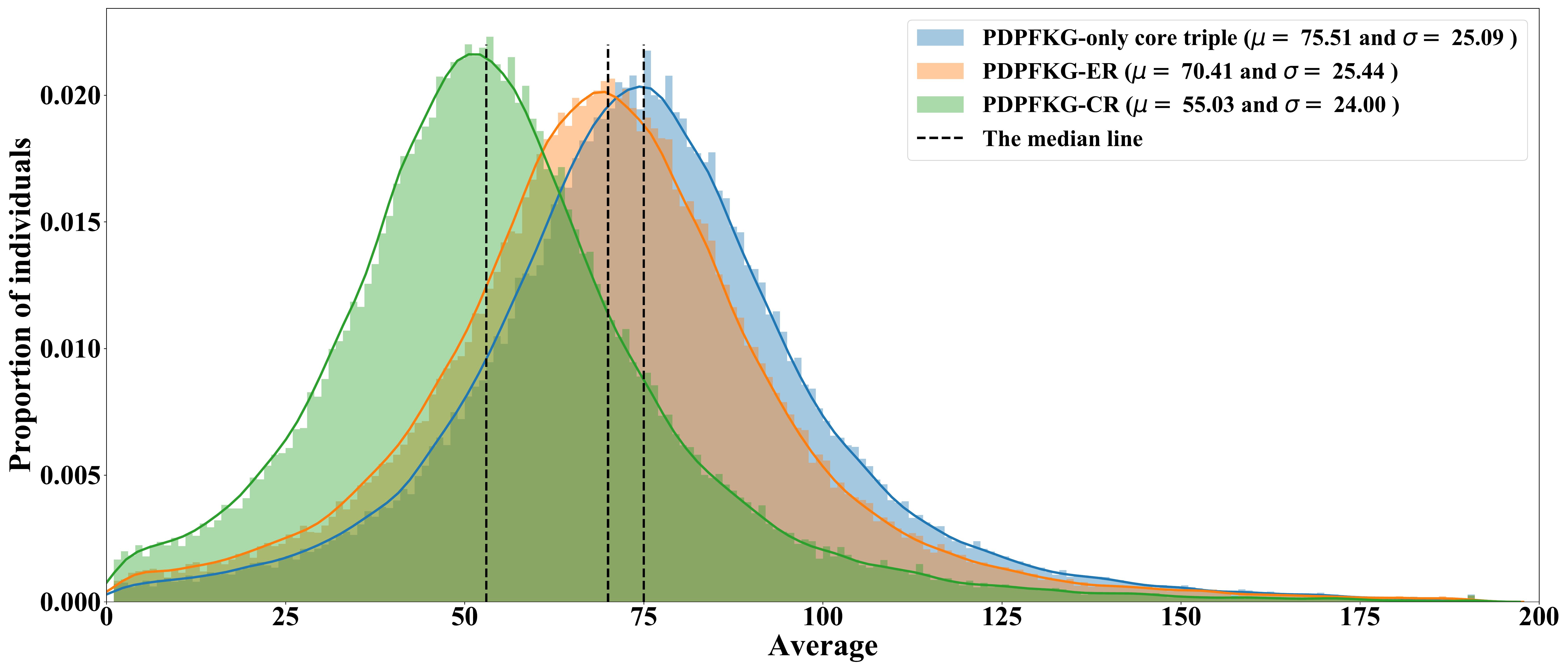}
\caption{Distribution of the average ranking of the individual's potential destinations.}
\label{fig:indi_dis}
\end{figure}

The distributions $H$ about the average of individuals' prediction results are visualized in Fig.~\ref{fig:indi_dis}. PDPFKG with only core triple (blue bar) indicates the PDPFKG whose TKG only contains the core triple (see Table.~\ref{tab:KG_triple}) for TKGEM training (Its aggregated-level result is shown in Fig.~\ref{fig:discussion_info}(b), Section.~\ref{sec:dis1}). The aggregated-level results of PDPFKG-ER and PDPFKG-CR correspond to Fig.~\ref{fig:our method}(b) and Fig.~\ref{fig:our method}(c), respectively. 

There are two aspects of valuable information we can conclude from  Fig.~\ref{fig:indi_dis}. First, the shapes of $H$ indicate the performances of PDPFKG follows a single-peak distribution from the individual perspective. It means most individuals' results performed similarly overall and concentrated in a range, with few individuals whose potential destinations are very predictable or extremely difficult to predict. Further, the distributions $H$ predicted by PDPFKG are very smooth, and we found they approximate normal distribution or student's $t$-distribution. 

Second, these three distributions $H$ have a similar shape but different positions. The distribution $H$ of PDPFKG-ER can be regarded as a left shift from PDPFKG with only core triple, indicating an overall improvement of the performance. Considering both are based on PDPFKG, the only difference is that PDPFKG-ER has richer data types (all triples shown in Table.~\ref{tab:KG_triple}). Thus it can be concluded that the introduction of valid data types would improve the performance under the framework PDPFKG. In other words, PDPFKG can effectively take advantage of the information contained in various data and enhance the prediction or discovery. It shows that PDPFKG is a scientific framework, and it has the potential to be further enhanced if new types of data are introduced. We have discussed the information reflected by the difference of $H$ of PDPFKG-ER and PDPFKG-ER when evaluating their aggregated performances in Section.~\ref{sec:over_per_vis}. We want to emphasize that our purpose of integrating the traffic zone's hotness information by combining rankings is to prove that the information PDPFKG learned is not statistical. Hotness ranking performs well in XuanCheng city and improves the performance of PDPFKG, but it is not always valid in different regions or cities with various contexts. For example, the performance of hotness ranking would be similar to random choice (RC) for a region that is uniformly visited.

\section{Discussion}

In this section, we first respond to the crucial points for PDPFKG development (e.g., the private relationship for TKG construction and embedding algorithm adaptation), which we mentioned in Section.~\ref{sec:Method}. Next, we discuss the application of location-based and deep learning methods to our task. Finally, we summarize and share our understanding of the advantages of knowledge graphs on the individual's trip prediction and knowledge discovery.


\subsection{Discussions on crucial points} \label{sec:dis1}
All the blue bar charts shown in this section have the same format and meaning as before (e.g., Fig.~\ref{fig:our method}), for which we have not labeled the horizontal and vertical axes for a better display effect.

\subsubsection{Non-core triples} \label{sec:non_coretriple}

\begin{figure}[ht]
\centering
\subfigure[With non-core triples]{\includegraphics[width=0.49\textwidth]{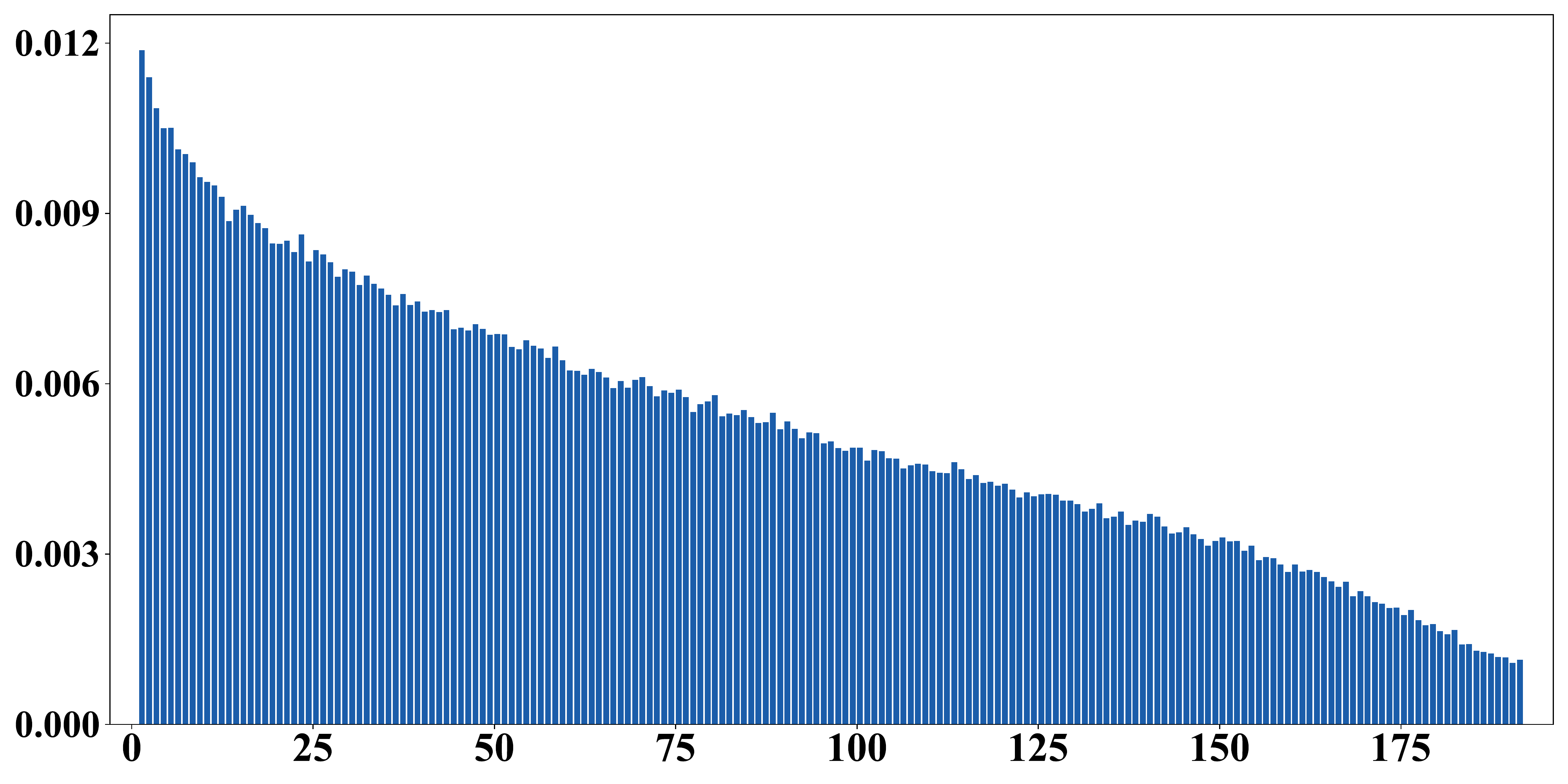}}
\subfigure[Only core triple]{\includegraphics[width=0.49\textwidth]{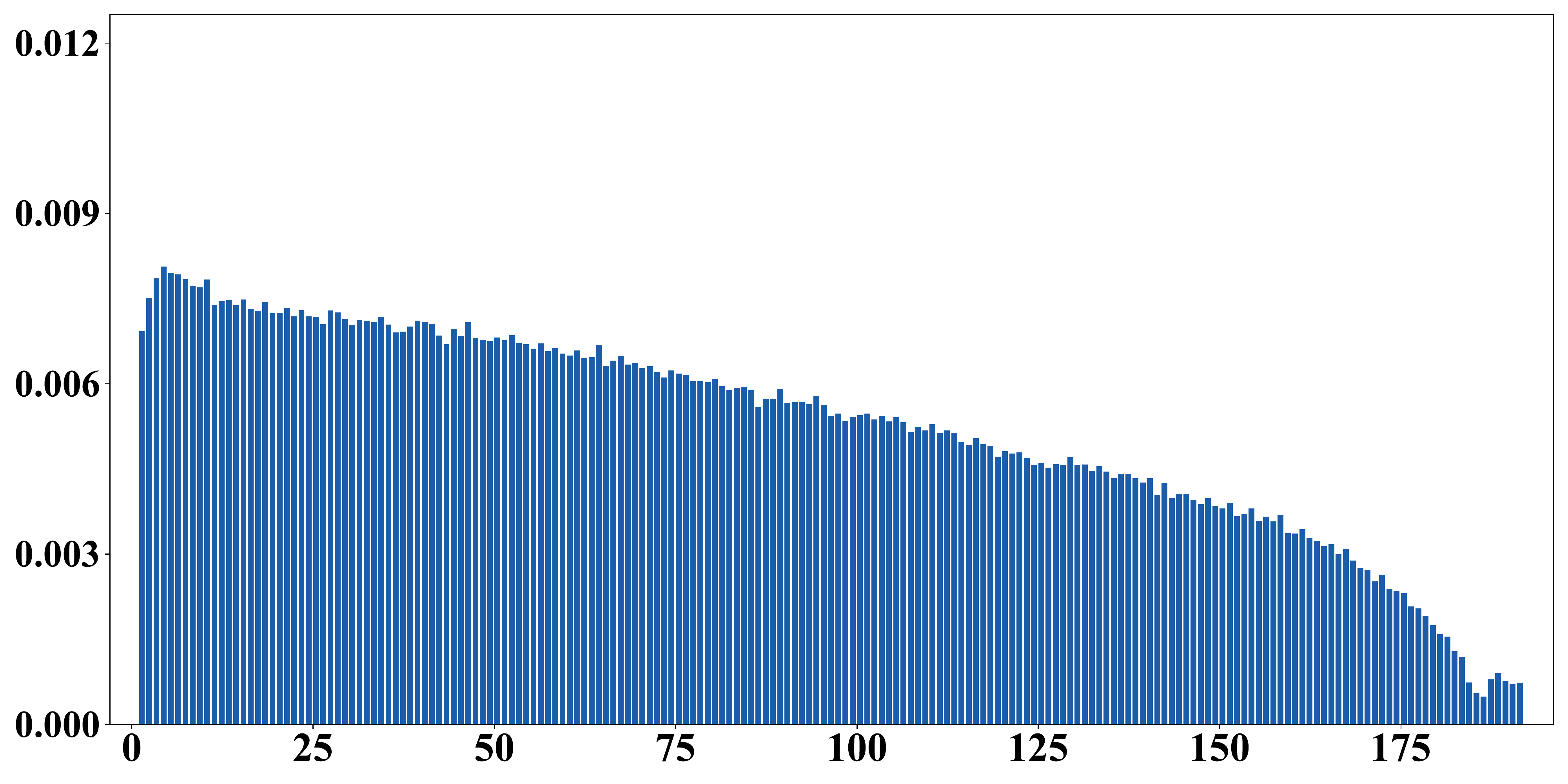}}

\caption{Performances of model with and without non-core triples.}
\label{fig:discussion_info}
\end{figure}

PDPFKG works as long as the core triple is available. Non-core triples function through influencing the core triple's representation during TKGEM training process, by which we want to enhance the prediction. Fig.~\ref{fig:discussion_info}(b) shows the $U$ of PDPFKG with only the core triple. It can be seen the improvement boosted by non-core triples is noticeable on aggregated level, and the comparison on the individual level has been shown and discussed in Section.~\ref{sec:ex_eva_ini}. This proves the information we introduce to TKG (Section.~\ref{sec:Structure}) is valid for potential destination prediction, and PDPFKG can use them effectively. 

\subsubsection{Private relationship}

\begin{figure}[ht]
\centering

\subfigure[Public relationship; dimension = 148]{\includegraphics[width=0.49\textwidth]{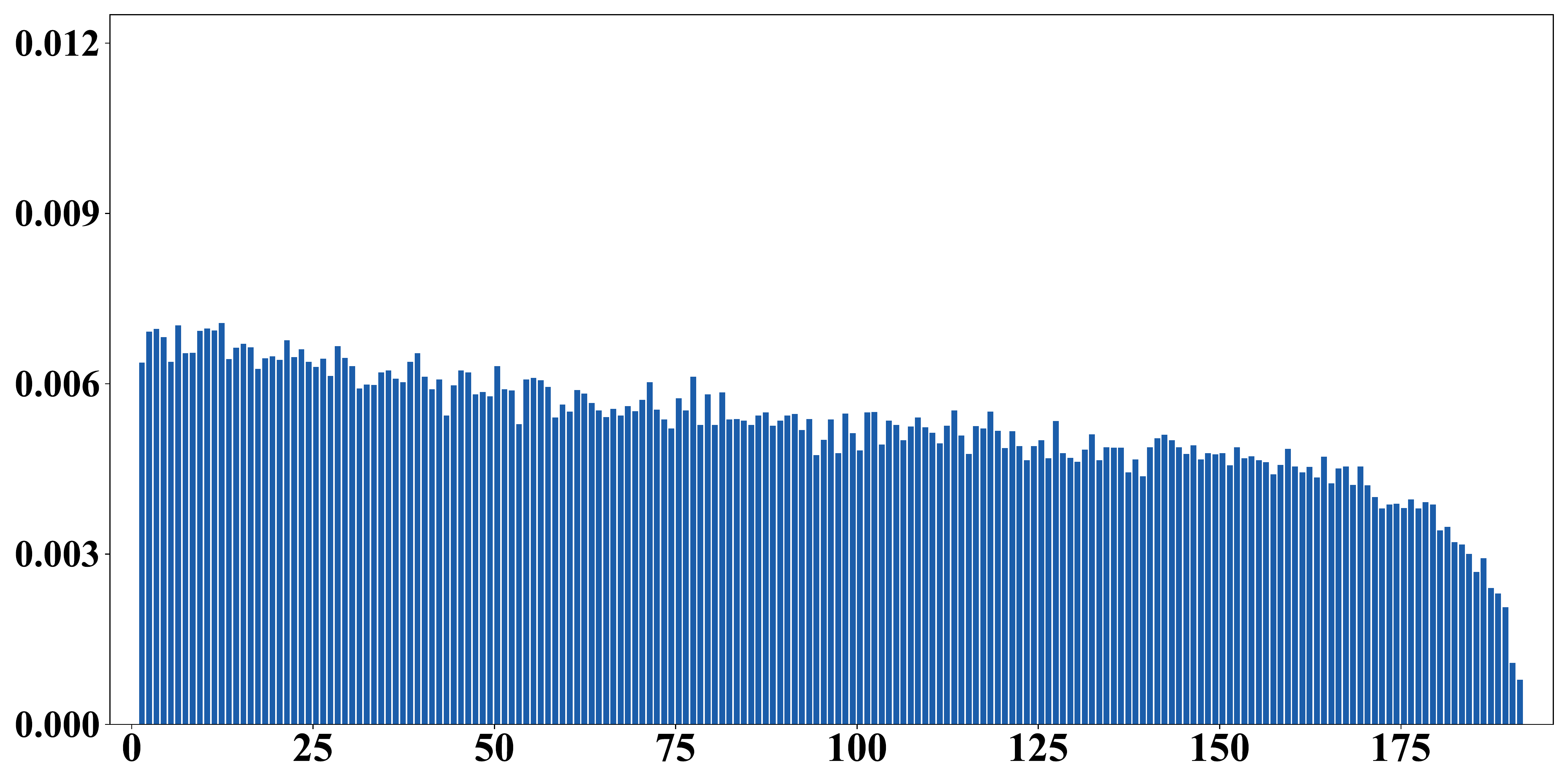}}
\subfigure[Public relationship; dimension = 600]{\includegraphics[width=0.49\textwidth]{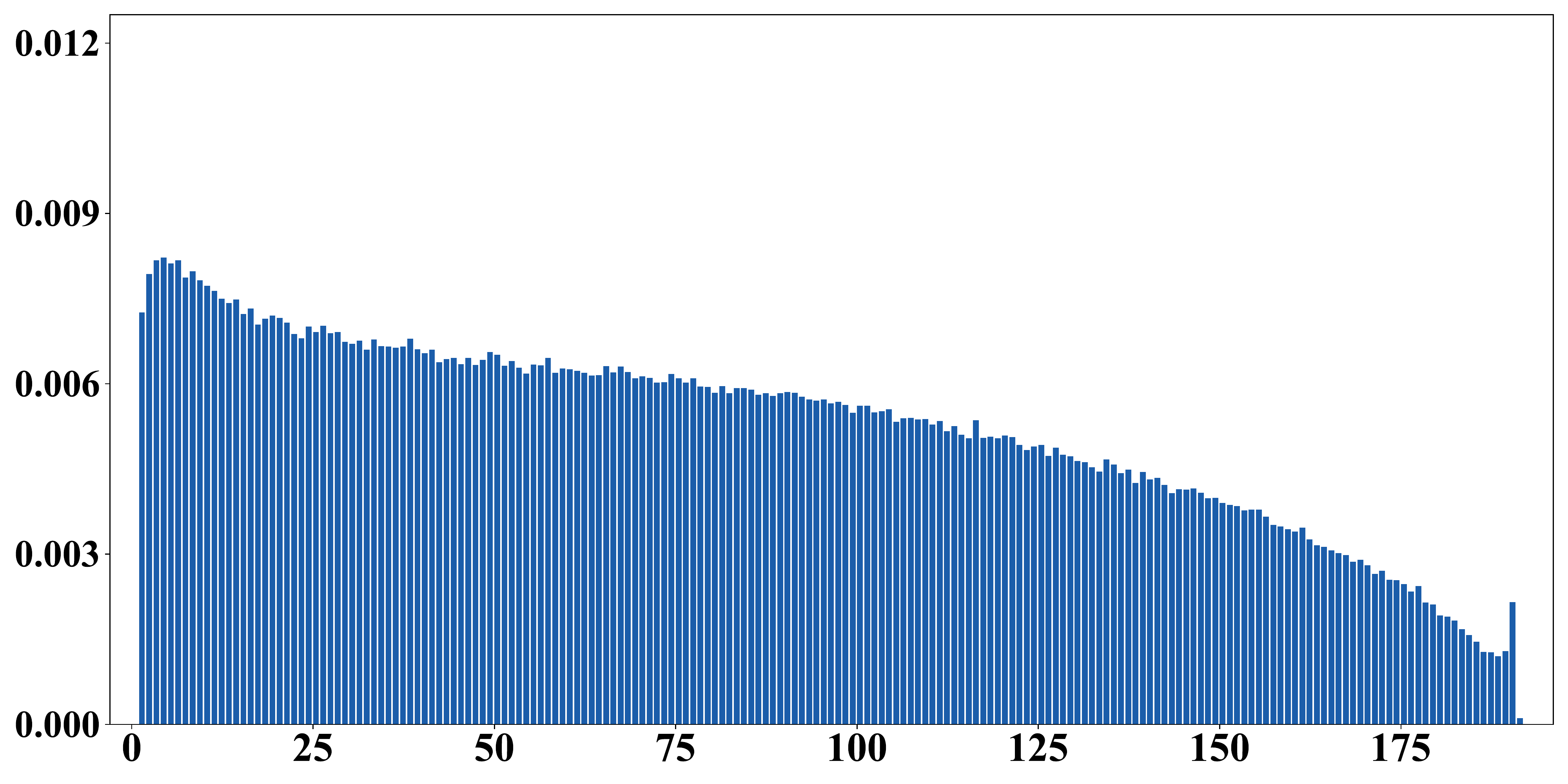}}

\caption{Performance of adopting public relationship.}
\label{fig:discussion_stru}
\end{figure}

In Section \ref{sec:Structure}, we propose the concept of private relationship, and claim it is more scientific compared to public relationship ($Choose\_D$) for TKG construction. Here we would like to discuss it through experiments, as well as explain the principles of relationship building proposed in Section.~\ref{sec:Structure} further.
First, We do an experiment with the settings of experiment Fig.~\ref{fig:discussion_info}(b), only adjusting the private relationship to public relationship. Its overall performance is shown in Fig.~\ref{fig:discussion_stru}(a), which has a large gap with Fig.~\ref{fig:discussion_info}(b). Next, we conduct experiments raising the embedding dimension of public relationship, and get Fig.~\ref{fig:discussion_stru}(b) that similar to Fig.~\ref{fig:discussion_info}(b) until the dimension reaches $600$. It indicates that the performance of Fig.~\ref{fig:discussion_stru}(a) is caused by insufficient representation of dimensions since public relationship has a higher complexity. Moreover, it proves the more complex the relationship is, the larger the optimal dimension is. 

Through the above experiments, it is conceivable that the optimal dimension of $Choose\_D$ and $Has\_POI$ will differ greatly due to the difference in complexity, which is bad for determining the training dimension. In addition, the variation of public relationship complexity with the data scale also leads to the optimal dimension being tied to the scale of the dataset.

\subsubsection{Negative sampling strategy}

\begin{figure}[ht]
\centering

\subfigure[Random replacement]{\includegraphics[width=0.49\textwidth]{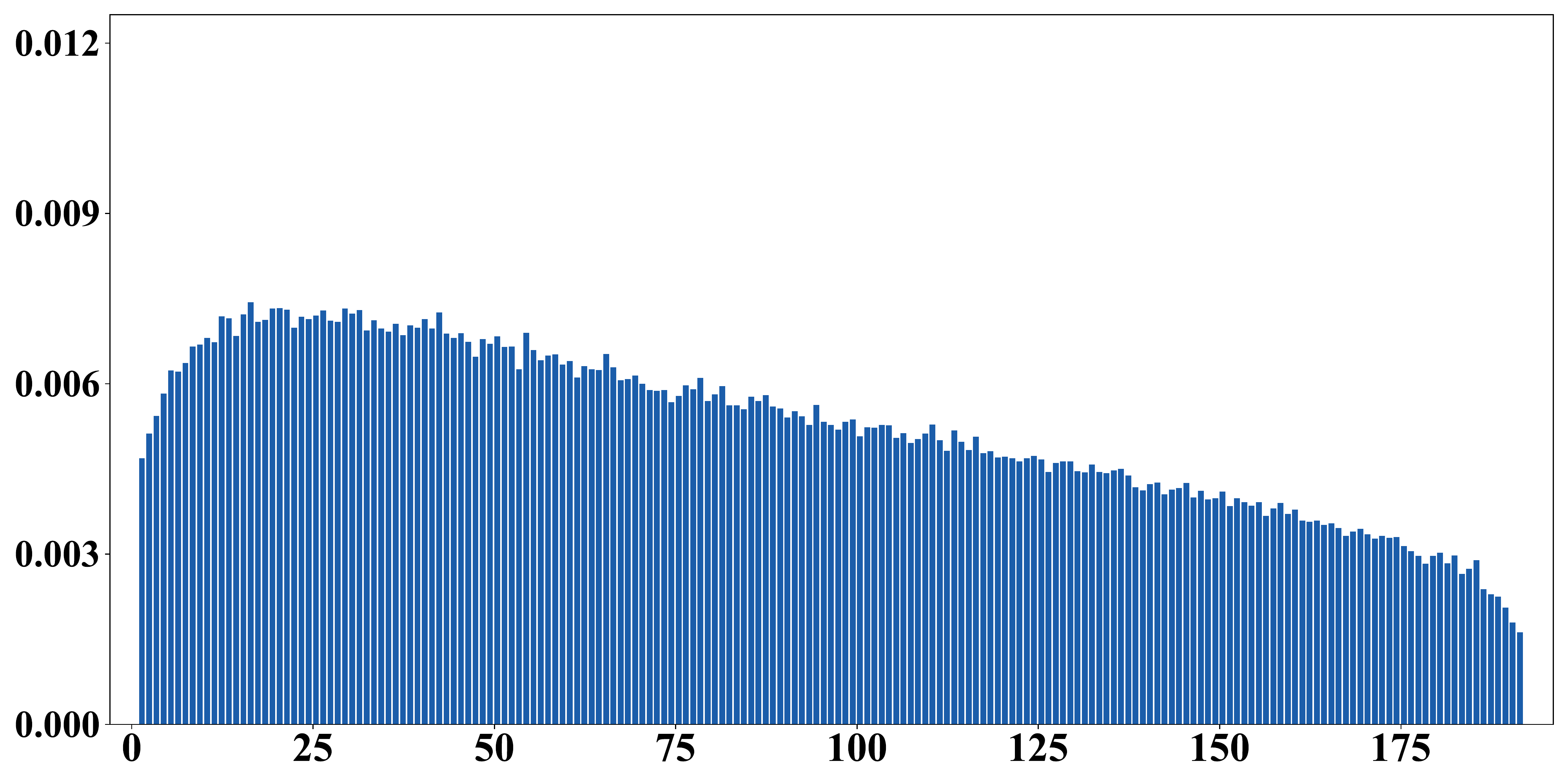}}
\subfigure[Controlled replacement]{\includegraphics[width=0.49\textwidth]{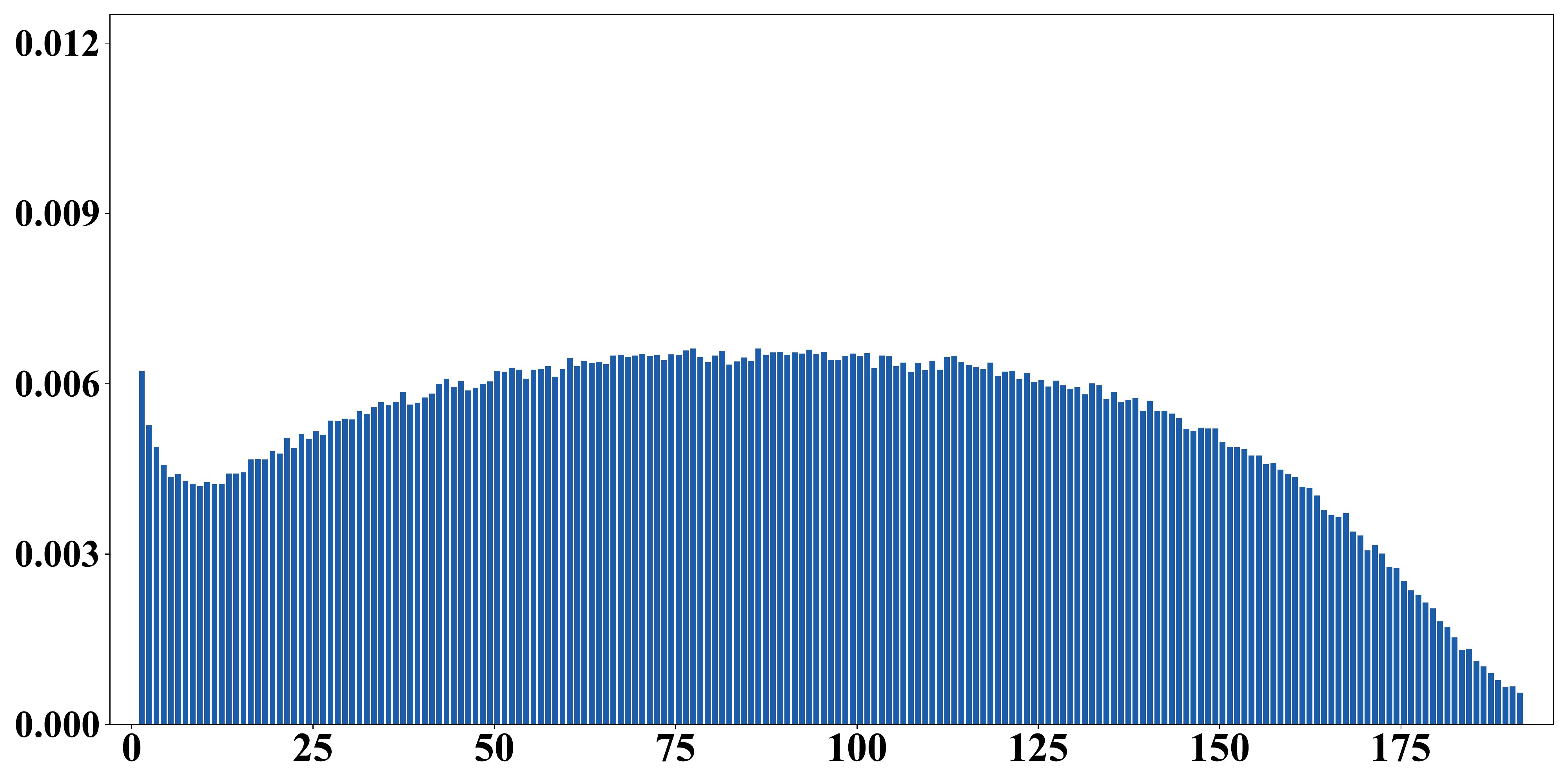}}

\caption{Performance of adopting the negative sampling strategy.}
\label{fig:discussion_neg}
\end{figure}

In Section \ref{sec:embedding}, we have mentioned that the negative sampling strategy is not adapted to TKG, which is constructed based on incomplete observation data. To further illustrate it, two models adopting the negative sampling strategy are trained, with the optimization objective changing to Eq.~\ref{eq:objective function}.
The first model generates negative samples by random replacement mentioned in Section \ref{sec:embedding1}.
The other one adopts a controlled replacement we designed, which guarantees that the type of replaced entity or relationship is different from the original one. These two models' performances are shown in Fig.~\ref{fig:discussion_neg}. The concentration of Fig.~\ref{fig:discussion_neg}(a) at the head becomes significantly worse. This is because many true core triples unobserved are trained as negative samples. Theoretically, controlled replacement does not produce possibly true triples. Nevertheless, it performs even worse and seems that it spoils the training. 

\subsubsection{Embedding dimension}

\begin{figure}[ht]
\centering
\subfigure[Dimension=37]{\includegraphics[width=0.32\textwidth]{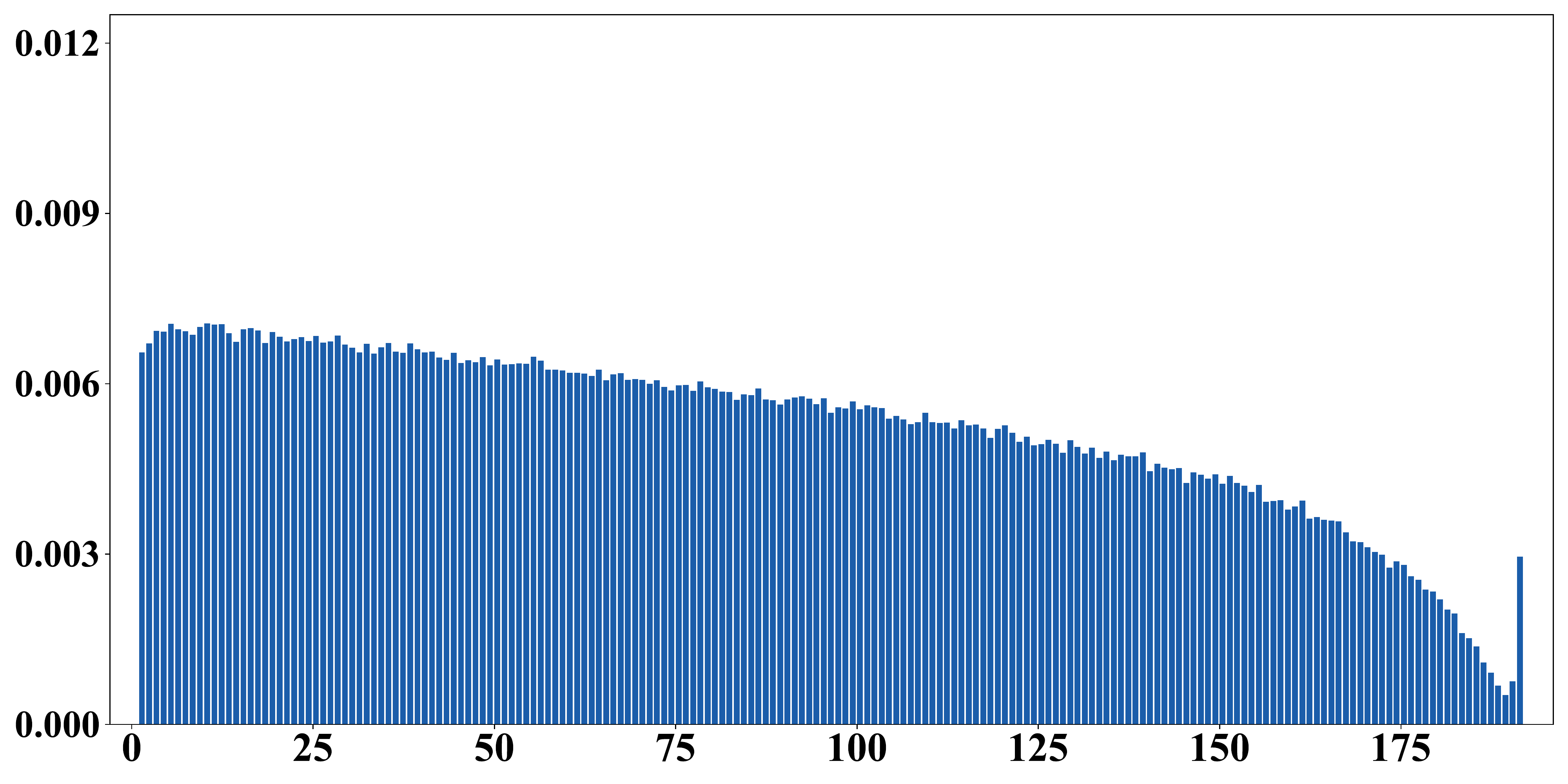}}
\subfigure[Dimension=76]{\includegraphics[width=0.32\textwidth]{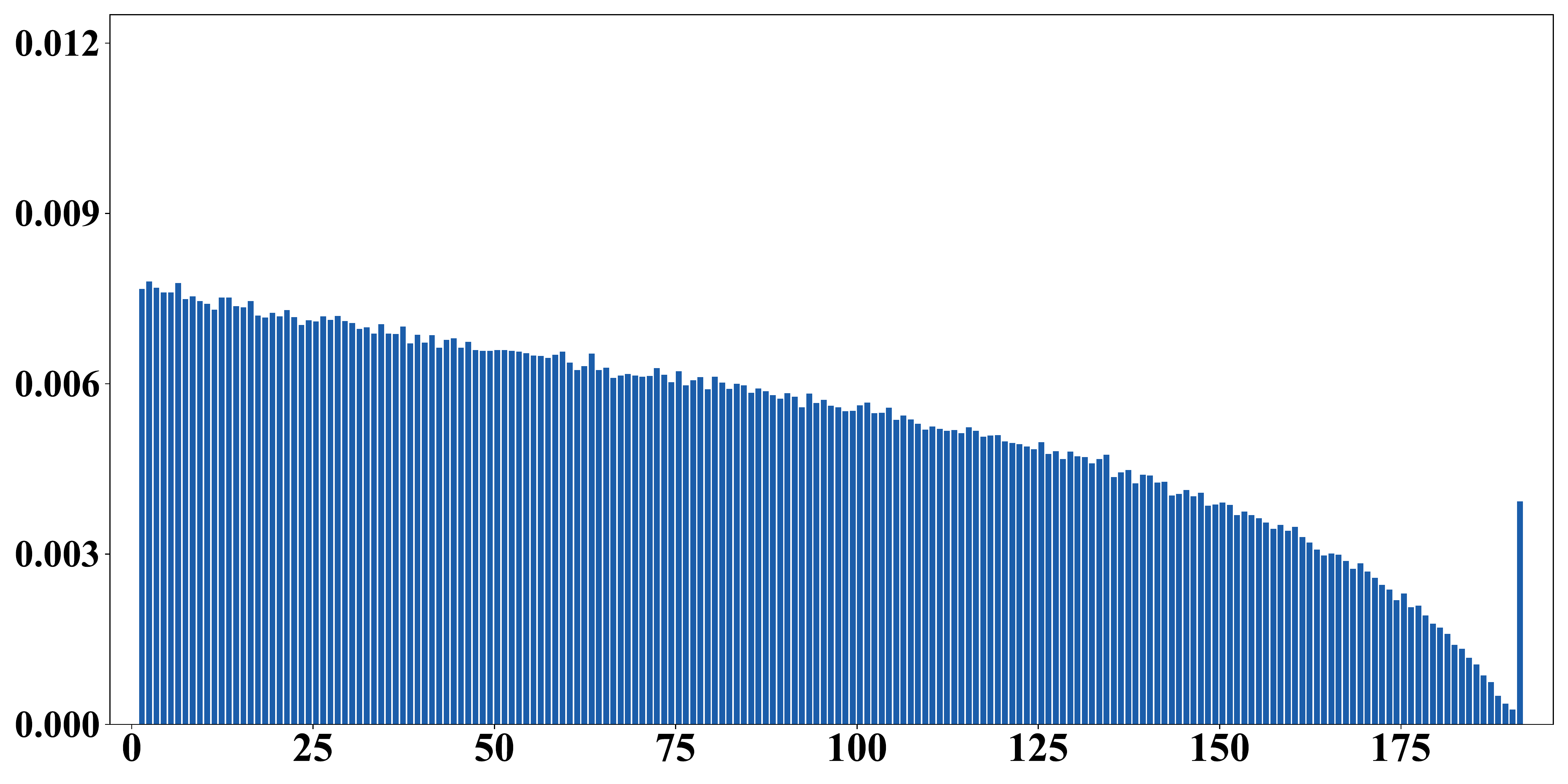}}
\subfigure[Dimension=106]{\includegraphics[width=0.32\textwidth]{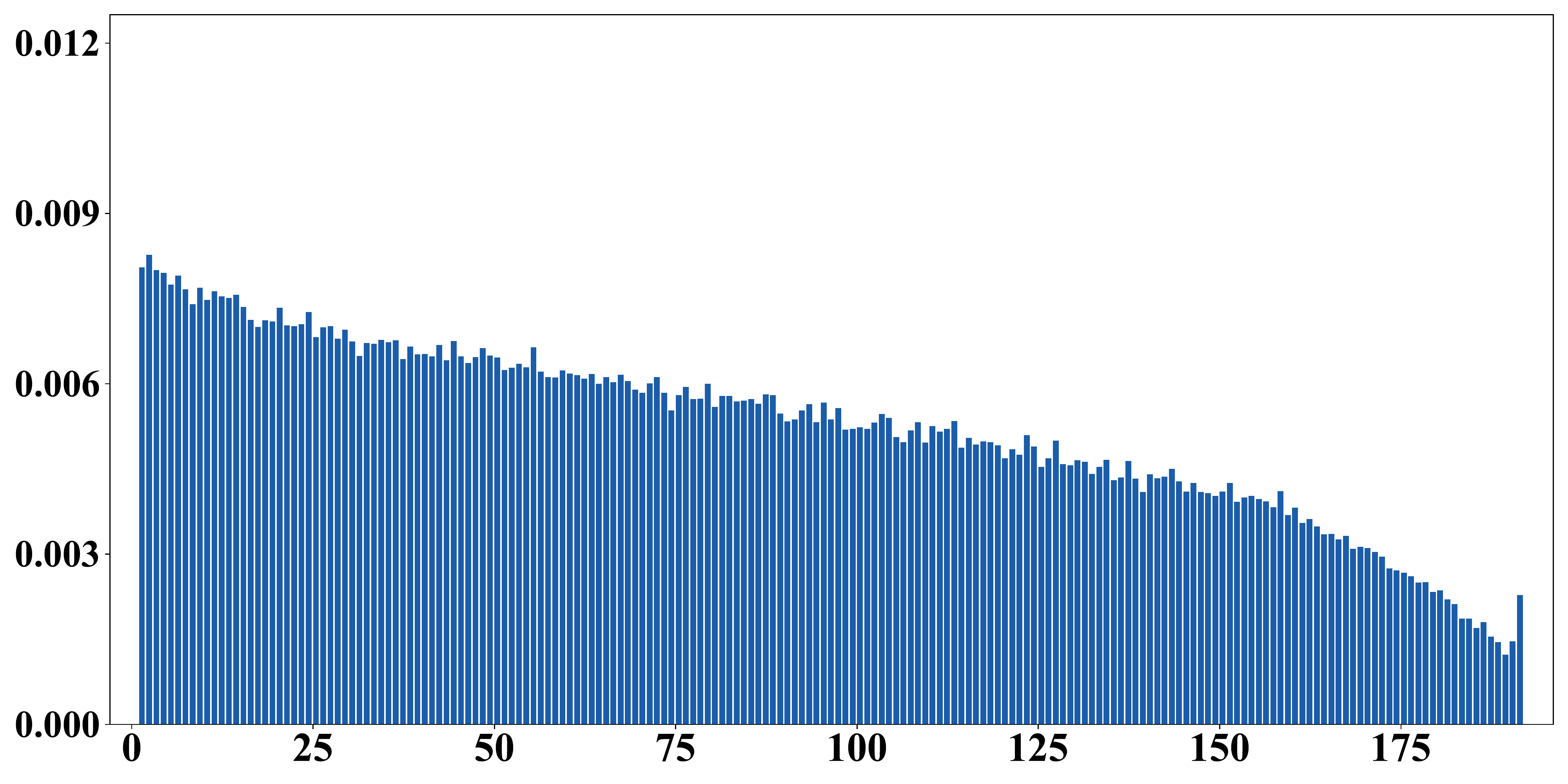}}
\subfigure[Dimension=124]{\includegraphics[width=0.32\textwidth]{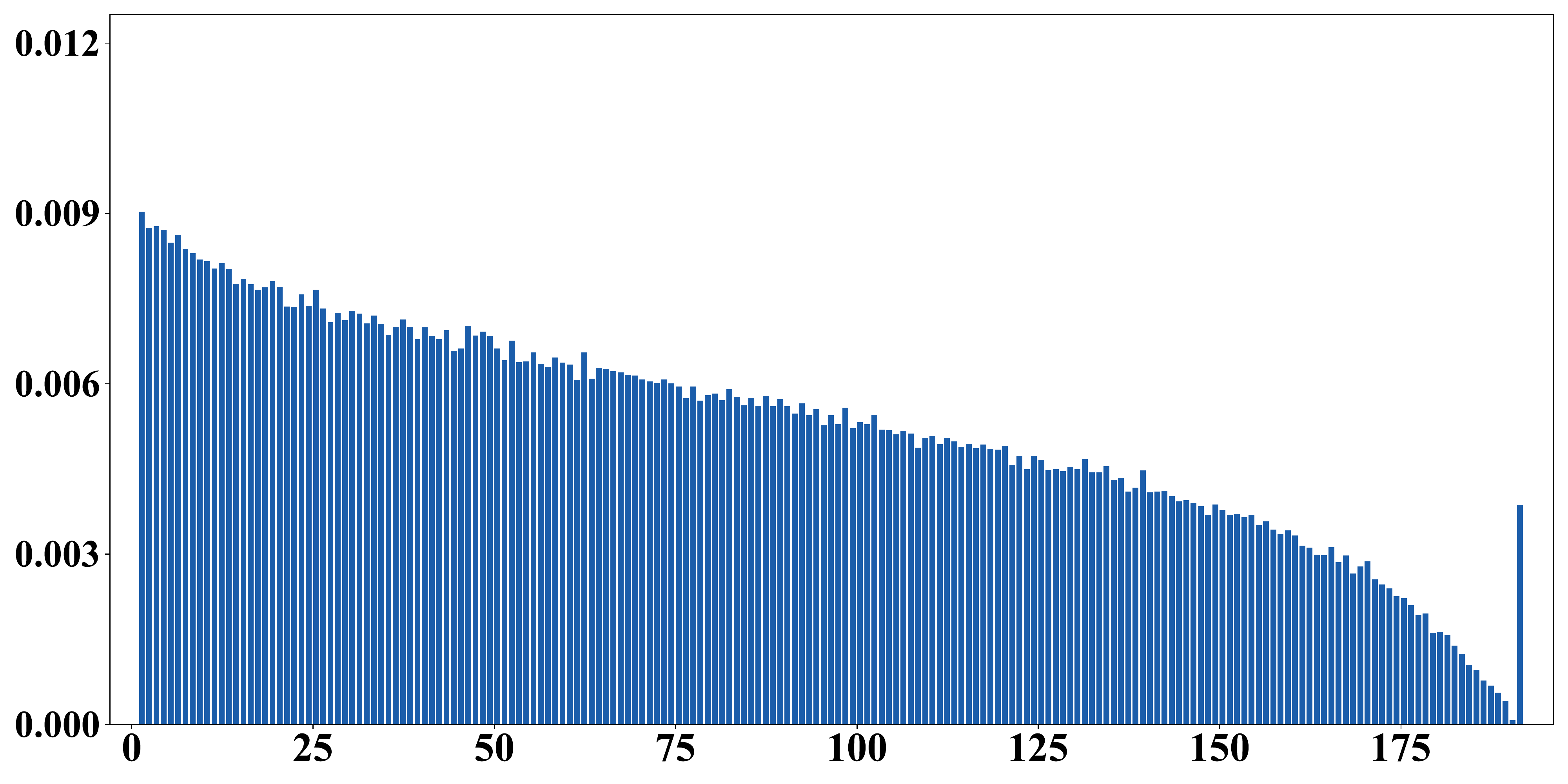}}
\subfigure[Dimension=148]{\includegraphics[width=0.32\textwidth]{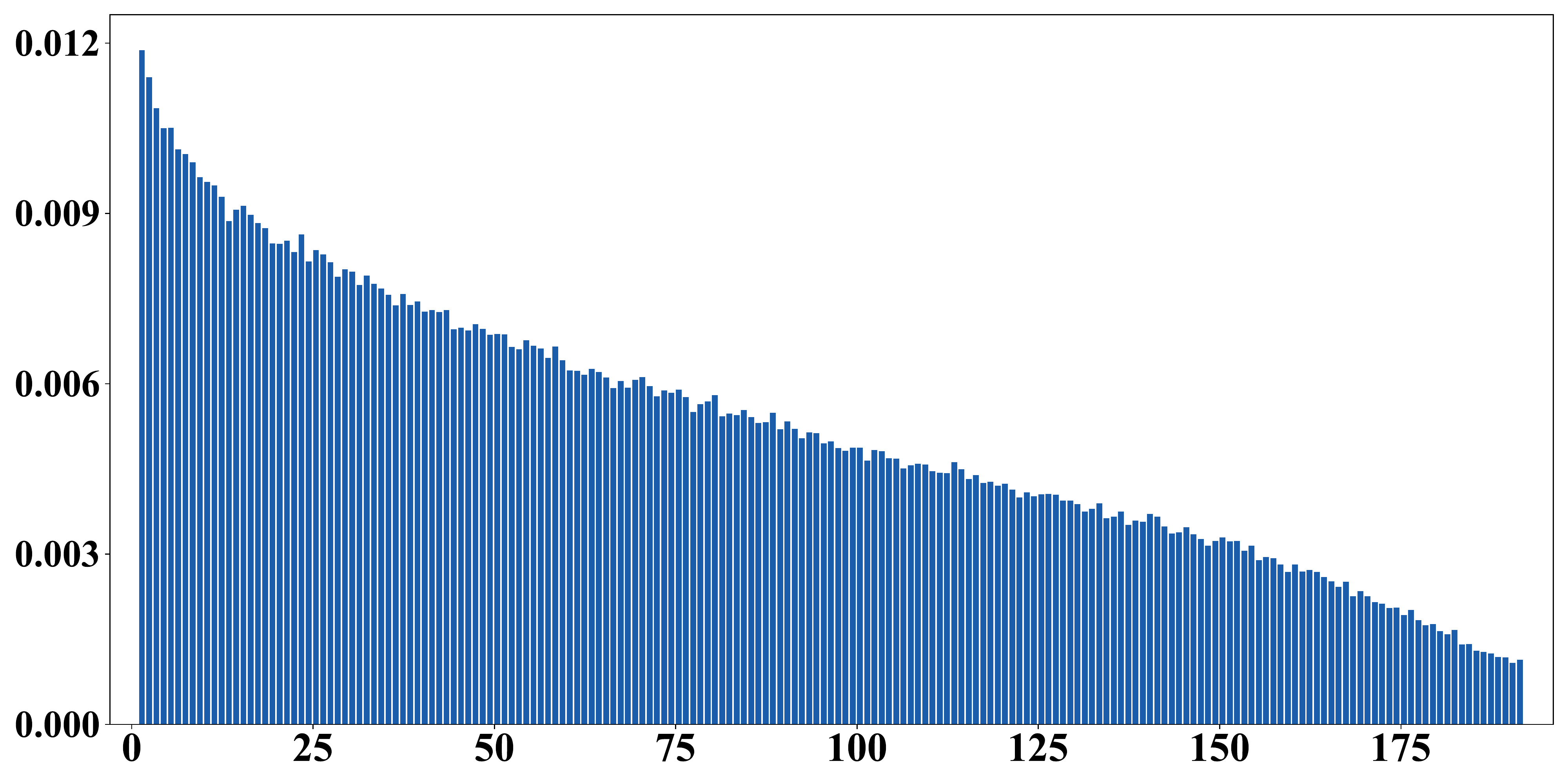}}
\subfigure[Dimension=160]{\includegraphics[width=0.32\textwidth]{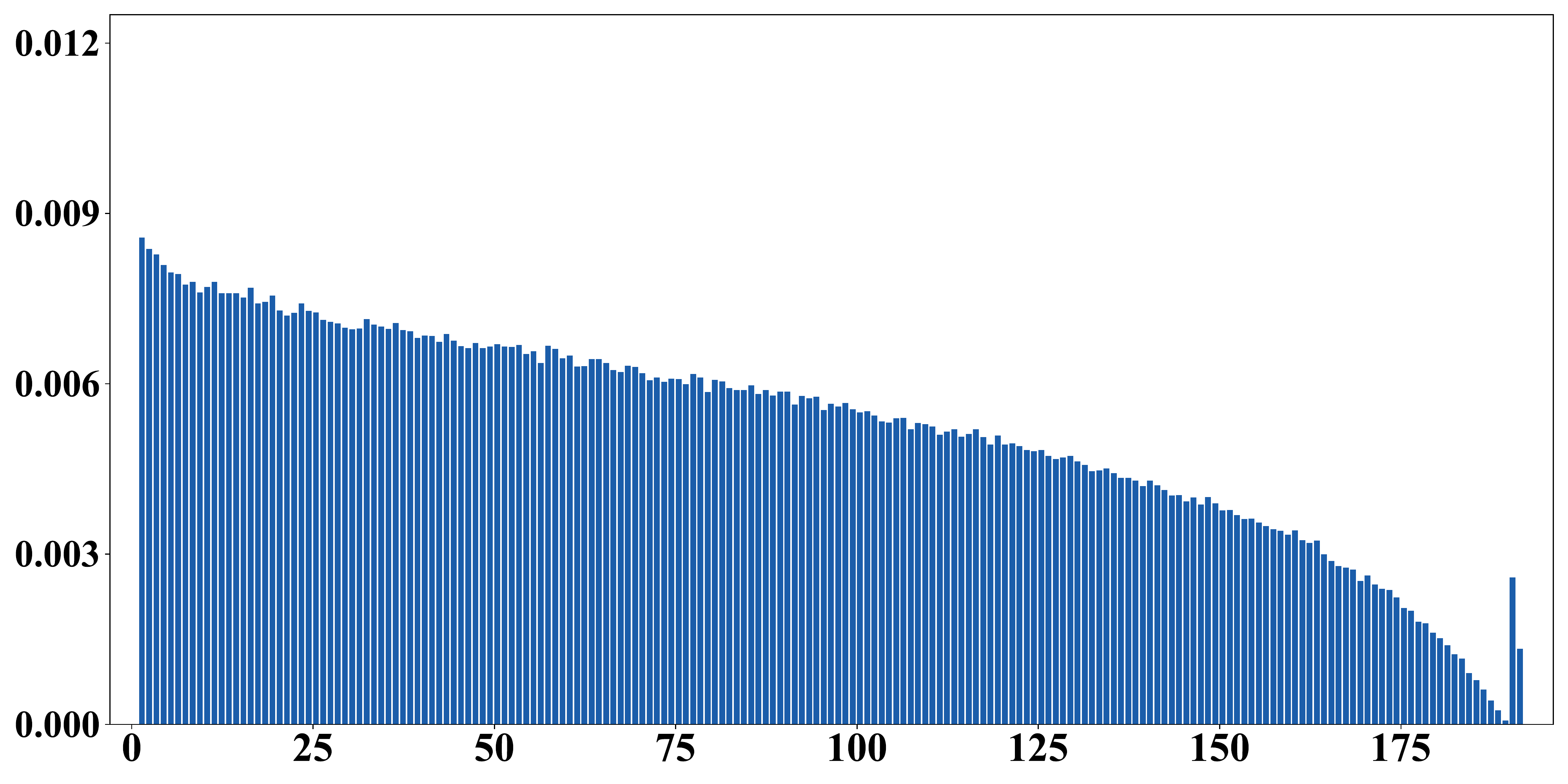}}
\caption{Performance of model in different dimensions.}
\label{fig:discussion of dimensions}
\end{figure}

The dimension of embedding is a critical parameter of TKGEM. In general, there is an optimal dimension that makes the model perform best for a given data and task. If the model's performance varies regularly with the dimension, it will significantly reduce the work of parameter adjusting. Fig.~\ref{fig:discussion of dimensions} shows the performances of TKGEM with different dimensions. It illustrates that the embedding dimension of TKGEM has a remarkable correlation with its performance. In addition, considering the migration of the optimal dimensions mentioned in Section \ref{sec:Structure}, we can migrate to other datasets by calibrating the optimal dimensions on one dataset. 

\subsubsection{Performance over different future time horizons} \label{sec:diff_period}

\begin{figure}[ht]
\centering
\subfigure[$T^f$=7 days]{\includegraphics[width=0.32\textwidth]{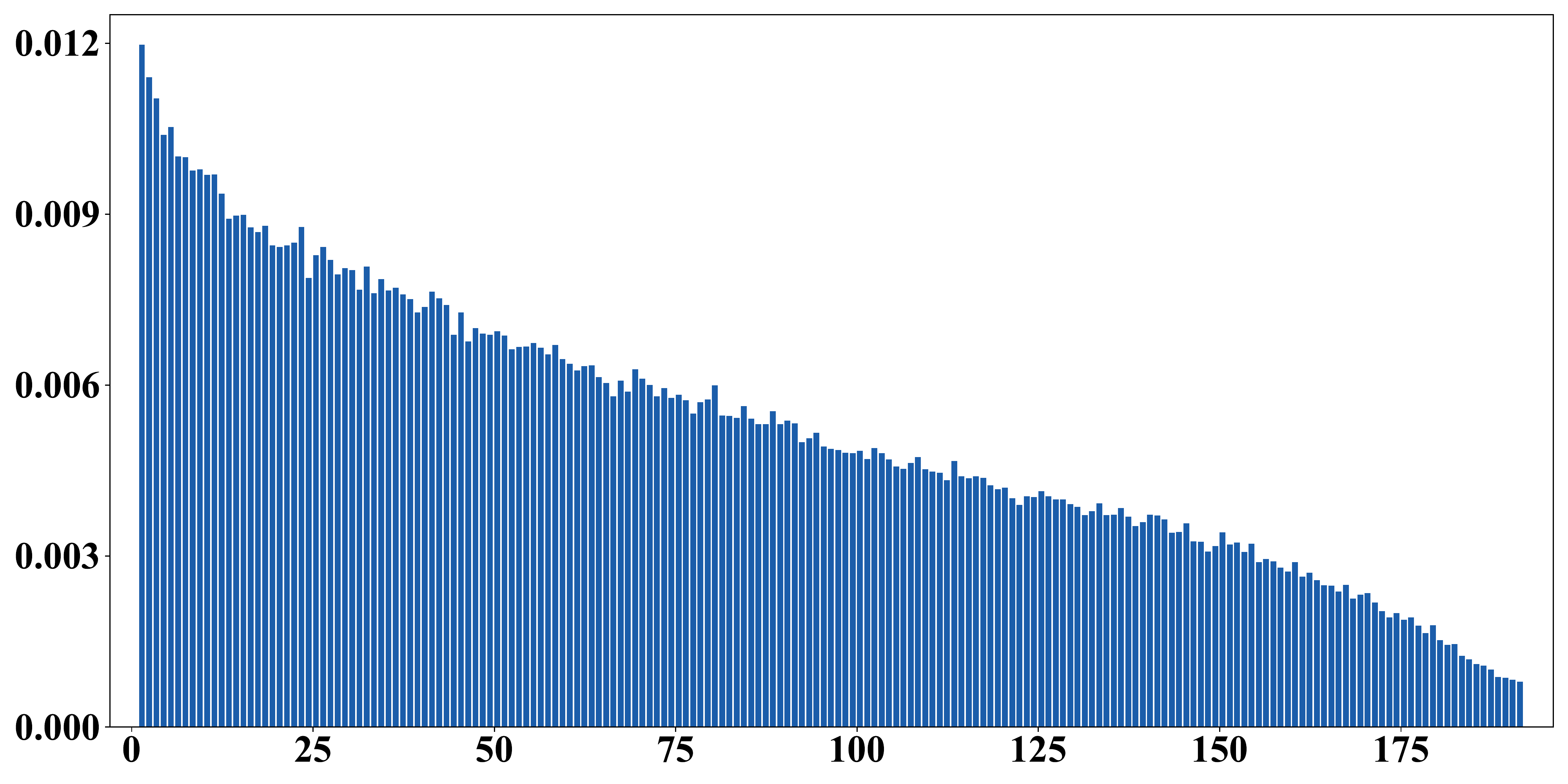}}
\subfigure[$T^f$=21 days]{\includegraphics[width=0.32\textwidth]{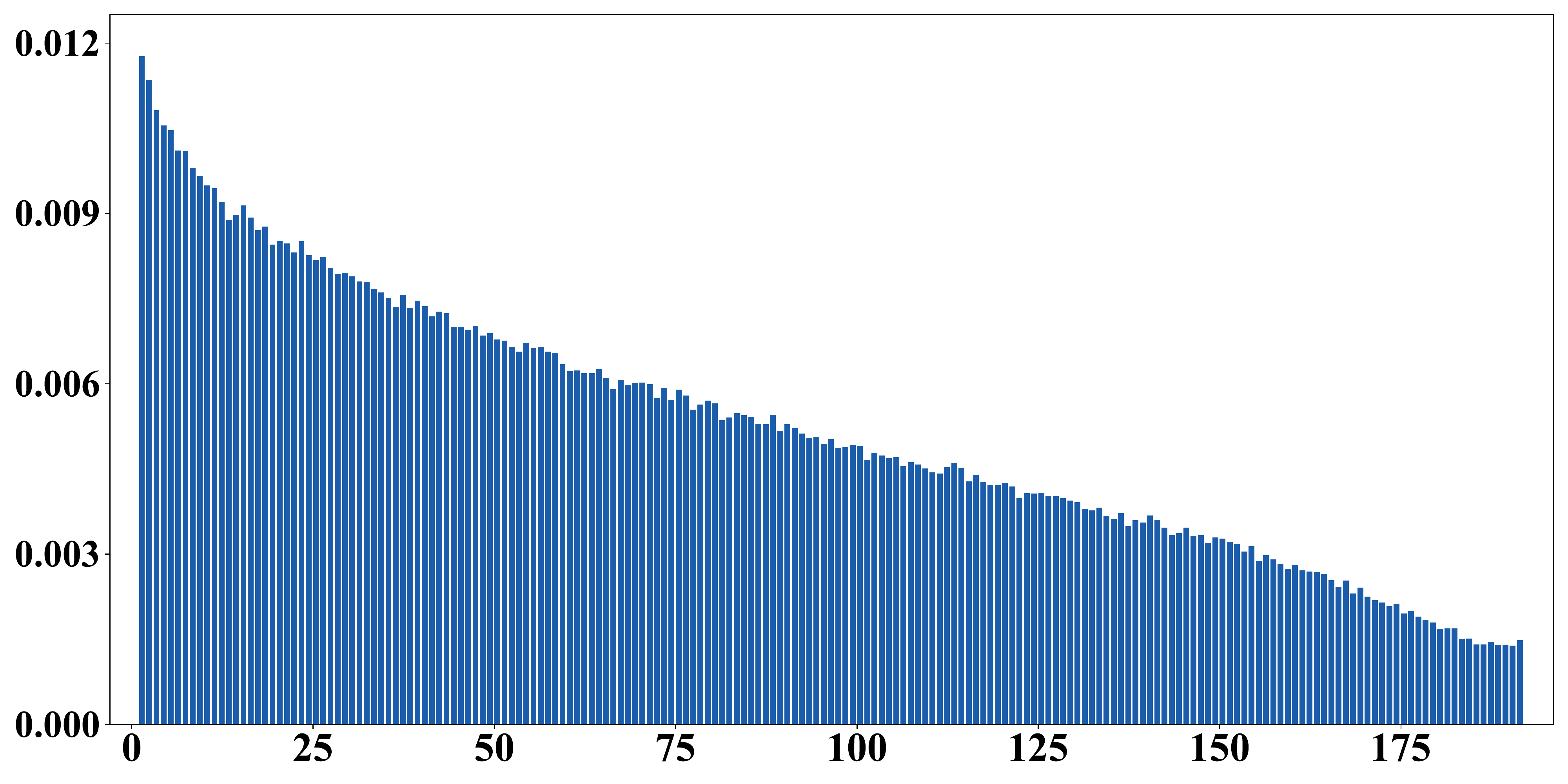}}
\subfigure[$T^f$=28 days]{\includegraphics[width=0.32\textwidth]{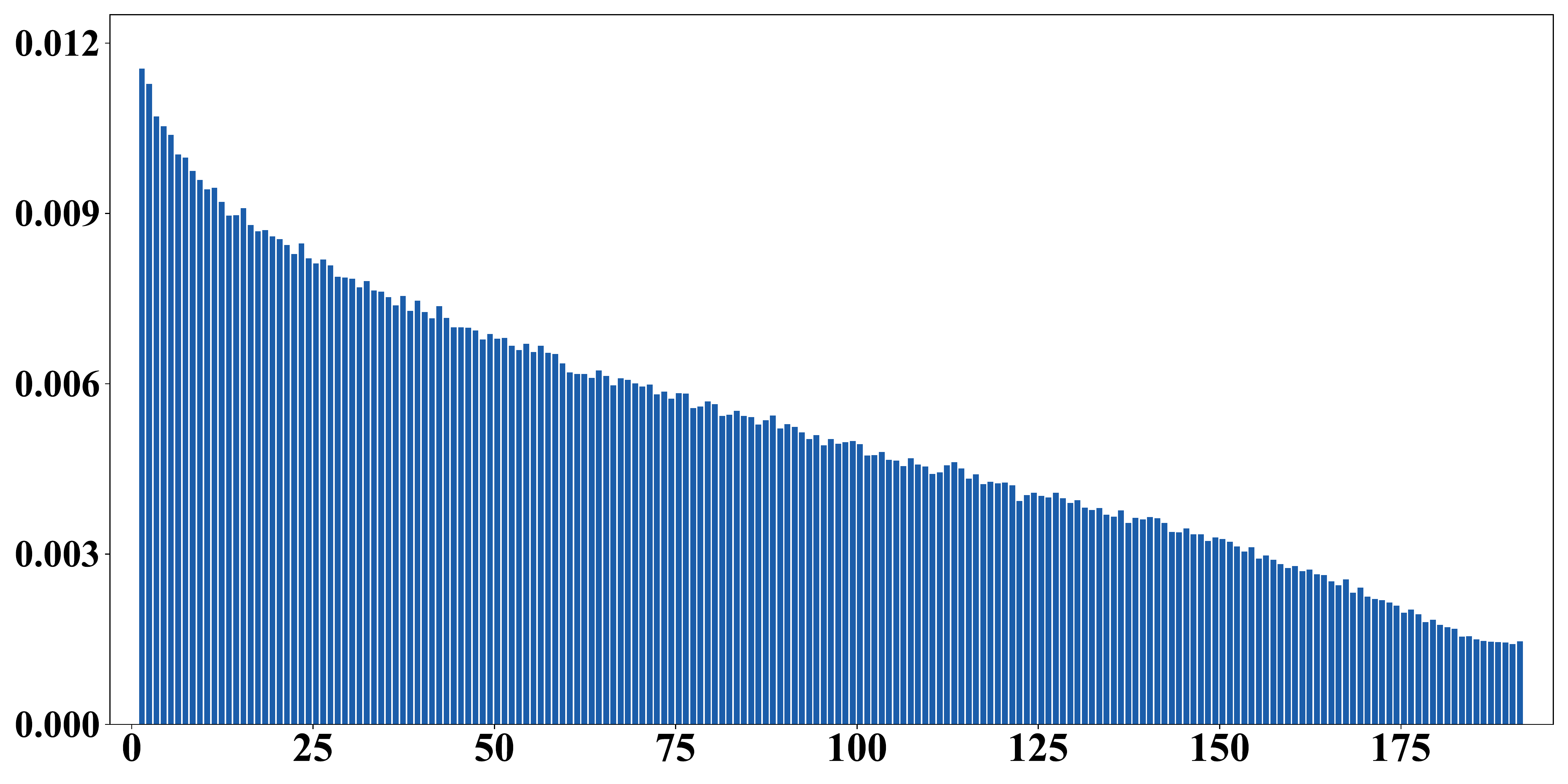}}

\caption{Performance over different future time horizons.}
\label{fig:discussion_days}
\end{figure}

In Section \ref{sec:exper}, we use the first week of original data for training and evaluate with the data of the following two weeks ($T^o=7 \ days \ \& \ T^f=14 \ days$). According to Table.~\ref{tab:percentage of accidental destination and potential destination}, individuals may present new potential destinations as the extension of $T^f$. On the other hand, evaluating with a shorter $T^f$ can better reflect the model's performance on potential destinations that individuals will visit in the short term. Thus to demonstrate the performance of our method more comprehensively, we conducted experiments with $T^f$ = 7, 21, and 28 days. The results are shown in Figure \ref{fig:discussion_days}, which shows the performance of PDPFKG is overall stable and effective under different $T^f$ (within 28 days).

\subsection{Application of other methods} \label{sec:other_method}

\subsubsection{Location-based methods}

There are location-based methods of human mobility field that consider the case of an individual choosing a new destination (or location). However, \cite{zhao2018individual} has presented a viewpoint that the location prediction problem is not particularly helpful for transportation applications. To explore it empirically, we select two popular models for location prediction problem and implement them on our task. One is the exploration and preferential rerun (EPR) model (\cite{song2010modelling}), which is a well known in human mobility domain. The other is the preferential exploration and preferential return (PEPR) (\cite{schlapfer2021universal}) model, which is an enhanced model based on EPR. It should be pointed out that the present location is required for these models to predict, which differs from PDPFKG, and it is the reason why we leave them out as baseline methods.

\begin{figure}[h]
\centering
\subfigure[Method based on EPR]{\includegraphics[width=0.46\textwidth]{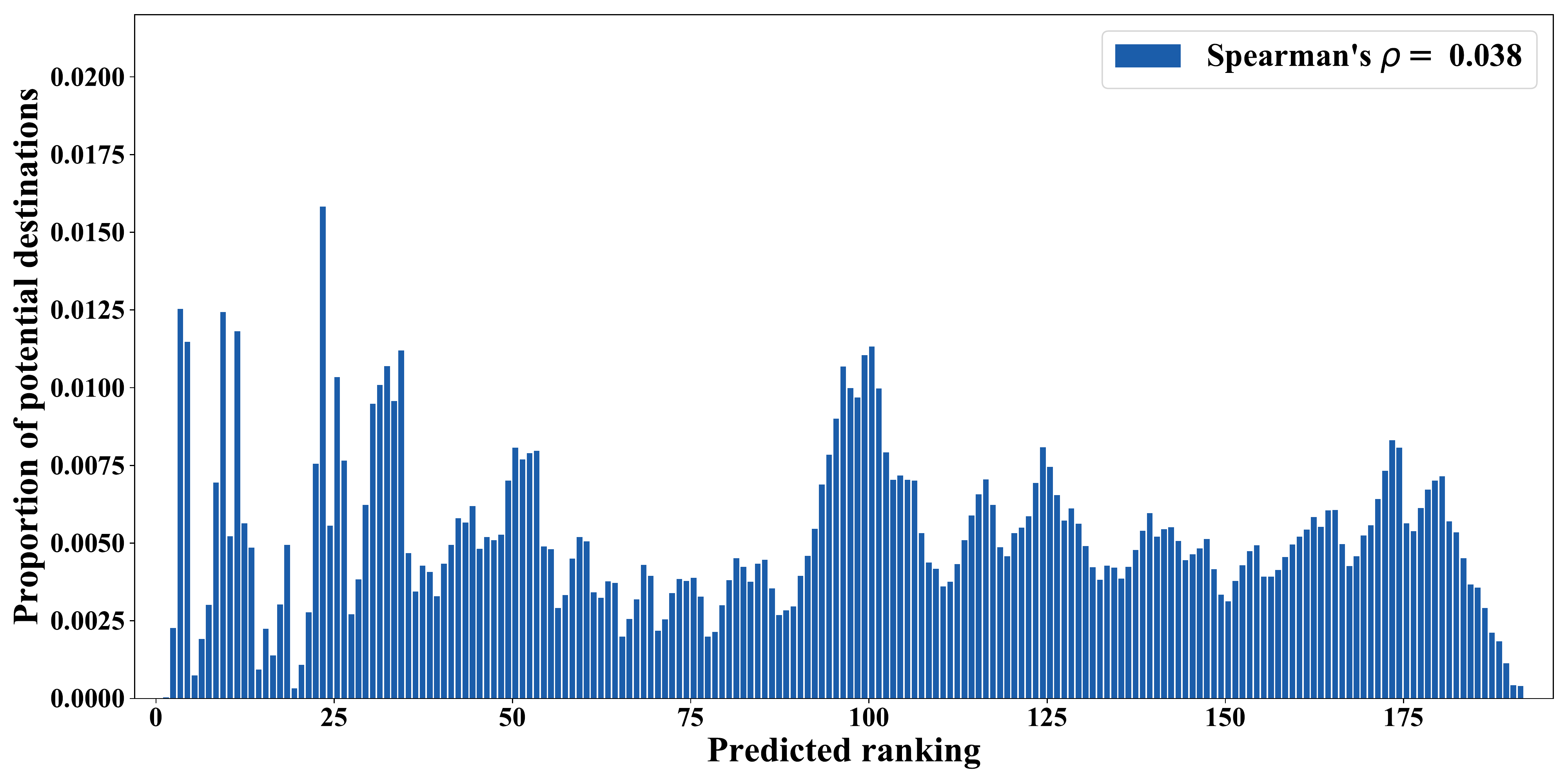}}
\subfigure[Method based on PEPR]{\includegraphics[width=0.46\textwidth]{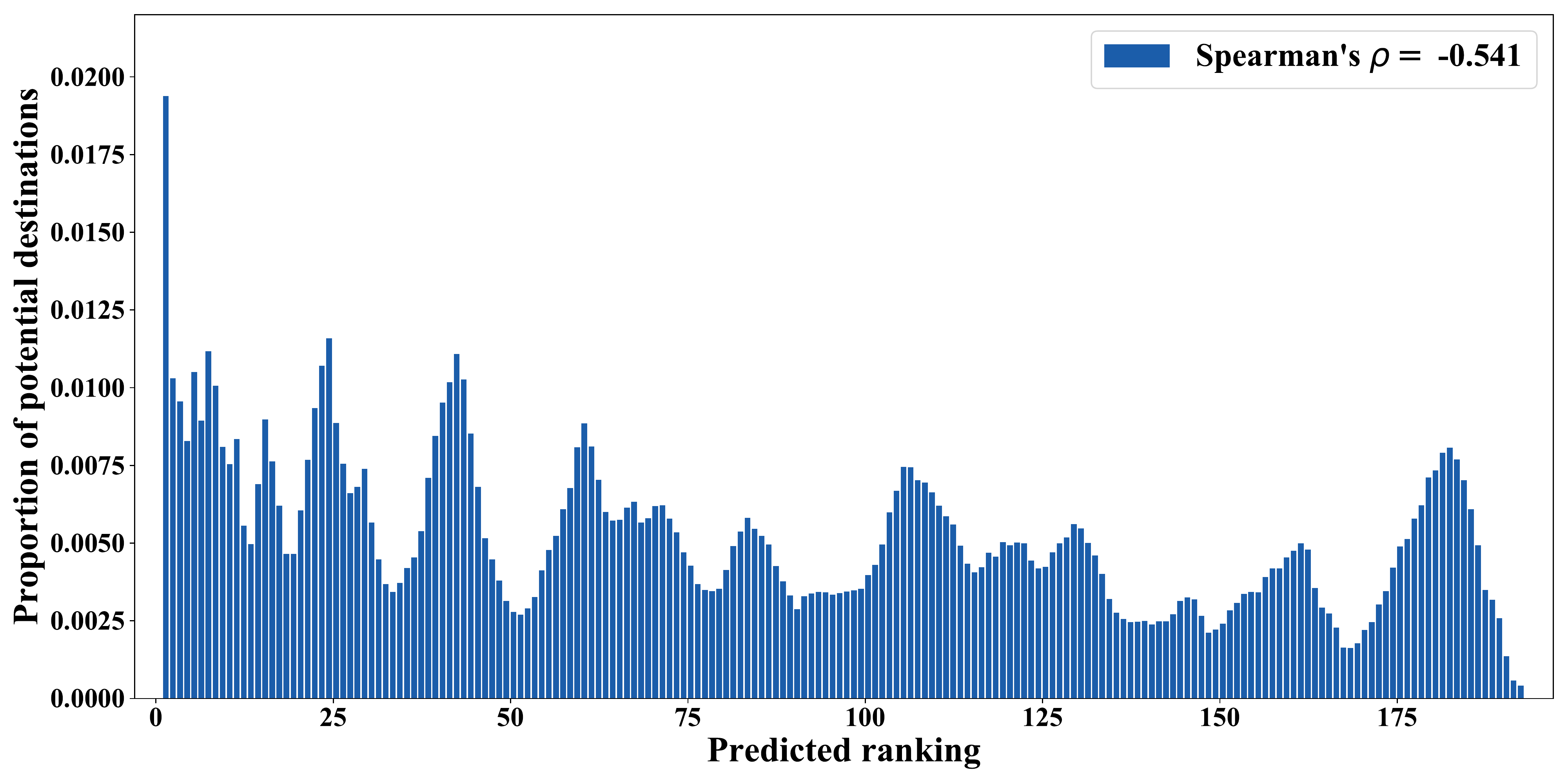}}
\caption{Performance of methods modified based on models in human mobility domain.}
\label{fig:epr}
\end{figure}

Due to the differences in the data used, we had to modify the two models for adaptation. First, we adopt the center point's coordinates (latitude and longitude) of traffic zones as the origin and destination, by which we obtain individuals' mobility records from location to location. Then the distribution of jump-size $\Delta j$ is needed. Both EPR and PEPR think aggregated $\Delta j$ follows a fat-tailed distribution proposed in \cite{brockmann2006scaling,gonzalez2008understanding}, and can be represented as $p(\Delta j)\sim |\Delta j|^{-1-\alpha}$ with $0<\alpha \leqslant 2$ ($\alpha$ needs to be calibrated). However, we found it is not adapted to urban transportation scenario after we get $\Delta j$ distribution, denoted as $J$, based on our first-week data (The distribution $J$ is shown in Appendix.~\ref{ap:app}. It should be noted that this distance is usually shorter than actual trip distance). From this perspective, this regularity of human mobility is not applicable to urban transportation. Next, we will make a prediction based on $J$. Given a present location of an individual, EPR model determines whether the individual returns to a previously visited location or explores a new location by $P_{new}=\rho S^{-\gamma}$. Here we only focus on the case of exploring new locations and assume the model has successfully judged it. For the new location $l_n$, EPR predicts it as a location at distance $\Delta j$ from the current location $l_o$, where $\Delta j$ is chosen from the $\Delta j$ distribution, and the direction is selected to be random. To align the predicted output with ours and keeping its idea, we modify this prediction logic as follows. For a data record that individual $v_n$ visited a new location (destination) $l_n$ after the first week, we set the coordinate of the origin traffic zone as present location $l_o$. For each $z_x \in Z-Z^o_n$, the distance to $l_o$ can be calculated by its coordinate $l_x$ with $l_o$, denote as $\Delta j_{o,x}$. Then we retrieve $p^J(\Delta j_{o,x})$ for each $z_x \in Z-Z^o_n$ from $J$, and take the ranking in descending order of $p^J(\Delta j_{o,n})$ as the prediction result of EPR model. Compared to EPR model, PEPR model is enhanced in the selection of directions. It preferentially selected directions towards regions of high visitation instead of random selection. For such a change, we take the combined ranking of the hotness ranking and the EPR's ranking as the predicted ranking of PEPR. Finally, we get the distributions $U$ of modified EPR and PEPR models and visualize them, see Fig.~\ref{fig:epr}.

The result of PEPR is better than EPR, but they all perform not well. We argue that this is mainly because urban vehicle travel distances ($\Delta j$) do not follow a pattern or the fat-tail distribution. Hence, it needs to be cautious to transfer some laws concluded by location-based studies to urban transportation field.

\subsubsection{Deep learning-based methods}

\begin{figure}[ht]
\setlength{\abovecaptionskip}{0.cm}
\centering
\includegraphics[width=0.9\textwidth]{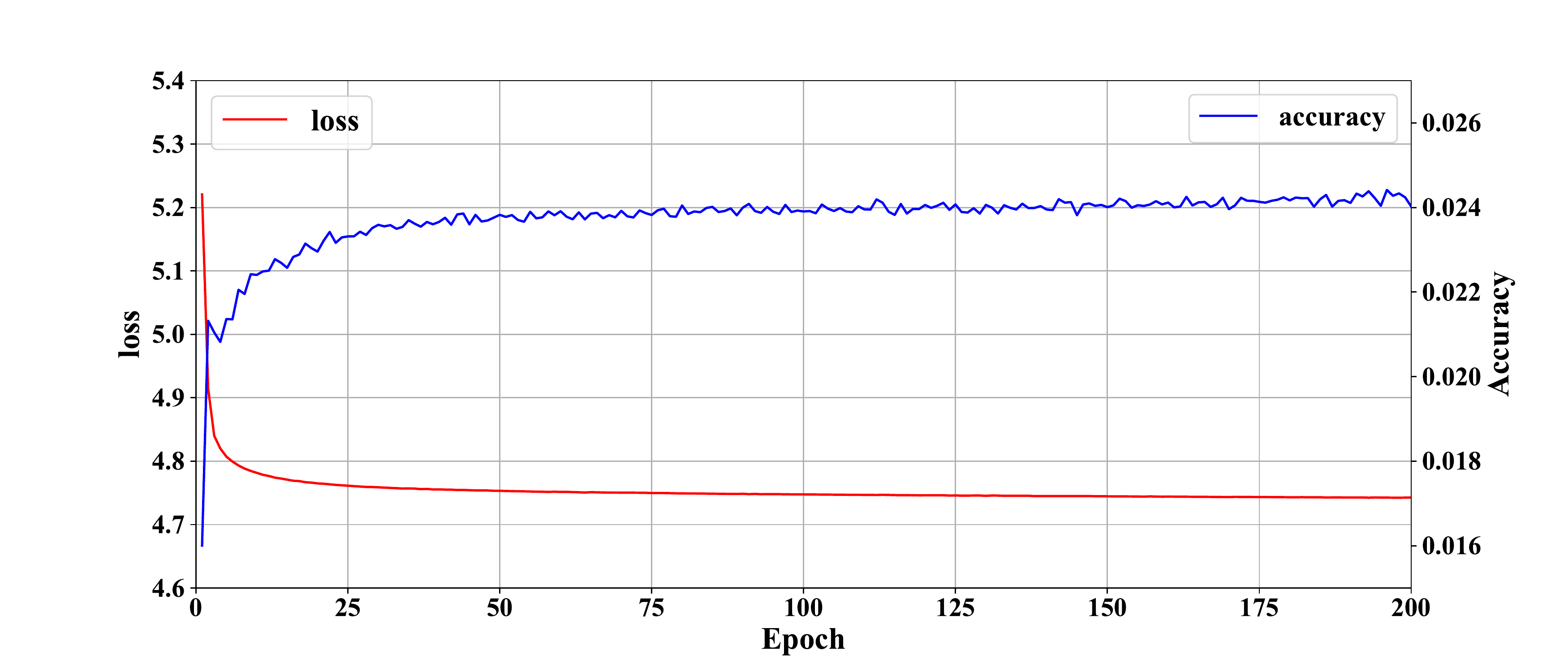}
\caption{Loss and accuracy of the adopted deep learning method.}
\label{fig:dl_loss}
\end{figure}

\begin{figure}[ht]
\setlength{\abovecaptionskip}{0.cm}
\centering
\includegraphics[width=0.75\textwidth]{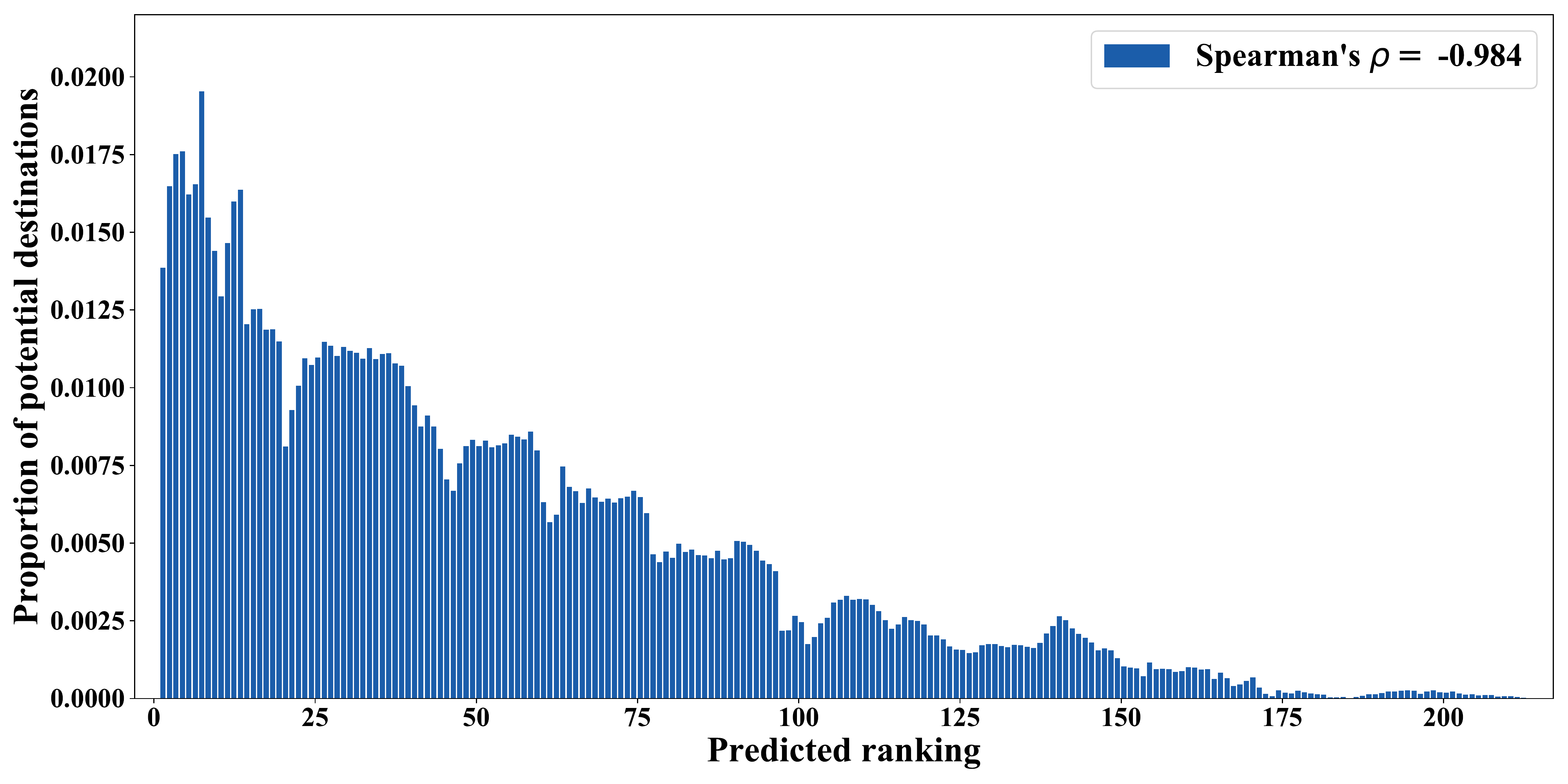}
\caption{Performance of the adopted deep learning model.}
\label{fig:dl_per}
\end{figure}

In the past, the methods that combine deep neural networks and supervised learning for end-to-end learning have achieved outstanding performance on destination prediction tasks. However, they are generally not good at handling individuals with low predictability (or sparse data) since the network under such a framework is essentially a statistical model, while these individuals' trip data does not show (or is not enough to find) the statistical regularity. Also, this is the reason why they always remove individuals with fewer trip records or high randomness.
To further illustrate experimentally, we construct a plain four-layers deep learning model and apply it to our task with the target individuals screened in Section.~\ref{sec:data scene}. This model is trained with the individual's identity as the feature and destination as the label to align the prediction logic (i.e., predict the ranking for each unobserved traffic zone given $v_n$). The cross-entropy loss is adopted for training, and a detailed parameter report is shown in Table.~\ref{tab:DL_par}, Appendix.~\ref{ap:app}. At the stage of prediction, we get unobserved traffic zones' rankings by the values of their neurons in the output layer (those already observed destinations do not participate in the ranking).
The loss and accuracy of the deep learning model during training are shown in Fig.~\ref{fig:dl_loss}. Besides, the $U$ refined by its predicted result is visualized in Fig.~\ref{fig:dl_per}. It performs very similarly to Fig.~\ref{fig:our method}(a), which is based on hotness ranking. Furthermore, we record the difference between each potential destination's ranking given by the deep learning model and its hotness ranking, denoted as $\Delta a$. After counting we found that $p(\Delta a \leqslant5)=0.981$ ($p(\Delta a =0)=0.312$; $p(\Delta a =1)=0.326$; $p(\Delta a =2)=0.181$; $p(\Delta a =3)=0.891$; $p(\Delta a =4)=0.048$; $p(\Delta a =5)=0.024$), and $  \Delta a \leqslant 21 \ \forall \Delta a$, which means what the deep learning model learns is statistical information. In other words, under low predictability data conditions, the adopted deep learning model degenerates into a simple statistical model.

Recently, some studies in other fields have dealt with data sparsity using deep learning methods. Their ways to achieve this, especially the ways of handling cold-start individuals (or users) in recommendation systems field, is worthy of attention. Basically, most of these researches relies on prior knowledge to make predictions on individuals with few or no records. For instance, based on the prior knowledge that people tend to explore the POIs near the ones they favour and the probability of a user visiting a POI is largely affected by local word-of-mouth about the POI when users travel in unfamiliar regions, \cite{yin2017spatial} extracted the neighborhood and popularity features of POIs, which are significant information for model training. Besides, this study believed users with the same role at a region are more likely to have similar preferences, and the cold-start users were predicted by the collective preferences of the crowd with the same role. In \cite{qian2019spatiotemporal}, the normalized ratio of common friends in two users’ social circles was introduced to handle cold-start users, which is based on the knowledge or common sense that the social relation will affect an individual's choice of POIs. In addition, with prior knowledge, specialized networks and mechanisms for training can also be developed. For example, the model proposed by \cite{yin2017spatial} employed unsupervised learning for representation learning first and adopted supervised learning in the last step.

In summary, prior knowledge and corresponding data are required for most deep learning methods to combat data sparsity, which can guide the development of models (e.g., network structures and optimize objectives). However, as we mentioned in Section.2, there is currently little common sense and no strong prior knowledge about the individual's potential destinations choice behavior to rely on. But we believe that the deep learning method has potential for this topic as the study progresses and more knowledge is discovered. Moreover, incorporating deep learning and knowledge graphs (e.g.,\cite{guo2020attentional}) is a promising way to enhance performance. 




\subsection{Advantages of knowledge graph} \label{dis:discuss2}

In this section, we would like to discuss some advantages of knowledge graph on the individual's trip prediction and knowledge discovery knowledge graphs.

First, knowledge graph is very powerful at organizing data and expressing associated information. Except for spatio-temporal information, which is commonly used for individual trip analyzing and predicting, there are many other factors that influence individual trip behaviors in different ways. They may be generated from other natural systems like the weather system or human societies (e.g., social relationships). Thus, a model that can organize these heterogeneous data and express various relationships is needed, which is the knowledge graph specialty. For instance, if individuals' social relationships are available, then it can be introduced to TKG by adding the triple $(Veh\_id)\leftarrow[social relationship]\rightarrow(Veh\_id)$.

Second, knowledge graph-based prediction logic is flexible and allows inconsistent inputs in the training and prediction stages. Many existing methods, especially end-to-end supervised ones, require the inputs of the training and prediction stages to be aligned. That is, all the information used for training must be given when predicting. This makes the information that is available but not accessible at the prediction stage cannot be used. Also, once an item of features is missing, the prediction will fail. However, for knowledge graph-based models, the prediction logic can be adjusted through the schema design, and inconsistencies in the inputs of the two stages are allowed. For example, if we want to change the prediction logic to a destination prediction based on the individual with its trip time, modifying the core triple to $(Veh\_id)-[trip\_time]\rightarrow(Zone)$ will achieve it. Besides, knowledge discovery tasks can be supported by specific schema. For instance, under the schema of TKG, we train the model using spatio-temporal and POI information, while the prediction only needs the individual. In this case, it is more like obtaining knowledge attached to individuals through training using rich data.

Lastly, there are mature knowledge graph technologies to handle traffic elements' complex semantics. The semantics of traffic elements are complex. For instance, travelers who visit the same traffic zone may have different purposes. The pattern of individual trip behavior is also not stable, such as it may behave differently on weekdays versus holidays. Besides, the meaning of the same period will be varied among travelers. These require models capable of handling complex semantics. Otherwise, it would be misdirected. We think one of the factors leading to the poor performances of CF and MD, especially bad performances in the tail of $U$, is that they all default to things having only a single meaning. In contrast, as we mentioned in Section.~\ref{sec:embedding1}, knowledge graph embedding algorithm TransH allows the entities have different representations on different relationships, by which it is able to handle the complex semantics of trip scenarios. In addition to TransH, different knowledge graph-based models with different capabilities and computational performance are available.

\section{Conclusion and future works}

In this paper, we propose a knowledge graph-based potential destination discovery (or prediction) framework PDPFKG for low predictability individuals. To associate data of individuals, we first construct a trip knowledge graph (TKG) by organizing original data by adopting knowledge graph.
When building relationships, we introduce the private relationship. 
It reduces and balances the complexity of the relationships and makes their complexity independent of the data scale, ensuring the consistency of the optimal dimension of relationships. Further, we apply a specialized embedding model to TKG and get the trip knowledge graph embedding model (TKGEM). In TKGEM, entities and relationships are parameterized and computable. The specialization of the generic model mainly includes training strategies and optimization objectives, which are adapted to TKG and our task. Benefitting from the schema of TKG, the training of TKGEM can achieve overall optimization.
Next, we implement PDPFKG on low predictability individuals in a city-scale vehicular dataset of the real world. Experiments demonstrate PDPFKG is valid and overall performs superior to other methods. In detail, the predicted ranking of individuals' unobserved destinations given by PDPFKG strongly correlates with their choice behavior. In particular, the smooth aggregated ranking distribution of potential destinations with a very low confusion degree implies individuals' behavior in choosing potential destinations may follow a certain pattern (from the association perspective). Although PDPFKG's performance on concentration degree (recall) is not yet very well, experiments show that it is a scientific framework with great potential for improvement. Lastly, we thoroughly discuss the innovation points of this paper or critical points of the methodology. Also, we discuss the applicability and potential of other methods on our topic and the advantages of knowledge graph.

Our future work will focus on the following two topics: 1) Explore how to improve the performance of PDPFKG. Section.~\ref{sec:ex_eva_ini} and \ref{sec:non_coretriple} have shown additional information can improve the performance of PDPFKG. Next, we will consider introducing more data types, such as the adjacency between traffic zones.
2) Explore the correlation between group sizes and the predictability of low-predictability individuals. PDPFKG indicates that individuals with low predictability are not completely unpredictable when put in a group. It can be determined that there is a correlation between the size of the group and predictability. It would be a meaningful effort to reveal the relationship between them.

\section*{Acknowledgments}

This research was supported by the National Natural Science Foundation of China (No. U21B2090 and No. U1811463).

\newpage


\bibliography{mybibfile}

\newpage

\begin{appendices}
\section{}\label{ap:app}

\begin{figure}[ht]
\setlength{\abovecaptionskip}{0.cm}
\centering
\includegraphics[width=0.55\textwidth]{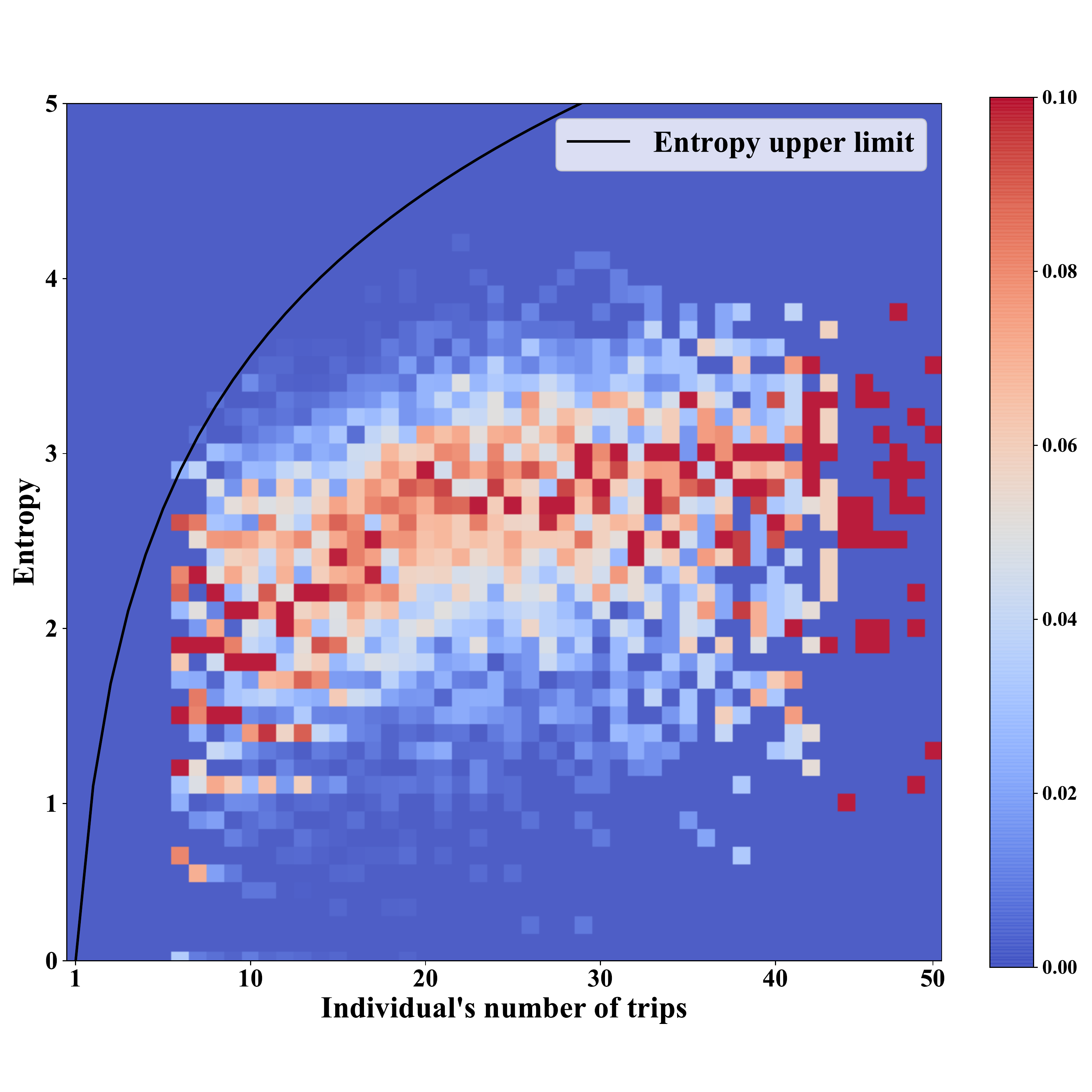}
\caption{Entropy of destinations of eliminated individuals.}
\label{fig:shang_tq7}
\end{figure}

\begin{figure}[ht]
\centering
\subfigure[Accidental destination]{\includegraphics[width=0.46\textwidth]{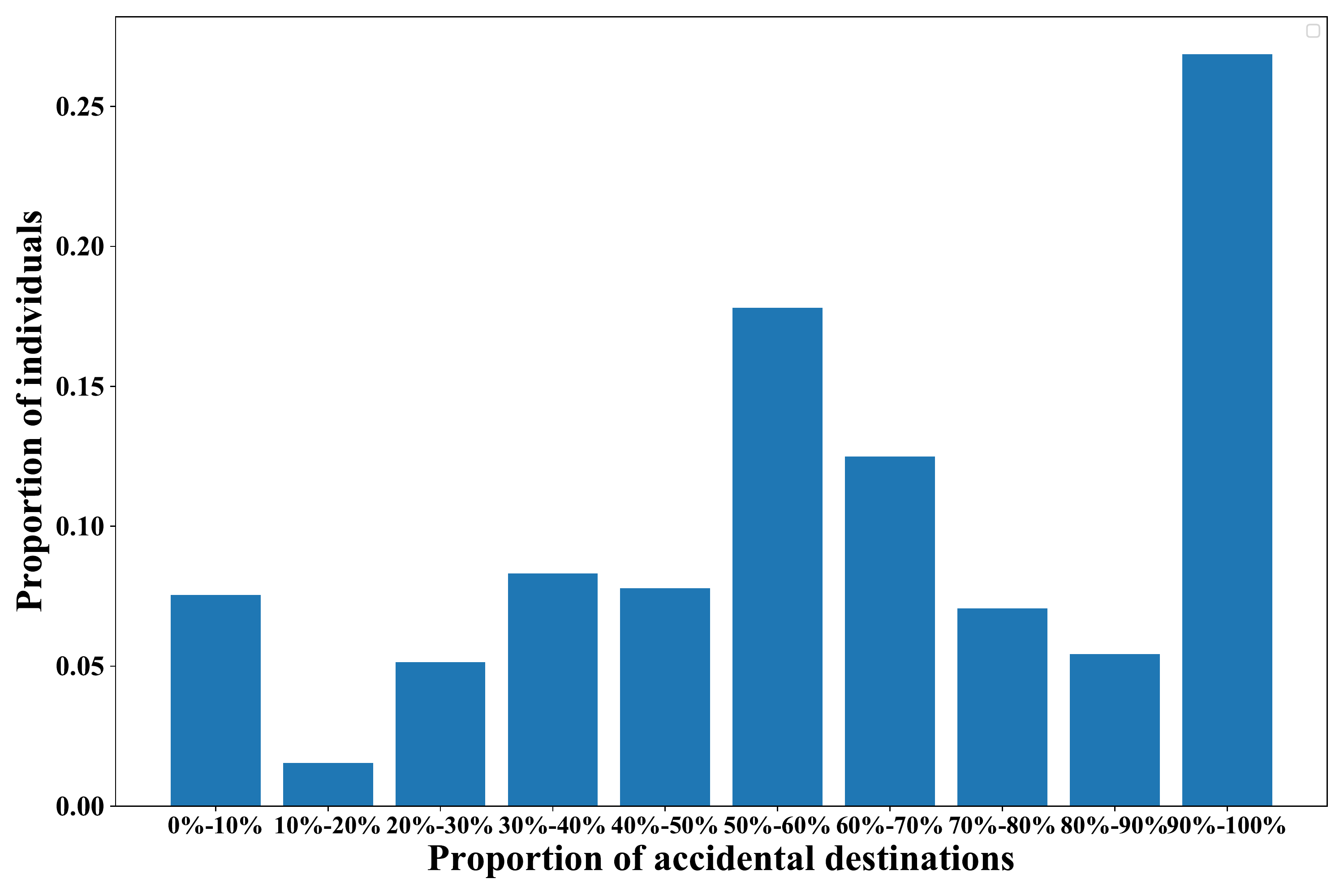}}
\subfigure[Potential destination]{\includegraphics[width=0.46\textwidth]{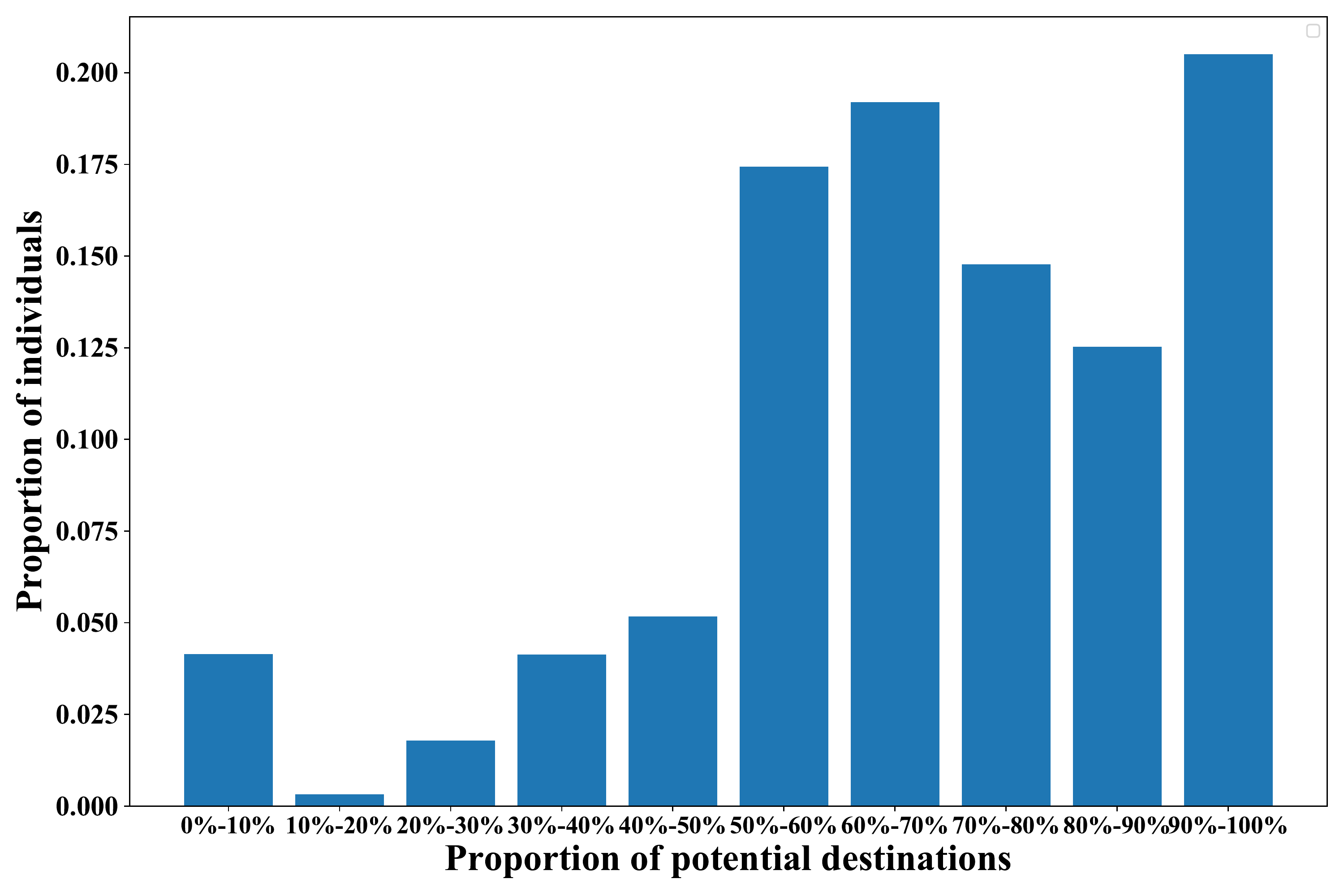}}
\caption{Proportional distribution of accidental and potential destinations with $T^o$=7days and $T^f$=14days.}
\label{fig:acc_po}
\end{figure}

\begin{figure}[ht]
\setlength{\abovecaptionskip}{0.cm}
\centering
\includegraphics[width=0.9\textwidth]{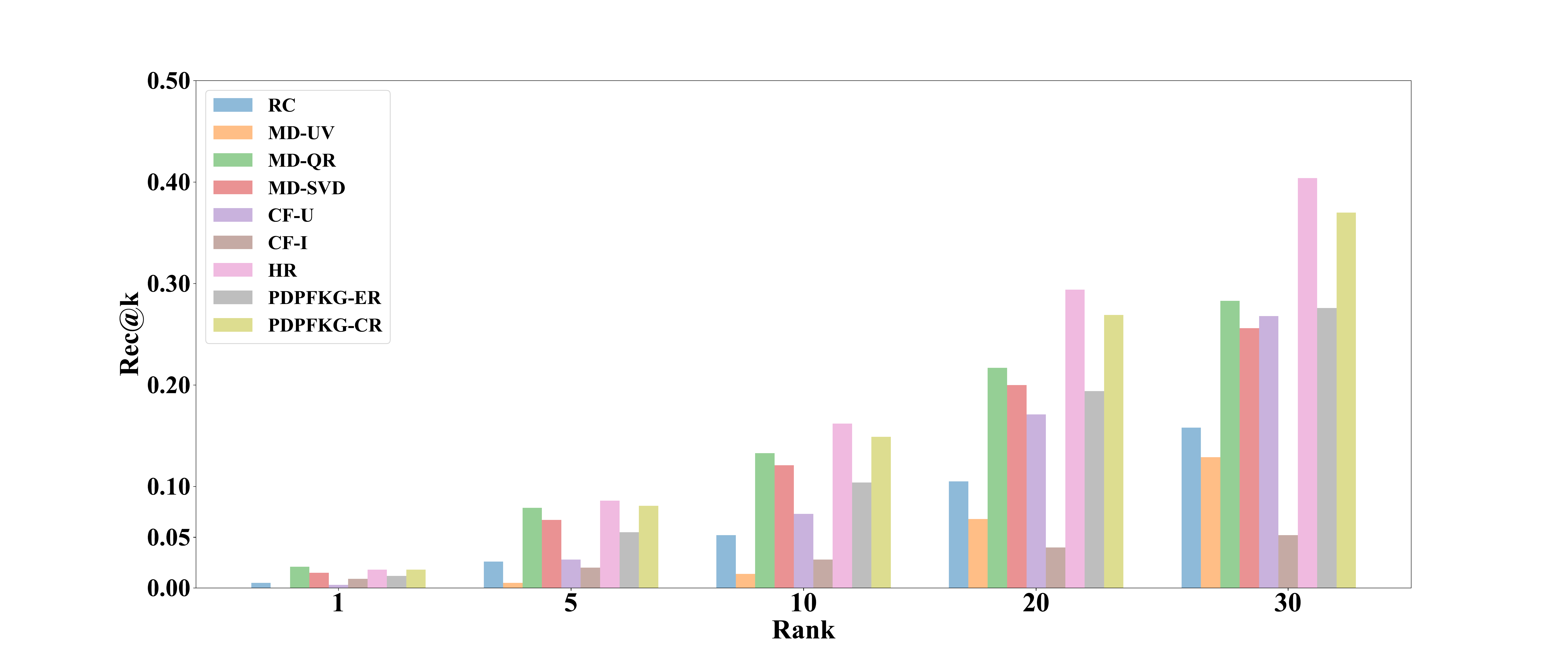}
\caption{Comparisons with different methods in terms of Rec$@$k.}
\label{fig:accurecy}
\end{figure}

\begin{figure}[ht]
\setlength{\abovecaptionskip}{0.cm}
\centering
\includegraphics[width=0.9\textwidth]{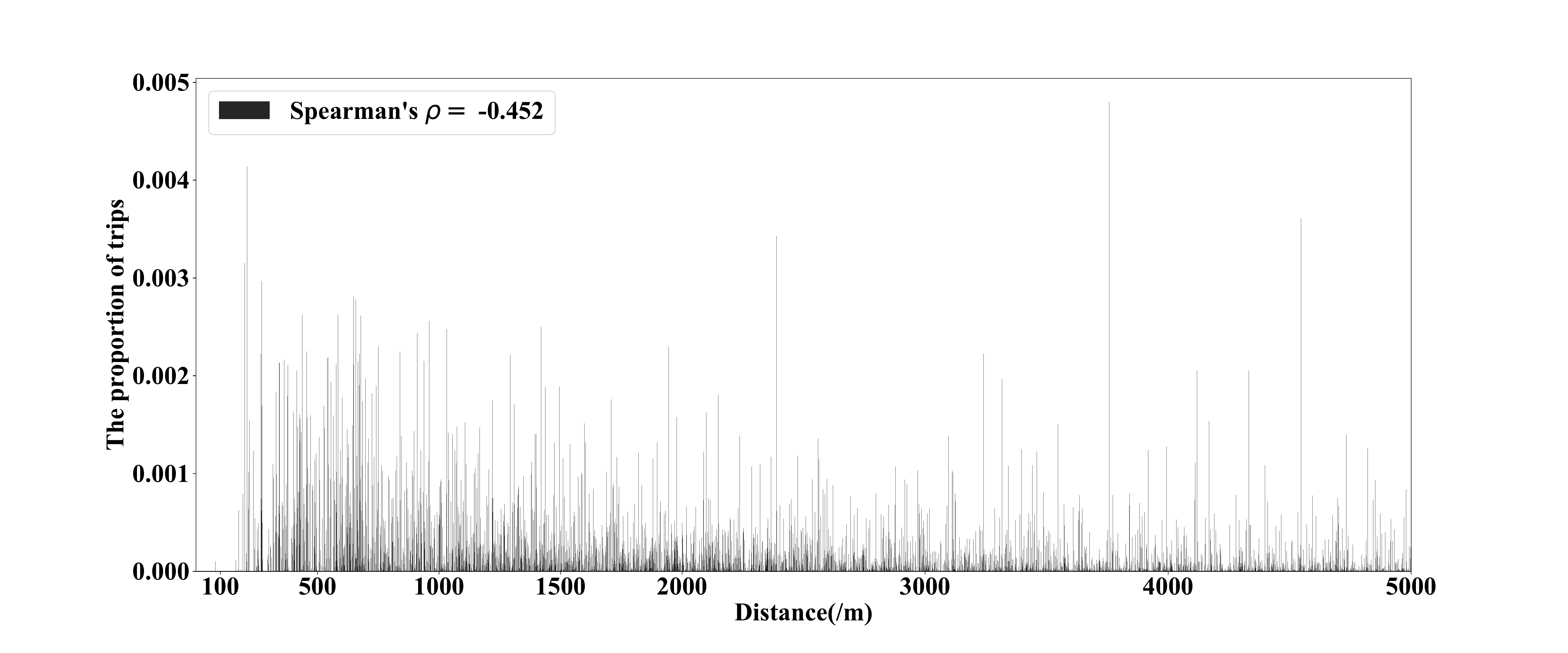}
\caption{Distribution of individual mobility distance in XuanCheng city.}
\label{fig:distance_dis}
\end{figure}

\begin{table}[ht]
\centering
\caption{Parameter report of the adopted deep learning model.}
\label{tab:DL_par}
\begin{tabular}{l|c}
\hline
\textbf{Parameter}      & \multicolumn{1}{c}{\textbf{Value}}      \\ 
\hline
Layers & 4 \\
Embedding Layer Size & 148 \\
Hidden Layer Sizes & 400,200   \\
Training Batch Size & 128  \\
Learning Rate  & 0.001   \\
Epochs & 200  \\
Optimizer   & Stochastic Gradient Descent (SGD)  \\ \hline

\end{tabular}
\end{table}

\end{appendices}

\end{document}